\documentclass{article}

\usepackage[nonatbib,preprint]{neurips_2024}
\usepackage[numbers]{natbib}

\usepackage[utf8]{inputenc} 
\usepackage[T1]{fontenc}    
\usepackage{hyperref}       
\usepackage{url}            
\usepackage{booktabs}       
\usepackage{amsfonts}       
\usepackage{nicefrac}       
\usepackage{microtype}      
\usepackage{xcolor}         

\usepackage{multirow}
\usepackage{amsmath}
\usepackage{makecell}
\usepackage{multirow}
\usepackage{mathrsfs}
\usepackage[ruled]{algorithm2e}
\usepackage{cellspace}

\usepackage{makecell}
\usepackage{bm}
\usepackage{graphicx}
\usepackage{paralist}
\usepackage{enumitem}
\usepackage[utf8]{inputenc} 
\usepackage[T1]{fontenc}    
\usepackage{hyperref}       
\usepackage{url}            
\usepackage{booktabs}       
\usepackage{amsfonts}       
\usepackage{nicefrac}       
\usepackage{microtype}      
\usepackage[ruled]{algorithm2e}
\usepackage{threeparttable}

\usepackage{graphicx}
\usepackage{subfigure}
\usepackage{booktabs} 
\usepackage{caption}
\usepackage{color}
\usepackage{framed}
\usepackage{amssymb}
\usepackage{amsfonts}
\usepackage{mathrsfs}
\usepackage{mathtools}
\usepackage{array}
\usepackage{amsthm}
\usepackage{verbatim} 
\usepackage{enumerate}
\usepackage{bbm}
\usepackage{commath}
\usepackage{wrapfig}
\usepackage{amsbsy}
\usepackage{float}
\usepackage{amsmath}
\usepackage{tabularx}
\usepackage{listings}
\usepackage{xcolor}
\usepackage{colortbl}
\usepackage{tcolorbox}
\usepackage{bbding}
\usepackage{multirow}
\usepackage[toc,page,header]{appendix}
\usepackage{minitoc}
\usepackage{titletoc}
\usepackage{CJKutf8}
\usepackage{subcaption}	

\newcommand{\vpara}[1]{\vspace{0.07in}\noindent\textbf{#1 }}
\newcommand{\data}{VisScience\xspace}

\title{VisScience: An Extensive Benchmark for Evaluating K12 Educational Multi-modal Scientific Reasoning}

\author{%
  Zhihuan Jiang$^{13\dagger*}$,
  Zhen Yang$^{13\dagger*}$,
  Jinhao Chen$^{23\dagger}$,
  Zhengxiao Du$^{13}$,
  Weihan Wang$^{13}$,\\
  \textbf{Bin Xu$^{1}$,
  Jie Tang$^{1}$} \\ \\
  \textsuperscript{1}Tsinghua University \quad
  \textsuperscript{2}Beihang University \quad
  \textsuperscript{3}Zhipu.AI 
\\ \\
}

\begin{document}

\maketitle

\renewcommand{\thefootnote}{\fnsymbol{footnote}}
    \footnotetext[1]{ZHJ and ZY contributed equally. Emails: \texttt{\{jiang-zh21, yangz21\}@mails.tsinghua.edu.cn}}
    \footnotetext[2]{Work done while Zhihuan Jiang, Zhen Yang and Jinhao Chen interned at Zhipu AI.}
\renewcommand{\thefootnote}{\arabic{footnote}}

\begin{abstract}

Multi-modal large language models (MLLMs) have demonstrated promising capabilities across various tasks by integrating textual and visual information to achieve visual understanding  in complex scenarios. Despite the availability of several benchmarks aims to evaluating MLLMs in tasks from visual question answering to complex problem-solving, most focus predominantly on mathematics or general visual understanding tasks. This reveals a critical gap in current benchmarks, which often overlook the inclusion of other key scientific disciplines such as physics and chemistry. To address this gap, we meticulously construct a comprehensive benchmark, named \data, which is utilized to assess the multi-modal scientific reasoning across the three disciplines of mathematics, physics, and chemistry. This benchmark comprises 3,000 questions drawn from K12 education — spanning elementary school through high school — equally distributed across three disciplines, with 1,000 questions per discipline. The questions within \data span 21 distinct subjects and are categorized into five difficulty levels, offering a broad spectrum of topics within each discipline. With \data, we present a detailed evaluation of the performance of 25 representative MLLMs in scientific reasoning. Experimental results demonstrate that closed-source MLLMs generally outperform open-source models. The best performance observed include a 53.4\% accuracy in mathematics by Claude3.5-Sonnet, 38.2\% in physics by GPT-4o, and 47.0\% in chemistry by Gemini-1.5-Pro. These results underscore the strengths and limitations of MLLMs, suggesting areas for future improvement and highlighting the importance of developing models that can effectively handle the diverse demands of multi-modal scientific reasoning.

\end{abstract}

\section{Introduction}

\begin{figure}[htbp]
\centering
\begin{minipage}{.55\textwidth}
  \centering
  \includegraphics[width=1\linewidth]{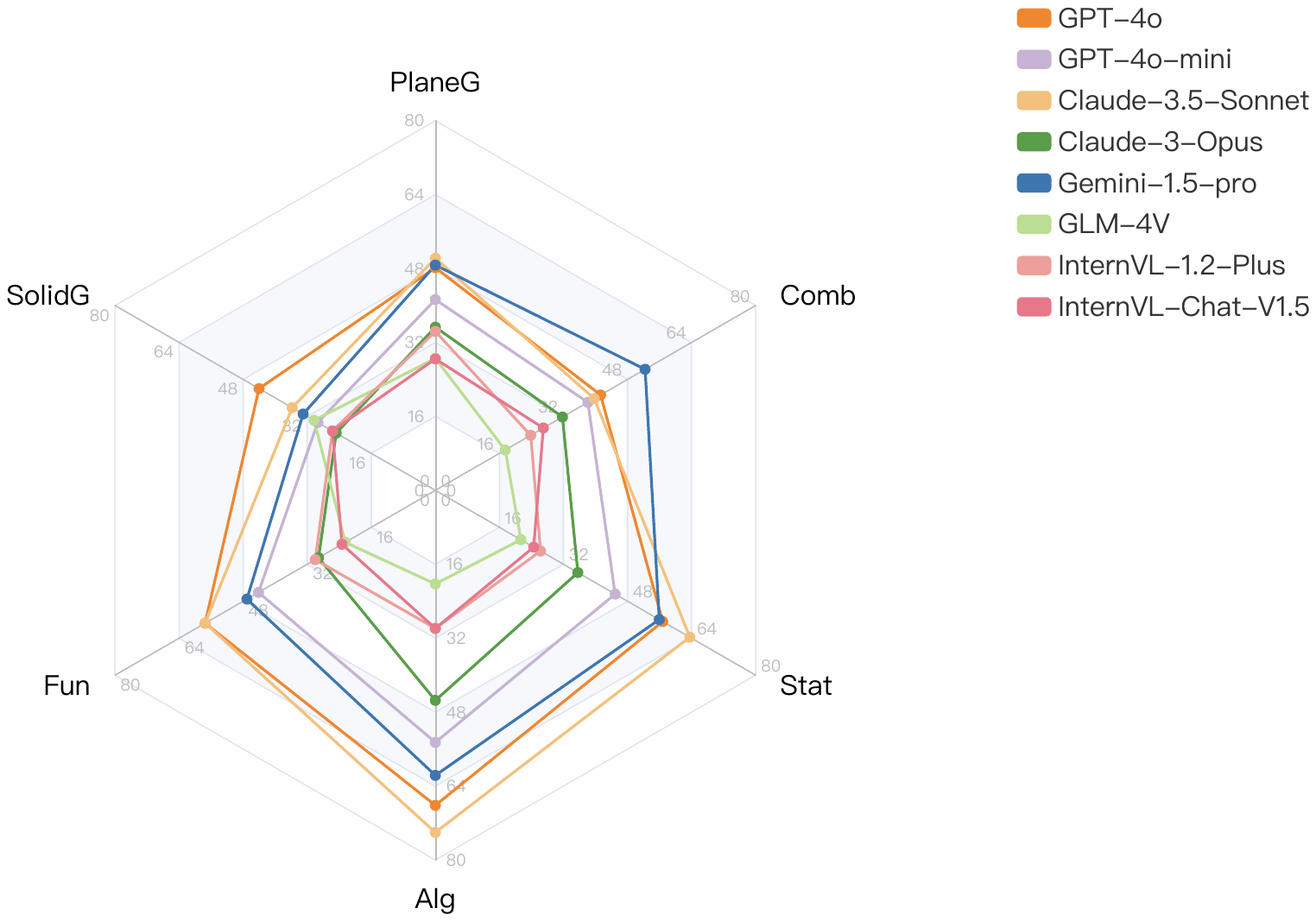}
  \vspace{-0.1cm}
\end{minipage}%
\begin{minipage}{.45\textwidth}
  \centering
  \includegraphics[width=1\linewidth]{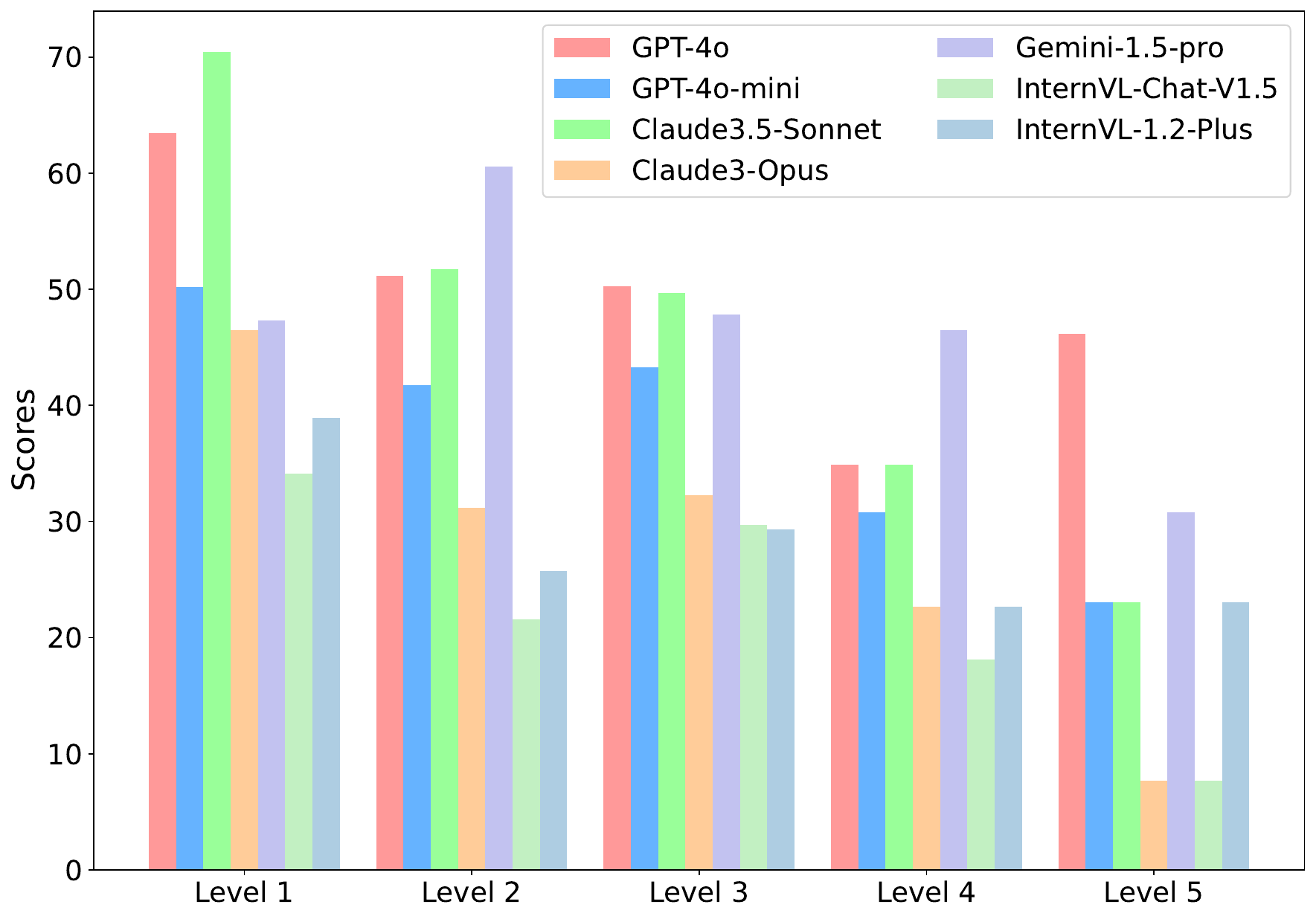}
  \vspace{-0.1cm}
\end{minipage}
\vspace{0.5cm}
(a) Mathematics
\centering
\begin{minipage}{.55\textwidth}
  \centering
  \includegraphics[width=1\linewidth]{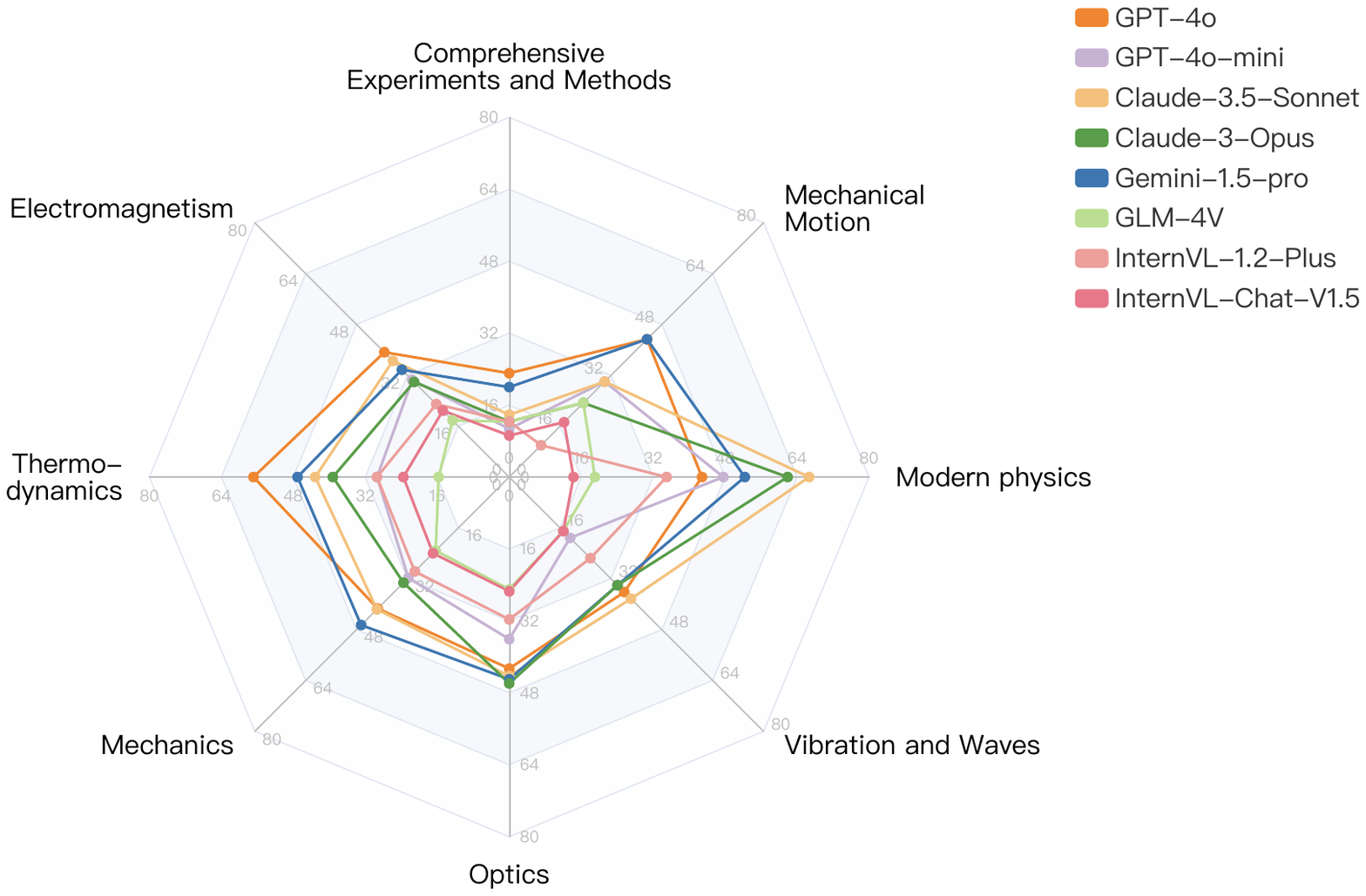}
  \vspace{-0.4cm}
\end{minipage}%
\begin{minipage}{.45\textwidth}
  \centering
  \includegraphics[width=1\linewidth]{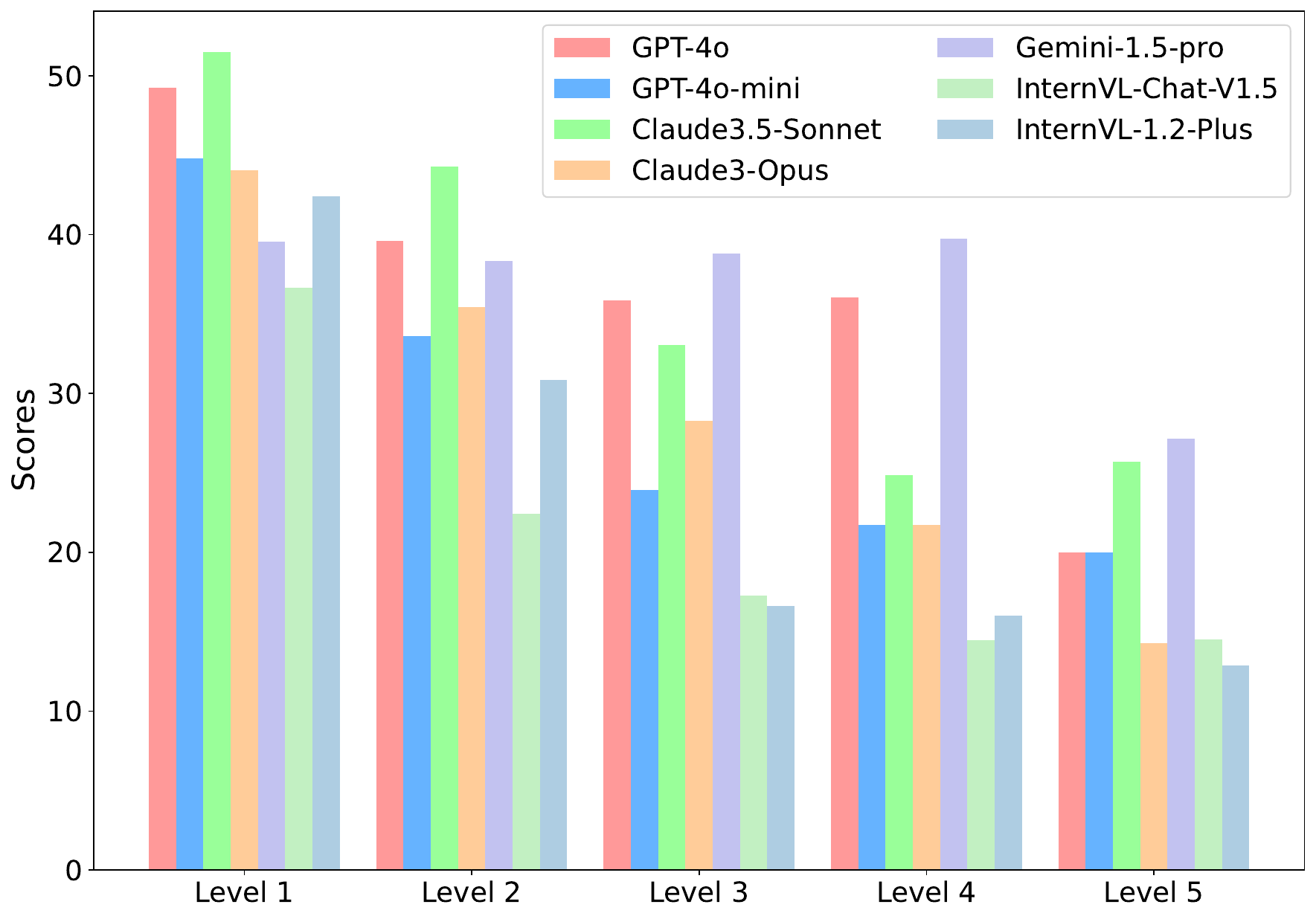}
  \vspace{-0.4cm}
\end{minipage}
\vspace{0.5cm}
(b) Physics
\centering
\begin{minipage}{.55\textwidth}
  \centering
  \includegraphics[width=1\linewidth]{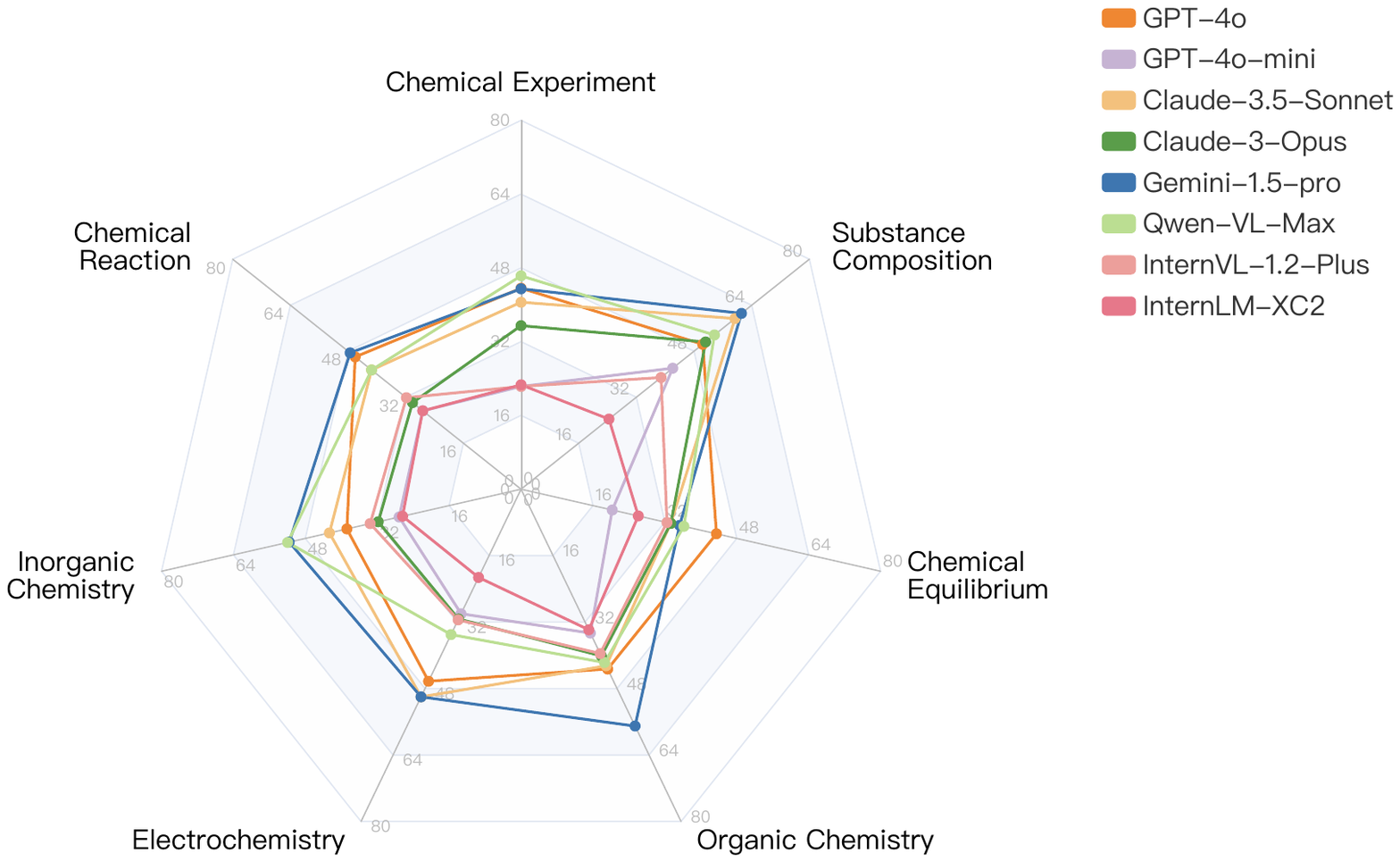}
  \vspace{-0.4cm}
\end{minipage}%
\begin{minipage}{.45\textwidth}
  \centering
  \includegraphics[width=1\linewidth]{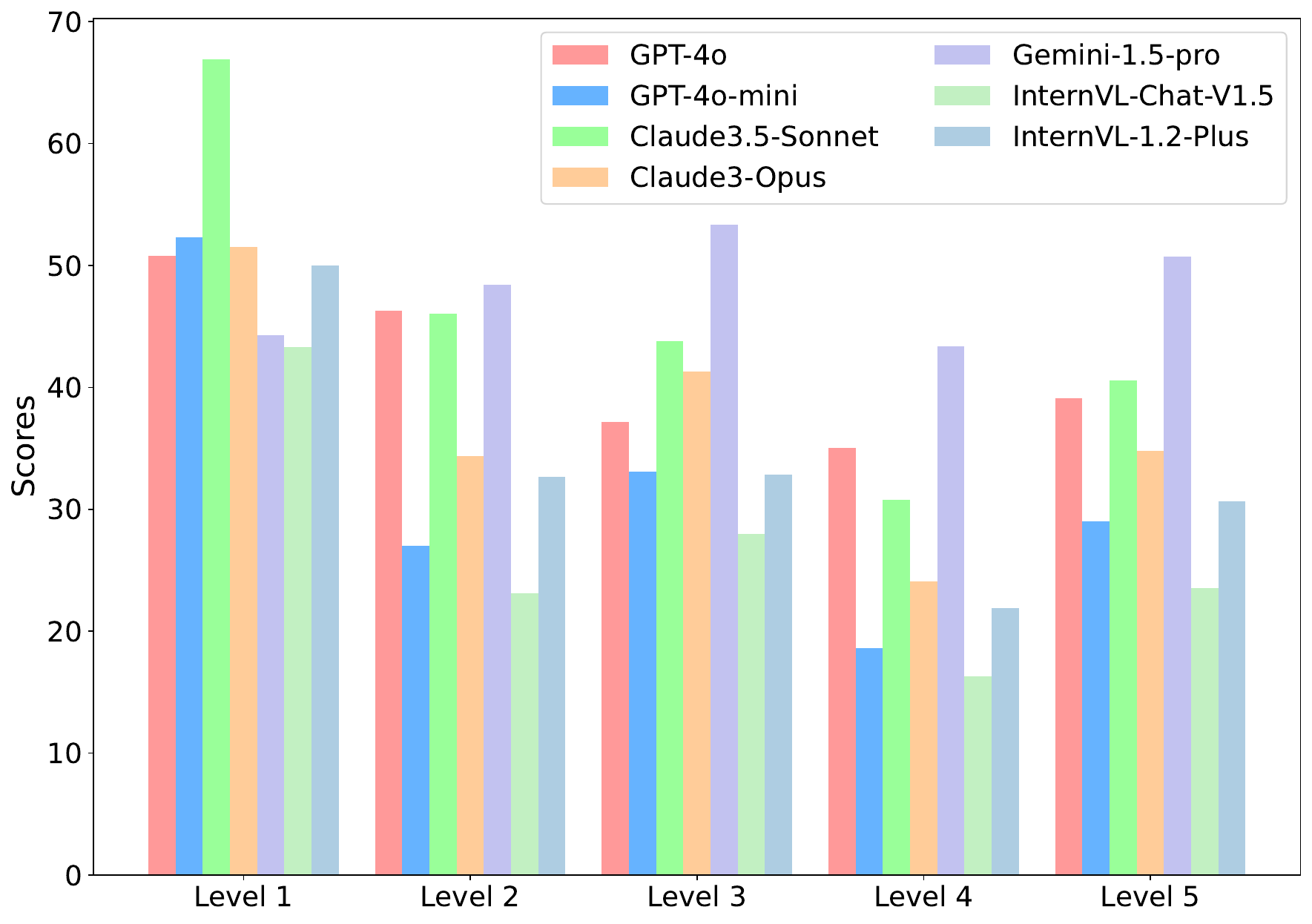}
  \vspace{-0.4cm}
\end{minipage}
(c) Chemistry
\caption{The accuracies of representative MLLMs on \data across different subjects and difficulty levels. (Left) The accuracies on different subjects. (Right) The accuracies on various difficulty levels.}
\label{fig:data_performance}
\end{figure}

Recently, large language models (LLMs)~\cite{openai2022chatgpt,achiam2023gpt,glm2024chatglm,touvron2023llama1,touvron2023llama2,bai2023qwen,brown2020language,chowdhery2023palm,anil2023palm2} have demonstrated remarkable capabilities across a wide range of tasks, including natural language understanding, text generation, and complex problem solving. The success of LLMs facilitates the development of multi-modal large language models (MLLMs)~\cite{openai2023gpt4v,team2023gemini,Claude3,liu2024visual,liu2024improved,ye2023mplug,ye2024mplug}, which extends these capabilities by integrating the ability to process and analyze both textual and visual information. Evaluation is a significant component in assessing the ability of these MLLMs across various tasks, which has attracted widespread attention and developed rapidly in recent years. For instance, several benchmark datasets are proposed to evaluate the ability of MLLMs in general visual understanding, including MME~\cite{fu2023mme}, MMMU~\cite{yue2024mmmu}, MMBench~\cite{liu2023mmbench}, MMStar~\cite{chen2024we}, and SEED-Bench~\cite{li2023seed}.

As a primary evaluation domain, mathematical reasoning presents specific challenges, requiring models to handle complex mathematical problems accompanied by visual information. Previous works~\cite{chen2021geoqa,chen2022unigeo,cao2022augmented} focus on geometric problems, resulting in the emergence of various evaluation datasets such as GeoQA~\cite{chen2021geoqa}, Geometry3K~\cite{lu2021inter}, and UniGeo~\cite{chen2022unigeo}. Subsequently, several benchmark datasets~\cite{lu2023mathvista,zhang2024mathverse,wang2024measuring} extend the scope of mathematical reasoning beyond geometry to encompass various branches such as arithmetic, algebraic, statistics, logic, and functions. Notably, MathVista also contains a portion of scientific datasets such as TQA~\cite{tqa}, SciBench~\cite{wang2023scibench}, and ScienceQA~\cite{lu2022learn}. However, despite these advancements, there remains some issues:

\begin{itemize}
    \item \textit{Existing benchmarks often focus narrowly on specific mathematics, neglecting other crucial scientific disciplines like physics and chemistry.}

    \item \textit{Existing benchmarks are often collected from limited sources, resulting in a lack of natural difficulty levels and leading to an incomplete evaluation of models' capabilities.}    

    \item \textit{Current benchmarks are predominantly available in a single language, limiting the evaluation of MLLMs' multilingual capabilities.}    
\end{itemize}

To address the limitations of existing benchmarks and provide a more comprehensive evaluation benchmark, we introduce a more expansive evaluation benchmark, named \textbf{\data}, integrating both textual and visual information. This benchmark is designed to assess the performance of MLLMs in multi-modal scientific reasoning tasks across disciplines like physics and chemistry alongside mathematics. To construct this benchmark, we gather a total of 450,000 questions from K12 education and meticulously select 3,000 questions as the final dataset, with each discipline containing 1,000 questions. This benchmark spans a comprehensive range of knowledge points across different chapters, with difficulty levels ranging from 1 to 5, ensuring that models are assessed on both basic and challenging problems.

In order to better understand MLLMs' performance on more detailed subjects within three disciplines, we categorize \data into several subjects across each discipline. Specifically, we divide the mathematical part of \data into six subjects such as \textit{plane geometry, solid geometry, functions and equations, algebraic operations, probability and statistics, and combinatorial mathematics}. For physics, the dataset is categorized as eight subjects, including \textit{mechanics, thermodynamics, comprehensive experiments and methods, mechanical motion, vibration and waves, optics, electromagnetism, and modern physics}. The chemistry section includes seven topics such as \textit{chemical experiments, organic chemistry, material composition, electrochemistry, chemical reactions, inorganic chemistry, and chemical equilibrium}. In summary, \data contains 21 subjects across the three disciplines of mathematics, physics, and chemistry.

We conduct extensive experiments on \data to evaluate the scientific reasoning abilities of 25 representative MLLMs. These models include close-source LLMs, close-source and open-source MLLMs, offering a comprehensive analysis of their performance across various disciplines. As illustrated in Figure~\ref{fig:data_performance}, the best performance is observed in close-source MLLMs, with distinct models excelling in different disciplines. In specific, Claude3.5-Sonnet achieves an accuracy of 53.4\% in mathematics, GPT-4o reaches a 38.2\% accuracy in physics, and Gemini-1.5-Pro records an accuracy of 47.0\% in chemistry. Among open-source models, InternVL-1.2-Plus performs best with accuracies of 30.1\% in mathematics, 24.8\% in physics, and 31.2\% in chemistry. Lastly, we systematically analyze the errors made by advanced models like GPT-4o on \data, which providing valuable insights into the specific domains where these models excel and where they struggle.

\section{\data Dataset}
In this section, we first illustrate the overview of our specially curated \data benchmark, designed to assess the capabilities of MLLMs in multi-modal scientific reasoning. Next, we introduce data generation process, which encompasses three core scientific disciplines: mathematics, physics, and chemistry. Lastly, we discuss the difference between our \data benchmark and existing benchmarks.

\subsection{Overview}
We introduce the \data benchmark, a meticulously curated collection aimed at evaluating the capabilities of multi-modal large language models (MLLMs) in multi-modal scientific reasoning, with a particular focus on bilingual tasks involving both English and Chinese. This dataset incorporates textual and visual contexts as inputs and spans three scientific disciplines, including mathematics, physics, and chemistry. Each discipline comprises 1,000 questions, meticulously gathered from different chapters to ensure comprehensive coverage of topics and concepts. The core statistics of the \data benchmark are presented in Table~\ref{tab:statistics}. The distributions of question length in \data are provided in Appendix~\ref{appendix:question_length}.  


In mathematics, the dataset can be divided into six key areas: plane geometry, solid geometry, functions and equations, algebraic operations, probability and statistics, and combinatorial mathematics. The physical component of the \data benchmark encompasses eight subjects, including mechanics, thermodynamics, comprehensive experiments and methods, mechanical motion, vibration and waves, optics, electromagnetism, and modern physics. The chemistry section of the \data benchmark includes seven topics such as chemical experiments, organic chemistry, substance composition, electrochemistry, chemical reactions, inorganic chemistry, and chemical equilibrium. A detailed introduction of each subjects within the three disciplines is available in Appendix~\ref{appendix:description_of_subjects}.

The primary objective of the \data benchmark is to provide a rigorous and diverse benchmark for assessing the multi-modal scientific reasoning capabilities of MLLMs. This benchmark aims to supplement existing benchmarks that predominantly focus on mathematical reasoning by broadening the scope to include expansive domains such as mathematics, physics, and chemistry. This benchmark aims to supplement existing benchmarks that predominantly focus on mathematical reasoning by broadening the scope to include expansive domains such as mathematics, physics, and chemistry. Through this enhancement, \data seeks to provide a more holistic measure of MLLMs' abilities across a wider spectrum of scientific disciplines.

\begin{table}[h!]
\centering
\resizebox{0.65\textwidth}{!}{%
\begin{tabular}{ll}
\toprule
Statistic & Number \\
\midrule
Total questions & 3000 \\
- multiple-choice questions & 2,053 (68.4\%) \\
- Free-form questions & 947 (31.6\%) \\
\midrule
Number of categories of math questions & 6 \\
Number of categories of physics questions & 8 \\
Number of categories of chemistry questions & 7 \\
Number of difficulty levels & 5 \\
\midrule
Unique number of images & 3,000 \\
Unique number of questions & 3,000 \\
Unique number of answers & 1,427 \\
\midrule
\multicolumn{2}{c}{\textit{Statistics with Chinese Language}} \\
\midrule
Maximum question length & 1297 \\
Maximum answer length & 112 \\
Maximum choice number & 5 \\
Average question length & 162.85 \\
Average answer length & 20.93 \\
\midrule
\multicolumn{2}{c}{\textit{Statistics with English Language}} \\
\midrule
Maximum question length & 418 \\
Maximum answer length & 92 \\
Maximum choice number & 5 \\
Average question length & 80.93 \\
Average answer length & 12.30 \\
\bottomrule
\end{tabular}}
\vspace{0.2cm}
\caption{Key statistics of \textsc{\data}.}
\label{tab:statistics} 
\vspace{-0.3cm}
\end{table}

\subsection{Data Generation}

\begin{figure}[t!]
    \centering
    \includegraphics[width=0.98\textwidth]{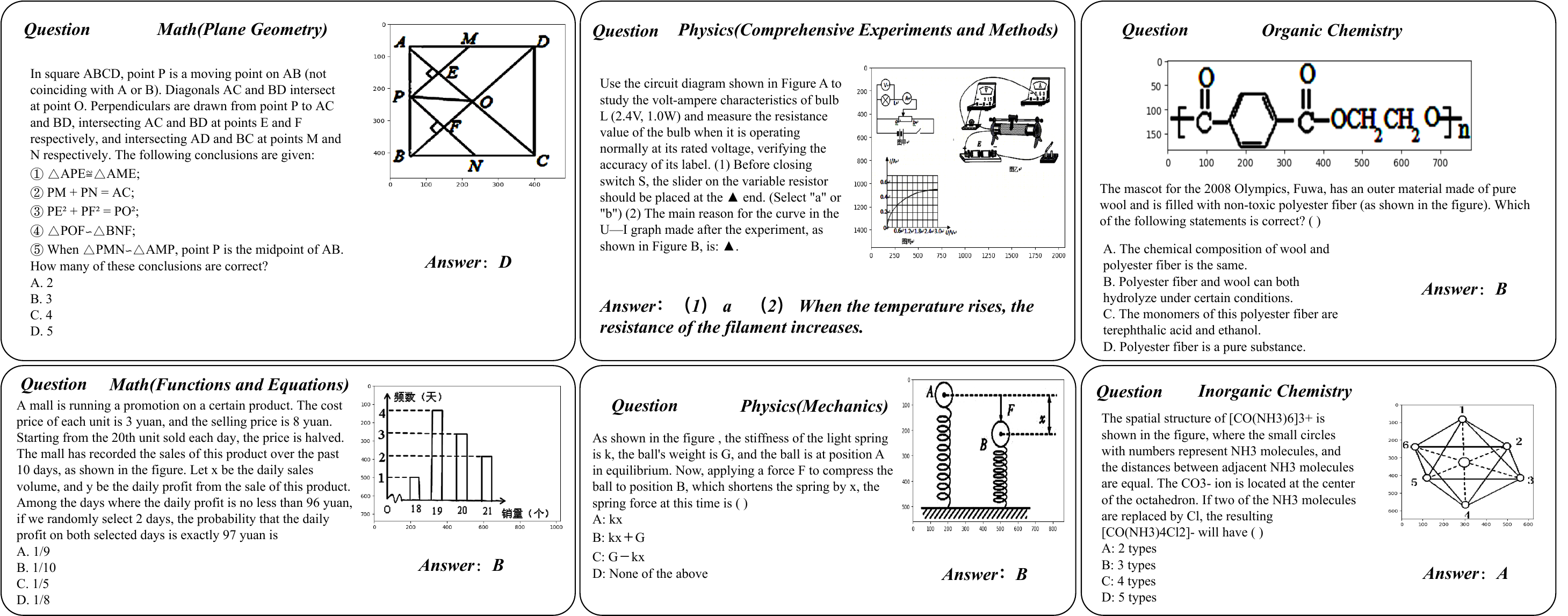}
    \caption{Examples of the \data benchmark comprising three disciplines: mathematics, physics, and chemistry.}
    \label{fig:data_case}
\end{figure}

The goal of the \data benchmark is to establish a comprehensive, bilingual (Chinese and English) benchmark for evaluating the capabilities of MLLMs in processing and understanding complex, scientifically-oriented tasks across various disciplines. In order to achieve this goal, we present a two-stage data generation pipeline to meticulously construct a benchmark dataset comprising 3,000 questions, evenly distributed with 1,000 questions each in the fields of mathematics, physics, and chemistry. Figure~\ref{fig:data_case} shows some examples sampled from the \data benchmark across three disciplines: mathematics, physics, and chemistry. More cases in \data are provided in Appendix~\ref{appendix: dataset_case}.

\vpara{Data Collection.} We gather a total of 450,000 questions from the disciplines of mathematics, physics, and chemistry, each enriched with visual information sourced from K12 education. This collection spans a comprehensive range of knowledge points across different chapters, with the difficulty levels scaled based on education grade. Consequently, we cluster 150,000 questions per discipline and carefully select 1,000 representative questions. These questions exemplify a range of difficulty levels and a variety of subjects, guided by the following principles:
\begin{itemize}
    \item \textit{Guaranteeing every knowledge point is included in \data benchmark.}

    \item \textit{Prioritizing the selection of questions from high-frequency knowledge points.}

    \item \textit{Ensuring a mixture of questions across various difficulty levels.}
\end{itemize}

In the end, the \data benchmark is constructed with 3,000 questions, with each of the three disciplines -- mathematics, physics, and chemistry -- contributing 1,000 questions. This approach ensures that the benchmark comprehensively covers a wide array of topics within each discipline, reflecting the breadth and depth required for a thorough assessment of MLLMs' capabilities.

\vpara{Data Annotation.} To improve the quality of the \data benchmark, we conduct multiple checks using both manual reviews and LLM assessments to confirm the completeness of each question. For textual content, we check for accuracy, coherence and relevance, ensuring that each question aligns with the corresponding scientific discipline and is free of ambiguities. For associated visual content, we rigorously screen out images that are incorrect, unclear, or lacking in detail, retaining only those that are clear and richly informative. To maintain the volume of the \data benchmark, we compensate for questions removed due to incomplete information by selecting new questions on identical topics from the original dataset. This approach ensures that the overall number of questions and the breadth of content coverage are consistently maintained. This verification process guarantees that both the textual and visual components of the \data benchmark is a reliable and effective tool for evaluating the capabilities of MLLMs in scientific reasoning.

\subsection{Data Analysis}

\begin{figure}[htbp]
\centering
\begin{minipage}{.50\textwidth}
  \centering
  \includegraphics[width=1\linewidth]{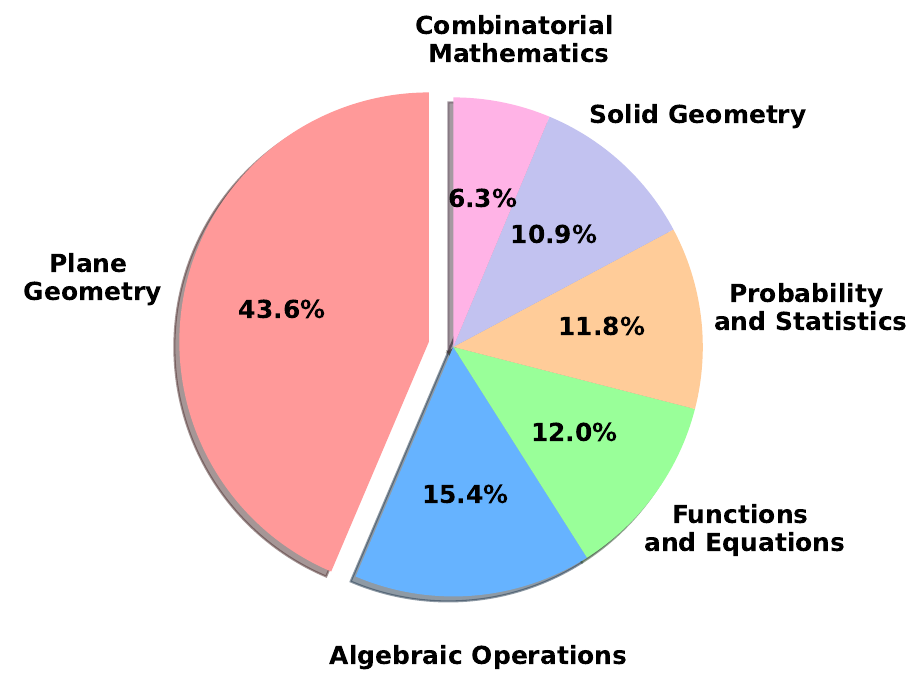}
\end{minipage}%
\begin{minipage}{.50\textwidth}
  \centering
  \includegraphics[width=1\linewidth]{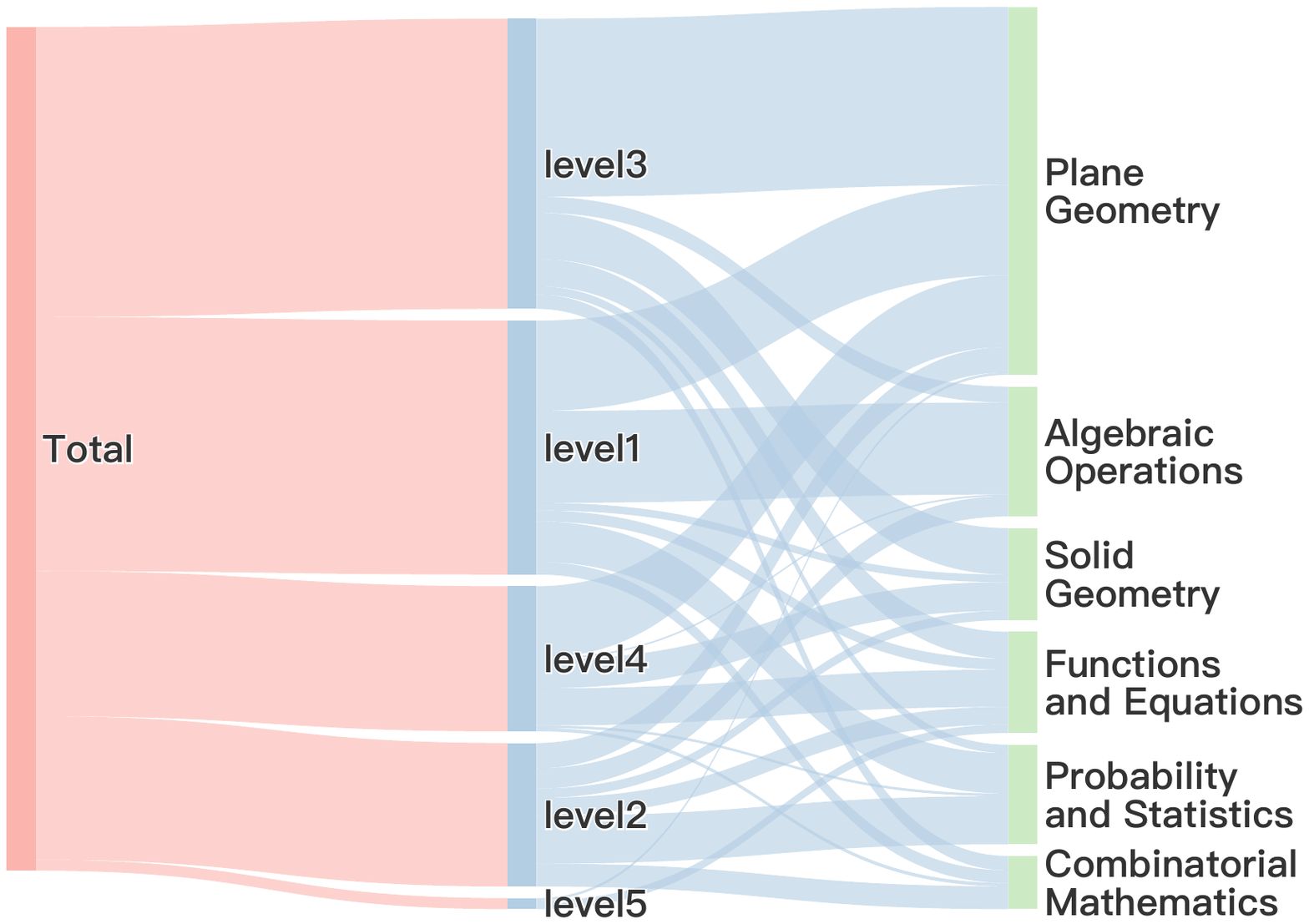}
\end{minipage}
(a) Mathematics
\label{fig:proportion_math}
\centering
\begin{minipage}{.50\textwidth}
  \centering
  \includegraphics[width=1\linewidth]{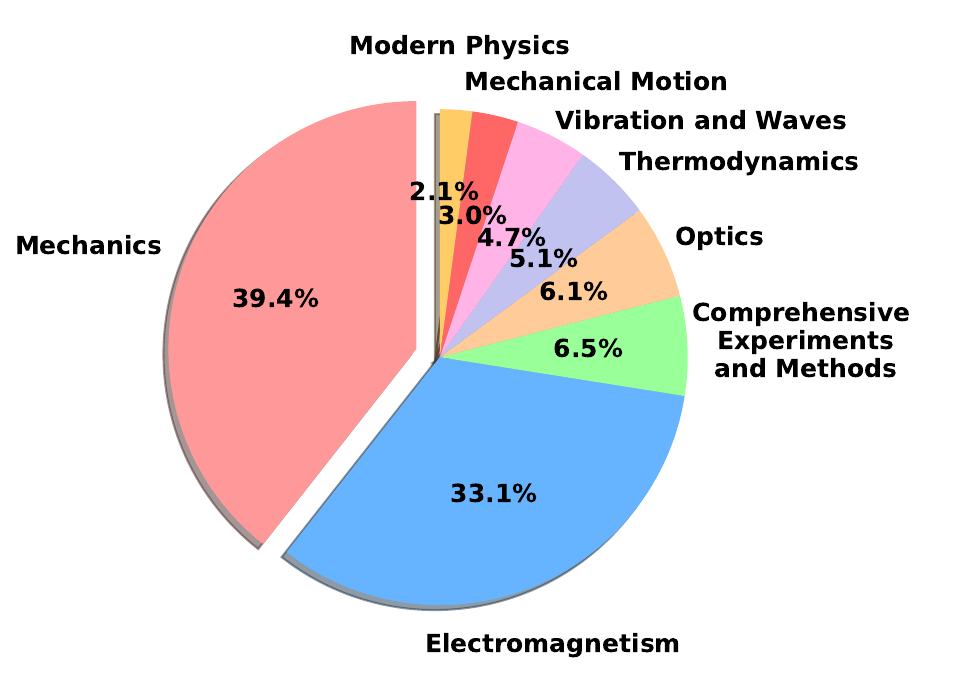}
\end{minipage}%
\begin{minipage}{.56\textwidth}
  \centering
  \includegraphics[width=1\linewidth]{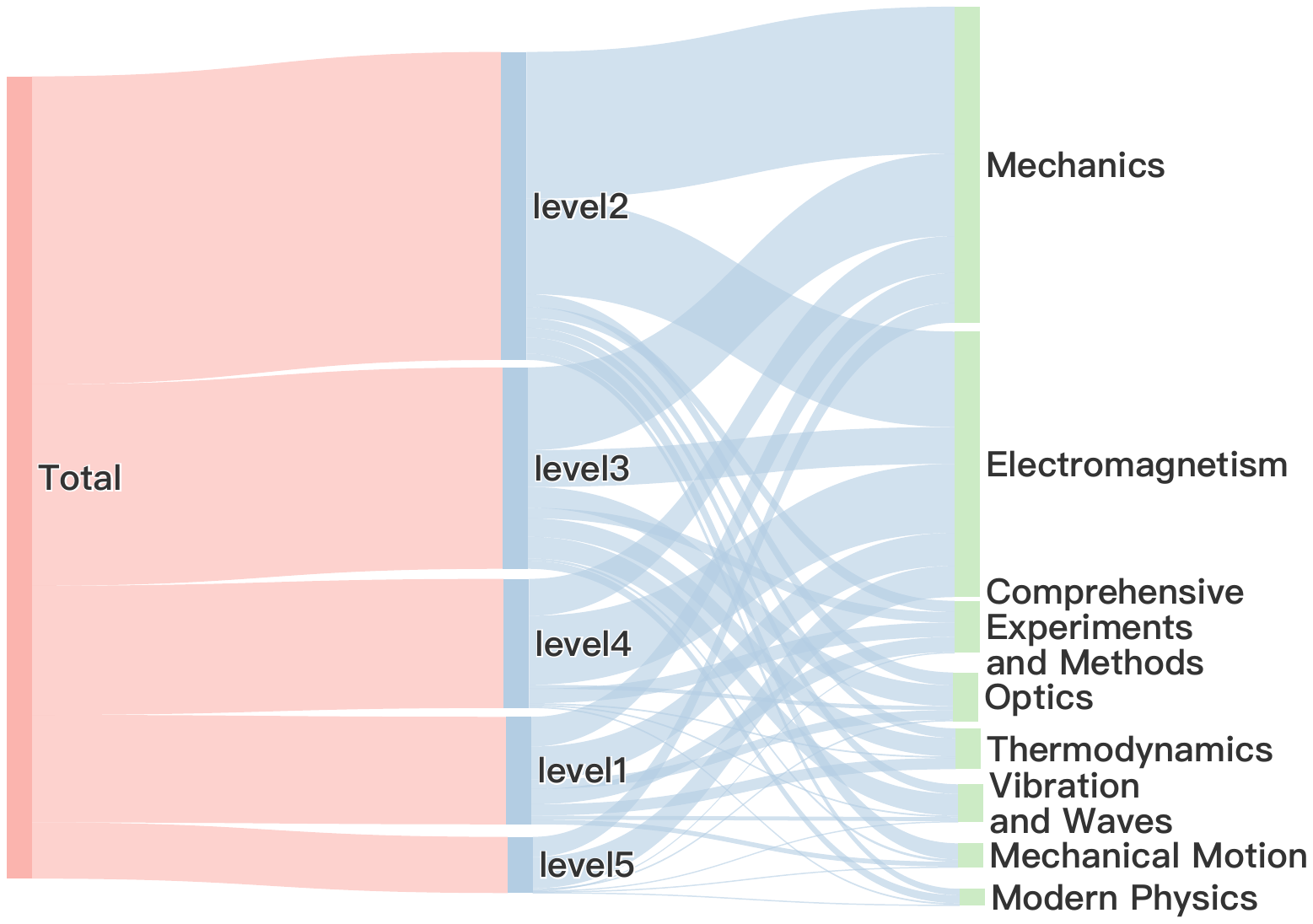}
\end{minipage}
(b) Physics
\label{fig:proportion_physics}
\centering
\begin{minipage}{.50\textwidth}
  \centering
  \includegraphics[width=1\linewidth]{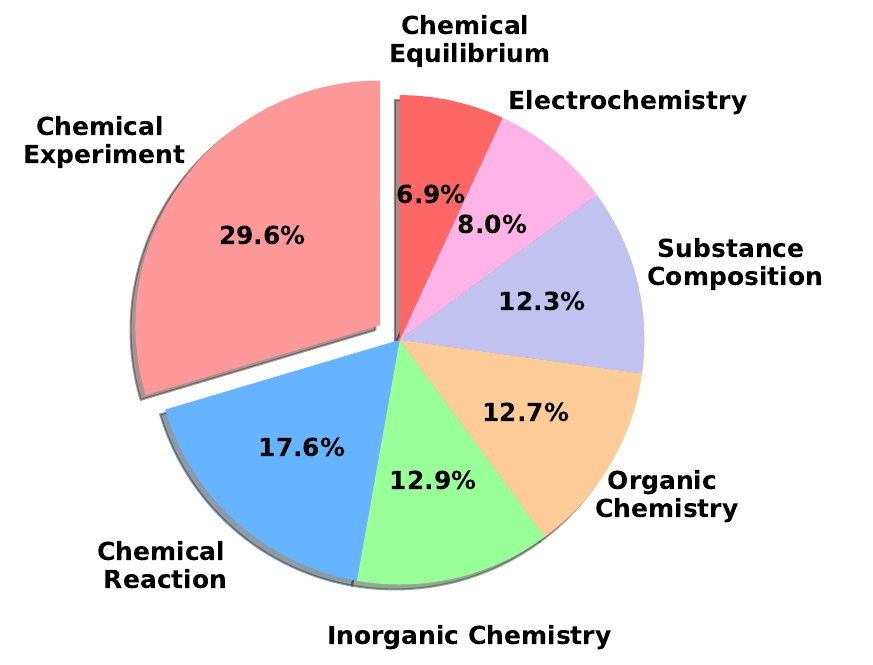}
\end{minipage}%
\begin{minipage}{.50\textwidth}
  \centering
  \includegraphics[width=1\linewidth]{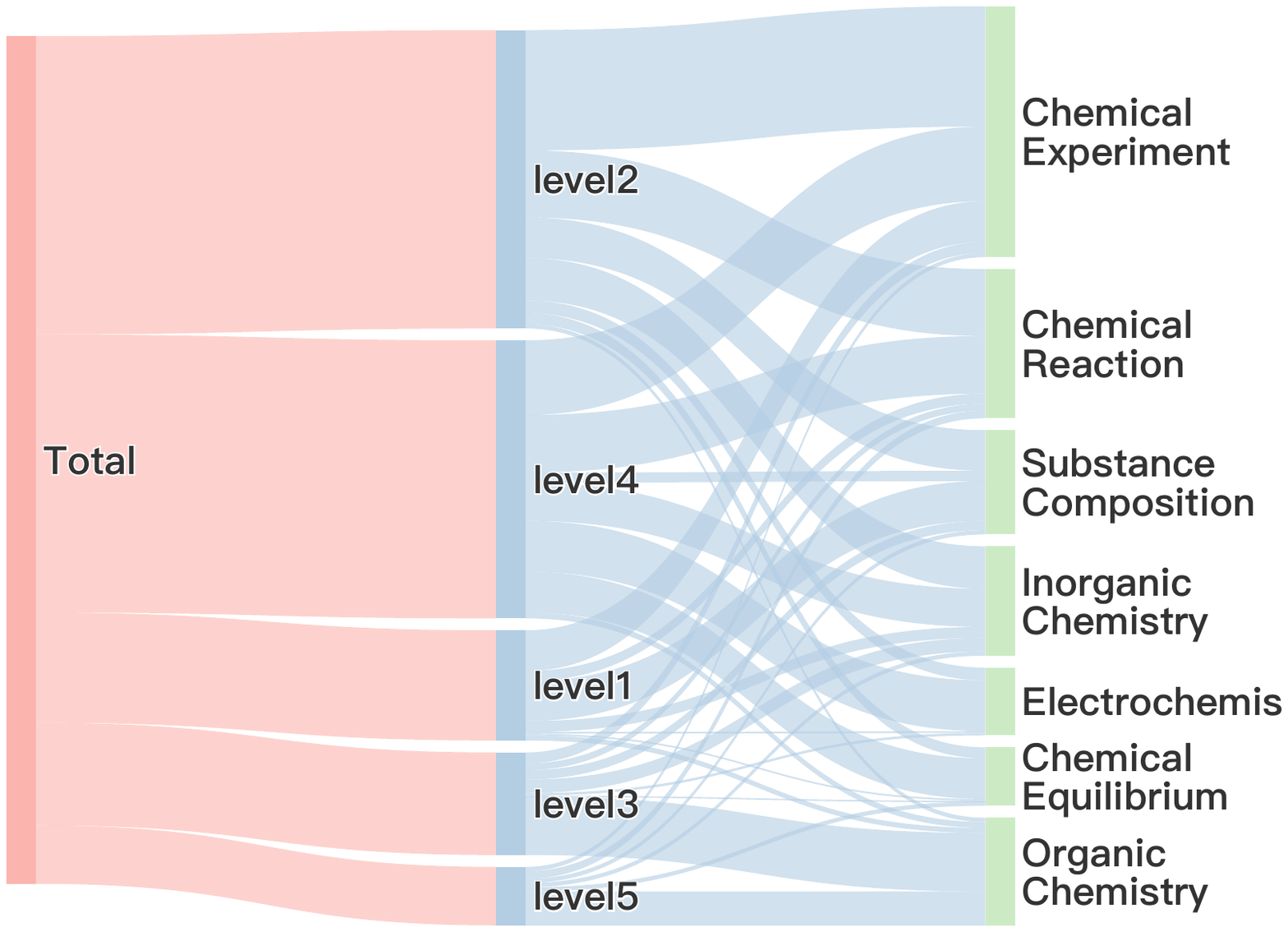}
\end{minipage}
(c) Chemistry
\caption{The distribution of detailed subjects and difficulty levels in the each discipline within the \data benchmark. (Left) The distributions of various subjects. (Right) The distributions of difficulty levels.}
\label{fig:benchmark_category}
\end{figure}

We utilize statistical analysis to assess subject distributions and difficulty levels within the \data benchmark. Figure~\ref{fig:benchmark_category} presents a visual representation of the categorization of question within the \data benchmark. This illustration shows the distribution of questions dedicated to each subject area -- mathematics, physics, and chemistry -- and details the distribution across various difficulty levels, ranging from 1 to 5.

\vpara{Subject Distributions.} To categorize each discipline into more detailed subjects, we first utilize LLM to segment the overall discipline into specific topics based on knowledge points and terminologies presented in the questions. Subsequently, we conduct a manual review of these categories to confirm its rationality and appropriateness, ensuring that each question is accurately categorized. As shown in Figure~\ref{fig:benchmark_category}, the mathematical part of the \data benchmark is divided into six subjects, i.e., plane geometry (43.6\%), algebraic operations (15.4\%), functions and equations (12\%), probability and statistics (11.8\%), solid geometry (10.9\%), and combinatorial mathematics (6.3\%). Furthermore, the distributions for physics and chemistry disciplines are presented in the figure, providing a comprehensive overview of the scope of the \data benchmark within these scientific fields.

\vpara{Difficulty Levels.} To classify the questions into distinct difficulty levels, we first utilize LLM for the initial sorting, and then conduct a manual verification. The questions within each discipline are stratified into five difficulty levels ranging from 1 to 5, defined as follows: Basic, Easy, Intermediate, Advanced, and Expert. Figure~\ref{fig:benchmark_category} shows the distribution of difficulty levels, providing a visual representation of the distribution of questions across different difficulty levels. Each discipline demonstrates a unique profile of topic distribution across the difficulty levels. For instance, in the field of mathematics, \textit{plane geometry} is classified at the intermediate level, \textit{algebraic operations} are positioned at the basic level, and \textit{functions and equations} appears at the highest difficulty level, reflecting their various placement within educational curricula. In physics, \textit{mechanics} dominates the introductory level, which belongs to a fundamental concept in physics education. \textit{Electromagnet} is positioned at the highest difficulty level, demanding the application of various advanced knowledge points. In the discipline of chemistry, \textit{organic chemistry} and \textit{chemical equilibrium} represent the pinnacle of K12 chemical education, requiring deep conceptual understanding and the ability to apply knowledge to complex scenarios.

\subsection{Comparison with Other Benchmarks}
We compare the \data benchmark with 5 existing benchmarks, including MathVista~\cite{lu2023mathvista}, Math-Vision~\cite{wang2024measuring}, CMMMU~\cite{zhang2024cmmmu}, ScienceQA~\cite{lu2022learn}, and SciBench~\cite{wang2023scibench}.

\vpara{\data vs MathVista.} MathVista is a comprehensive multi-modal benchmark for mathematical reasoning, comprising data from 28 existing datasets and 3 newly collected datasets. In MathVista, the majority of questions are annotated after collecting images, which results in a certain homogeneity within the data. In contrast, \data directly collects its questions from K12 education, featuring an average question length of 80.93 words. Such questions provide more contextual information, which  facilitate a more thorough evaluation of the models' reasoning capabilities. Unlike MathVista that encompasses only seven subjects within mathematics, \data offers a far broader scope, including 22 distinct subjects across mathematics, physics, and chemistry. Furthermore, \data distinguishes itself by being a bilingual benchmark, including both Chinese and English versions of questions. This feature is particularly advantageous as it assesses MLLMs' capabilities in scientific reasoning across different languages.

\vpara{\data vs Math-Vision.} Math-Vision is a mathematics benchmark derived from 19 competitions, covering 16 topics across 5 levels of difficulty. Different from Math-Vision that collected from competitions, \data spans a broader educational spectrum, incorporating a natural gradient of difficulty from elementary school to high school. Furthermore, \data extends beyond mathematics to include questions from physics and chemistry, significantly broadening its scope and applicability. While Math-Vision primarily focuses on the unique challenges of competitive environments, \data is grounded in real-world educational settings.

\vpara{\data vs CMMMU.} CMMMU comprises 12,000 manually collected multi-modal questions from university exams, quizzes, and textbooks, which covers 6 core subjects and 30 specific fields. Similar to \data, CMMMU is a bilingual benchmark, offering questions in both Chinese and English. Within this dataset, only 1,601 questions are dedicated to the disciplines of mathematics, physics, and chemistry, accounting for only 13.34\% of the total dataset. \data features a total of 3,000 questions, significantly outnumbering those in CMMMU dedicated to the same subjects. The questions in CMMMU are set at the university level, characterized by high difficulty, demanding that the model possesses substantial professional domain knowledge and expert-level reasoning abilities. In contrast, \data comes from K12 education, with a broader range of difficulty. This range allows \data to more comprehensively evaluate MLLMs' capabilities across different educational stages.

\vpara{\data vs ScienceQA.} ScienceQA is a newly developed benchmark featuring approximately 21,000 multimodal multiple-choice questions across a variety of science topics. In the ScienceQA dataset, 30.8\% of questions incorporate both image and text contexts, providing a multimodal benchmark to test MLLMs in scientific reasoning. The questions in ScienceQA have an average length of only 12.11 words. In contrast, \data also serves as a benchmark for evaluating the scientific reasoning abilities of MLLMs, but it typically features longer and more textually detailed questions. Specifically, the Chinese version of \data has an average question length of 162.85 words, providing a more comprehensive and intricate testing ground for evaluating the depth of detailed reasoning in MLLMs. Additionally, \data contains mathematical problems, further enriching the benchmark's scope by testing MLLMs on their mathematical problem solving capabilities alongside their scientific reasoning.

\vpara{\data vs SciBench.} SciBench is a benchmark developed to evaluate the reasoning capabilities of LLMs in solving collegiate-level scientific problems within the domains of mathematics, chemistry, and physics. The majority of the data in SciBench focuses on assessing the scientific reasoning of LLMs, it only includes 177 problems that incorporate visual elements to evaluate the performance of MLLMs. In contrast, \data is primarily focused on multimodal scientific reasoning, covering similar subjects such as mathematics, chemistry, and physics. \data differentiates itself by offering a more comprehensive range of difficulty levels and subjects, making it a broader benchmark for assessing the capabilities of MLLMs in scientific reasoning.

\section{Experiments}
In this section, we conduct experiments to evaluate a variety of MLLMs using the \data benchmark. The evaluation encompasses both close-source and open-source models, enabling a comprehensive analysis of their effectiveness in scientific reasoning. Besides, we provide a detailed error analysis of the advanced model GPT-4o.  

\subsection{Experimental Setup}

\vpara{Models.} We conduct our evaluation across a diverse array of models, including close-source text-only large language models (LLMs), close-source multi-modal large language models (MLLMs), and open-source MLLMs. This comprehensive assessment covers more than 20 models, which are listed below. The sources of models is reported in Appendix~\ref{appendix:model_sources}. 
\begin{itemize}
    \item Close-source text-only LLMs: ChatGPT~\cite{openai2022chatgpt}, GPT-4~\cite{achiam2023gpt}, Claude2~\cite{claude2}.

    \item Close-source MLLMs: Gemini-1.0-Pro~\cite{team2023gemini}, Gemini-1.5-Pro~\cite{team2023gemini}, GPT-4o~\cite{openai2024gpt4o}, Qwen-VL-Max~\cite{bai2023qwen-vl}, Qwen-VL-Plus~\cite{bai2023qwen-vl}, Claude3.5-Sonnet~\cite{claude3.5}, Claude3-Opus~\cite{Claude3}, GLM-4V~\cite{glm4v}, and Step-1V~\cite{step-1v}.
    
    \item Open-source MLLMs: mPLUG-Owl~\cite{ye2024mplug}, LLaMA-Adapter-V2~\cite{gao2023llama}, MiniCPM-Llama3-V2.5~\cite{hu2024minicpm}, LLaVA-1.5~\cite{liu2024llava}, DeepSeek-VL~\cite{lu2024deepseek}, ShareGPT4V~\cite{chen2023sharegpt4v}, SPHINX-Plus~\cite{gao2024sphinx}, InternLM-XC2~\cite{dong2024internlm}, InternVL-1.2-Plus~\cite{chen2023internvl}, InternVL-Chat-V1.5~\cite{chen2024internvl}, CogVLM~\cite{wang2023cogvlm}, CogVLM2~\cite{wang2023cogvlm}, and GLM-4V-9B~\cite{glm2024chatglm}. 
\end{itemize}


\vpara{Evaluation Details.} The evaluation process is conducted through two steps: generation and judgment. During the generation phase, the models are tasked with producing responses based on a set of questions. For zero-shot setting, we directly prompt the models with these questions without any examples. For 2-shot Chain of Thought (CoT) setting, we provide the models with two relevant examples before they are prompted with the questions. For MLLMs, we supply the models with the textual questions and the corresponding image to obtain their responses. During the judgment phase, we utilize GPT-4o to evaluate the models' responses by comparing them with the standard answers to assess  consistency. This phase involves calculating the accuracy across different subjects and levels. The prompts used in two phases is defined in Appendix ~\ref{appendix:prompts}.

\subsection{Experimental Results}

\vpara{Overall Results.} Table~\ref{tab:main_results_zh} demonstrates the performance of several models on \data within the version of the Chinese language. Experimental results show that the close-source models achieves best performance on \data. Specifically, Claude3.5-Sonnet achieves an accuracy of 53.4\% in mathematics, GPT-4o attains a 38.2\% accuracy in physics, and Gemini-1.5-Pro accomplishes an accuracy of 47.0\% in chemistry. Among open-source models, InternVL-1.2-Plus stands out, demonstrating robust capabilities across various scientific disciplines with accuracies of 30.1\% in mathematics, 24.8\% in physics, and 31.2\% in chemistry. Despite this, there is a notable disparity in performance between close-source and open-source models, with close-source models generally exhibiting superior performance. The performance of InternVL-1.2-Plus, although trailing behind the advanced close-source models such as GPT-4o, Claude3.5-Sonnet, and Gemini-1.5-Pro, showing significant potential for improvement. Notably, the performance in physics underscores unique challenges that necessitate targeted improvements in model training. This discipline often involves the interpretation of conceptual and numerical data, challenging the reasoning and computational abilities of MLLMs. As evidenced in Table~\ref{tab:main_results_zh}, even advanced models like GPT-4o achieve relatively lower accuracies in physics compared to other disciplines. Results on \data with the version of the English language are provided in Appendix~\ref{appendix:results_english}. 

\begin{table*}[t!]
    \centering
    \renewcommand{\arraystretch}{1.15}
    \small
    \resizebox{\textwidth}{!}{
    \begin{tabular}{cccccc}  
    \toprule
     Model & LLM& Input & Mathematics  & Physics & Chemistry  \\ 
    \midrule
    \multicolumn{6}{c}{\textit{Close Source Models} (APIs)}\\
    \midrule
    \multicolumn{6}{l}{\textit{Text-only LLMs}}\\
     Zero-shot ChatGPT & - & \textit{Q} & 22.4  & 22.7 &  18.6  \\
     Zero-shot GPT-4 & - & \textit{Q} & 25.9  & 30.4 &  33.1 \\ 
     Zero-shot Claude-2 &-& \textit{Q} &  27.3  & 22.0 & 24.4 \\ 
     Zero-shot Claude3-Opus &-& \textit{Q} & 29.3   & 30.8 &  32.5  \\
     Zero-shot Claude3.5-Sonnet &-& \textit{Q} & 29.7   & 35.3 &  36.9  \\
     Zero-shot GPT-4o &-& \textit{Q} &  31.1  & 38.0 & 39.6 \\   
     2-shot CoT Claude2 &-& \textit{Q} & 27.8  & 21.7 &  23.9  \\
     2-shot CoT ChatGPT &-& \textit{Q} & 20.2  & 18.6 & 21.3  \\ 
     2-shot CoT GPT-4 &-& \textit{Q} &  32.1  & 31.5 &  32.4 \\ 
     \midrule
     \multicolumn{5}{l}{\textit{Multi-modal LLMs}}\\
     Gemini-1.0-Pro & - &    \textit{Q}, \textit{I} & 26.6 & 23.70 & 27.8  \\
     Gemini-1.5-Pro & -  &    \textit{Q}, \textit{I} & 49.4  & 38.1 &  \cellcolor{red!25}{47.0} \\  
     GPT-4o & - & \textit{Q}, \textit{I} & 51.7  & \cellcolor{red!25}{38.2} &  41.6  \\
     GPT-4o-mini & - & \textit{Q}, \textit{I} & 42.6  & 29.8 &  28.4 \\
     Qwen-VL-Max & - & \textit{Q}, \textit{I} & 35.5 & 30.70&42.5\\ 
     Qwen-VL-Plus & - & \textit{Q}, \textit{I} & 27.6 & 26.5 & 37.7  \\
     Claude3.5-Sonnet & - & \textit{Q}, \textit{I} & \cellcolor{red!25}{53.4}  & 38.0 &  43.1  \\
     Claude-3 opus & - & \textit{Q}, \textit{I} & 34.4  & 31.1 &  34.1  \\
     GLM-4V & - & \textit{Q}, \textit{I} & 24.2 & 19.2 & 25.0 \\
     Step-1V  & - & \textit{Q}, \textit{I} & 28.1 & 23.5& 25.0\\
     \midrule
    \multicolumn{5}{c}{\textit{Open Source Models}}\\
    \midrule
    \multicolumn{5}{l}{\textit{General Multi-modal LLMs}}\\
    mPLUG-Owl   & LLaMA-7B  & \textit{Q}, \textit{I} & 7.6 & 8.3& 9.5\\
    LLaMA-Adapter-V2   & LLaMA-7B & \textit{Q}, \textit{I} & 9.6 & 10.3&10.8\\
    MiniCPM-Llama3-V2.5   & LLaMA3-8B & \textit{Q}, \textit{I} & 15.4 &17.9 & 19.5\\
    LLaVA-1.5   & Vicuna-13B & \textit{Q}, \textit{I} & 15.5 & 15.2& 18.8\\
    LLaVA-1.5    & Vicuna-7B & \textit{Q}, \textit{I} & 13.0 & 13.5 & 16.0\\
    DeepSeek-VL  & DeepSeek-LLM-7B  & \textit{Q}, \textit{I} & 8.3   &16.8 & 21.0\\
    ShareGPT4V  & Vicuna-7B & \textit{Q}, \textit{I} & 15.7 & 14.0& 19.0\\
    ShareGPT4V  & Vicuna-13B & \textit{Q}, \textit{I} & 16.4 & 14.9& 18.4\\
    SPHINX-Plus  & LLaMA2-13B & \textit{Q}, \textit{I} & 17.0 &15.3 & 20.4\\
    InternLM-XC2   & InternLM2-7B & \textit{Q}, \textit{I} & 24.9 &18.3 & 25.6 \\
    InternVL-1.2-Plus  & 
Nous-Hermes-2-Yi-34B & \textit{Q}, \textit{I} & \cellcolor{blue!25}{30.1}  & \cellcolor{blue!25}{24.8} & \cellcolor{blue!25}{31.2} \\
    InternVL-Chat-V1.5  & Mixtral 8*7B & \textit{Q}, \textit{I} & 26.9 & 20.8& 23.7\\
    CogVLM   & Vicuna-7B & \textit{Q}, \textit{I} & 16.7 & 14.5& 17.0\\
    CogVLM2   & LLaMA-3-8B & \textit{Q}, \textit{I} & 23.2 & 14.4 & 21.0\\
    GLM-4V-9B  &  GLM-4-9B & \textit{Q}, \textit{I} & 24.7 & 19.3& 22.5\\
    \bottomrule
    \end{tabular}} 
    \caption{\textbf{Results on \data within the version of the Chinese language across the disciplines of mathematics, physics, and chemistry.} For input, $Q$ represents for question, $I$ represents for image. The highest scores among close-source and open-source models are highlighted in \textcolor{red}{red} and \textcolor{blue}{blue}, respectively.}
    \label{tab:main_results_zh}
    \vspace{-0.4cm}
\end{table*}

\vpara{Results on Mathematics Across Different Subjects.} The mathematical part of \data encompasses a wide range of subjects, including plane geometry, solid geometry, functions and equations, algebraic operations, probability and statistics, and combinatorial mathematics. Table~\ref{tab:results_on_each} reports the comprehensive results across different mathematical subjects. It is evident that models like Claude3.5-Sonnet and GPT-4o in close-source MLLMs excel across multiple subjects, particularly in \textit{functions and equations}, \textit{probability and statistics}, and \textit{algebraic operations}. Conversely, open-source models show a more varied performance with notable strengths in certain areas but generally lower scores compared to close-source models. For instance, InternVL-1.2-Plus and InternVL-Chat-V1.5 perform relatively well in \textit{plane geometry}, and \textit{functions and equations}. These detailed performance on different subjects provide valuable insights into the specific strengths and weaknesses of various MLLMs. Additionally, results on physics and chemistry across different subjects are presented in Appendix~\ref{appendix:physics_subjects} and Appendix~\ref{appendix:chemistry_subjects}, respectively. Case studies illustrating correct responses by MLLMs can be found in Appendix~\ref{appendix: case_study}.

\begin{table*}[t!]
    \centering
    \renewcommand{\arraystretch}{1.15}
    \small
    \resizebox{\textwidth}{!}{
    \begin{tabular}{cccccccc}  
    \toprule
     \multirow{2}{*}{Model} & \multicolumn{6}{c}{Mathematics} \\
     & ALL &  PlaneG & SolidG & Fun & Alg &  Stat &  Comb \\
    \midrule
    \multicolumn{8}{c}{\textit{Close Source Models} (APIs)}\\
    \midrule
    \multicolumn{8}{l}{\textit{Text-only LLMs}}\\
     Zero-shot ChatGPT  & 22.40 & 20.18  & 11.93 &  18.33& 13.63 & 15.25 & 26.98  \\
     Zero-shot GPT-4  & 25.90 &  30.73 &18.35 & 28.33  & 17.53 & 24.58 & 33.33 \\ 
     Zero-shot Claude-2  & 27.30 & 27.06& 25.69 &  25.83  & 31.17 & 31.36 & 25.40 \\ 
     Zero-shot Claude3-Opus  & 29.30 & 30.28& 21.10 & 32.50   & 27.27 & 34.75 & 31.75  \\
     Zero-shot Claude3.5-Sonnet  & 29.70 & 33.94& 15.60 & 33.33  & 27.27 & 27.12 & 34.92  \\
     Zero-shot GPT-4o  &31.10 & 36.24 & 24.77 & 35.83  & 25.32 & 24.58 & 31.75 \\   
     2-shot CoT Claude2  & 27.80 & 30.05& 26.61 & 25.00  & 28.57 & 27.97 &  26.98  \\
     2-shot CoT ChatGPT  & 20.20 & 23.17& 20.18 & 19.17  & 17.53 & 22.88 &14.29  \\ 
     2-shot CoT GPT-4  & 32.10 & 37.16& 31.19 & 28.33  & 22.08 &  30.51 & 38.10 \\ 
     \midrule
     \multicolumn{8}{l}{\textit{Multi-modal LLMs}}\\
     Gemini-1.0-Pro  & 26.60 & 24.08 &   22.02 &23.73 & 35.71 & 29.66 & 34.92  \\
     Gemini-1.5-Pro  & 49.40 & 48.74  &  33.03 & 47.06  & 61.69 &  55.93 &  \cellcolor{red!25}{52.38} \\  
     GPT-4o  & 51.70 & 48.17 &  \cellcolor{red!25}{44.04} & \cellcolor{red!25}{57.50}  & 68.18 &  56.78 & 41.27  \\
     GPT-4o-mini  & 42.60 & 41.28  & 29.36 &  44.17 & 54.55 & 44.92 & 38.10 \\
     Qwen-VL-Max  & 35.50 & 34.86 & 27.52 & 35.83 & 50.00 &33.05 & 26.98\\ 
     Qwen-VL-Plus  & 27.60 & 27.98 & 18.35 & 29.17 & 31.17 & 35.59 & 20.63  \\
     Claude3.5-Sonnet  & \cellcolor{red!25}{53.4} & \cellcolor{red!25}{50.23} & 35.78 & \cellcolor{red!25}{57.50} & \cellcolor{red!25}{74.03} & \cellcolor{red!25}{63.56} & 39.68 \\
     Claude3-Opus  &  34.40& 35.31 &24.77 & 29.17& 45.45& 35.59 & 31.75  \\
     GLM-4V  & 24.20 & 28.57 & 30.28 & 22.50 & 20.26  & 21.37 & 17.46  \\
     Step-1V   & 28.10 & 31.68 &24.71 & 23.15&48.85 &40.57 &22.64 \\
     \midrule
    \multicolumn{8}{c}{\textit{Open Source Models}}\\
    \midrule
    \multicolumn{8}{l}{\textit{General Multi-modal LLMs}}\\
    mPLUG-Owl   & 7.60 & 6.19 &10.09 & 5.00& 12.34&7.63 &7.94 \\
    LLaMA-Adapter-V2    & 9.60 & 10.78 &10.09 &7.50 & 9.09&13.56 & 4.76\\
    MiniCPM-Llama3-V2.5    & 15.40 & 23.62 & 19.27& 15.83& 26.62& 26.27& 15.87\\
    LLaVA-1.5-13B   & 15.50 & 15.83 &15.60 & 12.50&18.83 & 14.41&14.29 \\
    LLaVA-1.5-7B    & 13.00 & 12.84  & 12.84 & 15.83 &  14.29& 11.86& 11.11\\
    DeepSeek-VL   & 8.30 &  13.99&8.26 &10.00 & 11.04& 10.17&7.94  \\
    ShareGPT4V-7B  & 15.70 & 16.06& 16.51 & 13.33 & 14.29 &  17.80 & 17.46 \\
    ShareGPT4V-13B   & 16.40 & 15.60& 11.93 &19.17  & 17.53 &22.03 & 14.29  \\
    SPHINX-Plus   & 17.00 & 21.79 & 19.27 & 15.83&20.13  & 22.88& 7.94 \\
    InternLM-XC2   & 24.90& 25.92 & 22.02& 22.50 & 27.92 & 27.97 & 20.63 \\
    InternVL-1.2-Plus  & \cellcolor{blue!25}{30.10} &  \cellcolor{blue!25}{34.40} & \cellcolor{blue!25}{25.69} & \cellcolor{blue!25}{30.00} & \cellcolor{blue!25}{29.87} & 26.27 & 23.81 \\
    InternVL-Chat-V1.5  &26.90 & 28.44 & \cellcolor{blue!25}{25.69} & 23.33 & \cellcolor{blue!25}{29.87} & 24.58 &  \cellcolor{blue!25}{26.98} \\
    CogVLM    &16.70 & 16.06 & 23.85 & 17.50 &17.53  & 13.56& 19.05\\
    CogVLM2   & 23.20 & 21.56 & 22.02& 29.17&22.73 & 26.27&20.63 \\
    GLM-4V-9B  & 14.70 & 25.23 & 20.18& 19.17& 27.27& \cellcolor{blue!25}{33.05} & 19.05 \\
    \bottomrule
    \end{tabular}} 
    \caption{\textbf{Results on the mathematical part of \data across different subjects.} Subjects: PlaneG: plane geometry, SolidG: solid geometry, Fun: functions and equations, Alg: algebraic operations, Stat: probability and statistics, Comb: combinatorial mathematics. The highest scores among close-source and open-source models are highlighted in \textcolor{red}{red} and \textcolor{blue}{blue}, respectively.}
    \label{tab:results_on_each}
\end{table*}

\subsection{Error Analysis}

\begin{figure}
    \centering
    \begin{minipage}{0.6\textwidth}
        \centerline{\includegraphics[width=\linewidth]{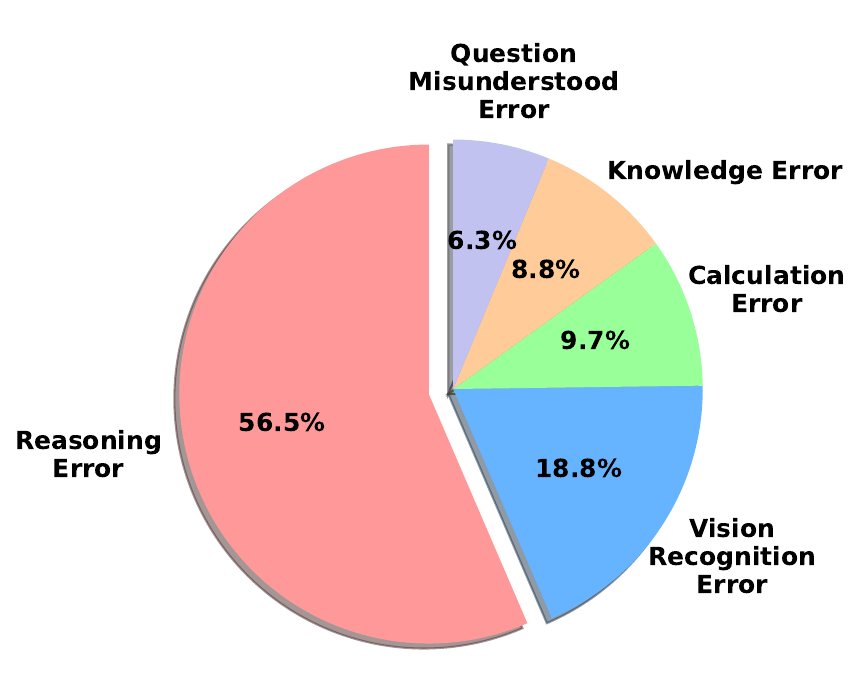}}
        \centerline{(a) Mathematics}
        \label{fig:math_error_distribution}
        \vspace{0.3cm}
        \centerline{\includegraphics[width=\linewidth]{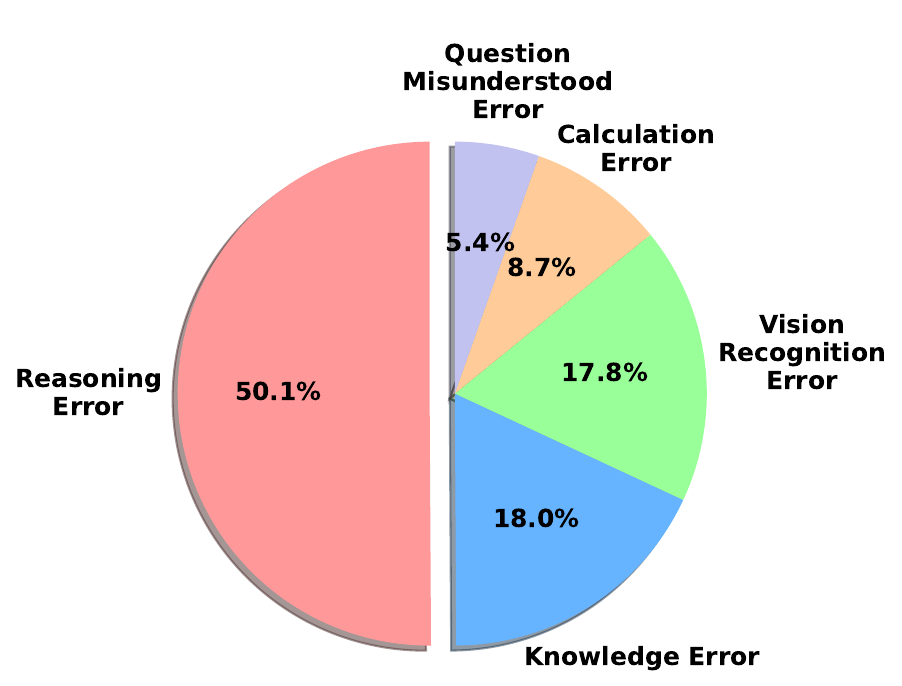}}
        \centerline{(b) Physics}
        \label{fig:physics_error_distribution}
        \vspace{0.3cm}
        \centerline{\includegraphics[width=\linewidth]{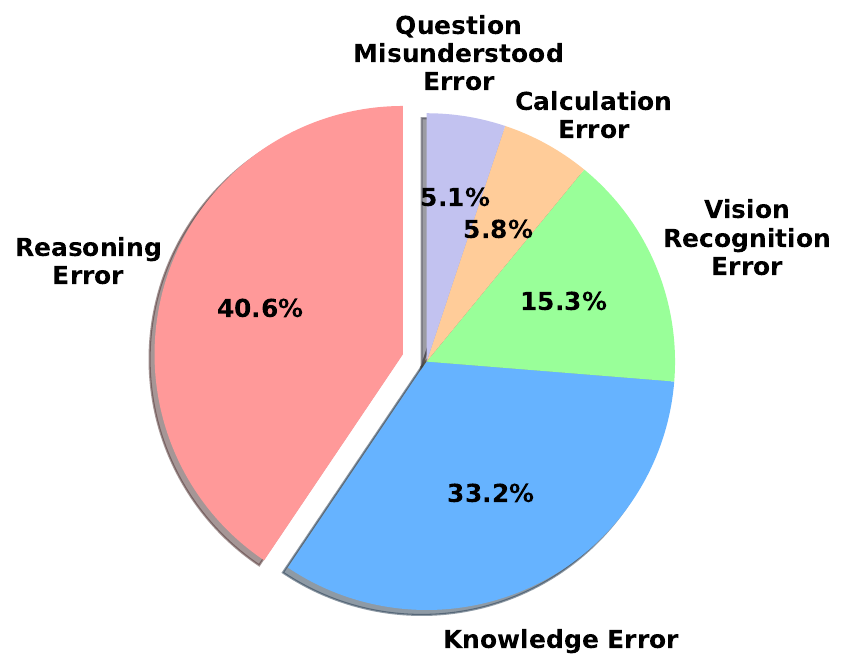}}
        \centerline{(c) Chemistry}
        \label{fig:chemistry_error_distribution}
    \end{minipage}
    \caption{Error distributions of GPT-4o on \data across the disciplines of mathematics, physics, and chemistry.}
    \label{fig:error_dist}
\end{figure}

\begin{figure}[hbpt]
    \centering
    \includegraphics[width=\textwidth]{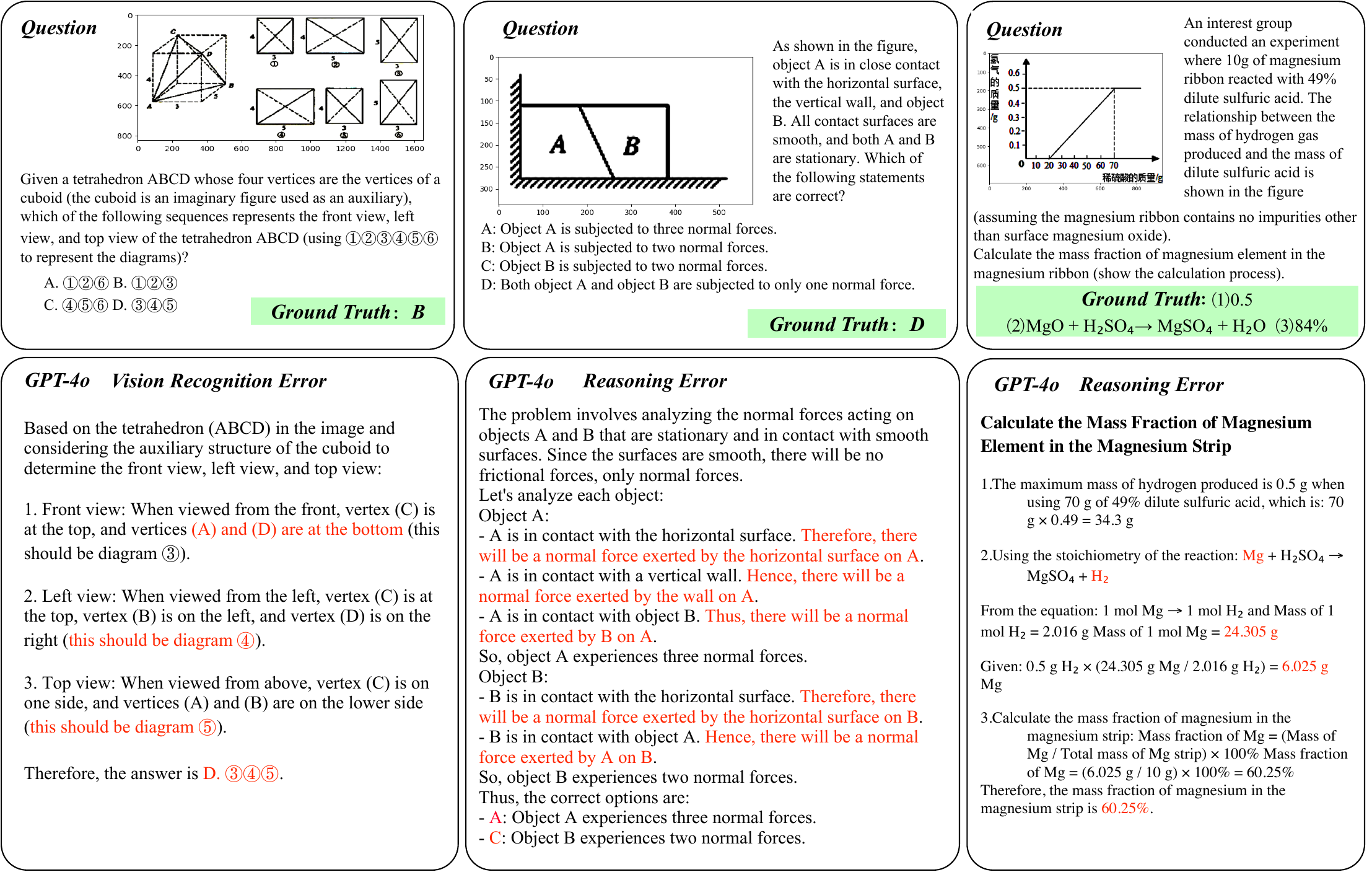}
    \vspace{-2mm}
    \caption{Cases of errors from GPT-4o in the disciplines of mathematics, physics, and chemistry.}
    \label{fig:error_case_all}
    \vspace{-3mm}
\end{figure}

To analyze the causes of errors in model responses, we meticulously review incorrect answers to identify common patterns. We specifically focus on the advanced MLLM, GPT-4o, to  illustrate specific instances of errors and their distributions across the disciplines of mathematics, physics, and chemistry. Figure~\ref{fig:error_dist} demonstrates the distributions of these errors, categorizing them into several types such as reasoning error, knowledge error, calculation error, vision recognition error, and question misunderstood error. Notably, across all disciplines, reasoning errors are the most prevalent, indicating a challenge in model's ability to solve scientific problems that involve visual information. Specifically, reasoning errors account for 56.5\% of the total errors in mathematics, 50.1\% in physics, and 40.6\% in chemistry, respectively. This is followed by knowledge error, which is particularly significant in chemistry, constituting 33.2\% of the errors in that discipline. Similarly, knowledge error also represent the second most common error type in physics. However, knowledge error in mathematics is less prevalent, making up only 8.8\% of the total errors. This indicates that while the model struggle with conceptual and fundamental principles in chemistry and physics, it demonstrate a better grasp of mathematical concepts. Vision recognition error is another significant type of error, accounting for 18.8\% of the errors in mathematics, making it the second most prevalent error type in this discipline. This error category is also significant in physics and chemistry, where it constitutes 17.8\% and 15.3\% of the errors, respectively. This type of error highlights the challenges faced by the model in processing and understanding visual information. Furthermore, calculation error accounts for a small portion of the errors, especially in chemistry, indicating that the model excels particularly in handling numerical computations. Figure~\ref{fig:error_case_all} shows some cases of reasoning error category in the disciplines of mathematics, physics, and chemistry. More detailed examples of these errors can be found in Appendix~\ref{appendix: error_cases}.

\section{Related Works}
\subsection{Multi-modal Reasoning Benchmarks}

Recently, the evaluation of multi-modal large language models (MLLMs)~\cite{openai2023gpt4v,team2023gemini,Claude3,bai2023qwen-vl,wang2023cogvlm,liu2024improved,liu2024visual} in various reasoning tasks has become increasingly crucial. So many benchmark datasets for these tasks span several categories are proposed like MME~\cite{fu2023mme}, MMMU~\cite{yue2024mmmu}, MMBench~\cite{liu2023mmbench}, MMStar~\cite{chen2024we}, SEED-Bench~\cite{li2023seed}, and CMMMU~\cite{zhang2024cmmmu}, which evaluate models' capabilities to apply logic and inference; mathematical reasoning; scientific reasoning, and agent-based reasoning. These benchmark datasets provide comprehensive measurements of MLLMs' capabilities in applying specialized knowledge and decision-making in simulated environments. For instance, MMMU covers university-level questions from six domains, which is utilized to assess MLLMs' advanced perception and reasoning abilities. CMMMU~\cite{zhang2024cmmmu} evaluates models' reasoning abilities across various disciplines through bilingual multi-modal questions in Chinese and English. Existing benchmark like ScienceQA~\cite{lu2022learn} is a specialized dataset designed to evaluate the capabilities of MLLMs, particularly in the domain of scientific reasoning. It contains over 21,000 questions that span a wide range of scientific disciplines, including physics, chemistry, biology, and earth science. Furthermore, several benchmarks such as MathVista~\cite{lu2023mathvista}, MathVerse~\cite{zhang2024mathverse}, and Math-Vision~\cite{wang2024measuring} are specially designed to evaluate the mathematical reasoning capabilities of MLLMs. MathVista comprises 6,141 examples drawn from 31 multi-modal datasets, with a commonly used minitest subset containing 1,000 examples specifically designed to evaluate the mathematical reasoning capabilities of MLLMs. MathVerse contains 2,612 high-quality, multi-subject mathematical problems covering key areas like plane geometry, solid geometry, and functions. Each problem is transformed into six distinct versions to provide varying levels of visual information: Text-dominant Version, Text-lite Version, Text-only Version, Vision-intensive Version, Vision-dominant Version, and Vision-only Version. Math-Vision is a carefully curated collection of 3,040 high-quality math problems, sourced from real math competitions, which covers 16 distinct mathematical disciplines and is graded across five levels of difficulty. 
While these benchmarks are valuable, they present limitations such as an overemphasis on mathematics and a broad array of topics that often lack depth in science-related questions and exhibit uneven difficulty levels. Our dataset addresses these shortcomings by providing 3,000 scientific reasoning questions across mathematics, physics, and chemistry, which is collected from K12 education. Additionally, it includes bilingual questions in Chinese and English, enriching the knowledge base and offering a more extensive range of difficulty levels to create a more comprehensive evaluation platform.

\subsection{Multi-modal Large Language Models}

Recently, the success of large language models (LLMs)~\cite{du2021glm,zeng2022glm,achiam2023gpt,gao2023llama,glm2024chatglm,bai2023qwen} has spurred the ongoing development of multi-modal large language models (MLLMs). These MLLMs~\cite{liu2024visual,liu2024llava,wang2023cogvlm,li2023blip,dai2024instructblip,bai2023qwen} expand upon traditional LLM capabilities by integrating the ability to process and analyze both text and images. For instance, models like miniGPT~\cite{zhu2023minigpt} and InstructBLIP~\cite{dai2024instructblip} attempt to utilize a trainable Q-Former or a linear layer to connect a frozen pretrained vision encoder and language model. Subsequently, LLaVA~\cite{liu2024visual,liu2024llava} presents visual instruction tuning, which achieves a end-to-end fine-tuning on a large multi-modal model (LMM) comprising visual encoder and language model. Currently, close-source MLLMs like Gemini~\cite{team2023gemini}, GPT-4v~\cite{openai2023gpt4v}, Qwen-VL~\cite{bai2023qwen-vl}, and Claude3~\cite{Claude3} demonstrate impressive capabilities in general image understanding and scientific reasoning. Besides, the development of open-source multi-modal large language models (MLLMs) continues to expand, providing an important complement to their closed-source models. These open-source MLLMs, such as mPLUG-Owl~\cite{ye2023mplug,ye2024mplug}, LLaMA-Adapter-V2~\cite{gao2023llama}, MiniCPM~\cite{hu2024minicpm}, LLaVA-1.5~\cite{liu2024improved}, LLaVA-NeXT~\cite{liu2024llava}, DeepSeek-VL~\cite{lu2024deepseek}, ShareGPT4V~\cite{chen2023sharegpt4v}, SPHINX~\cite{gao2024sphinx}, InternVL~\cite{chen2023internvl}, InternVL 1.5~\cite{chen2024far}, InternLM-XComposer2~\cite{dong2024internlm}, and CogVLM~\cite{wang2023cogvlm}, also achieves advance performance, further enriching the landscape of MLLM domain. Here, we utilize our specially curated benchmark to evaluate these MLLMs across tasks in mathematics, physics, and chemistry. This comprehensive evaluation aims to assess their capabilities in image understanding and scientific reasoning.

\section{Conclusion}
In this paper, we introduce a comprehensive benchmark, \data, to evaluate the capabilities of MLLMs in multi-modal scientific reasoning. \data comprises 3,000 questions across three disciplines of mathematics, physics, and chemistry, spanning 21 subjects and 5 difficulty levels. We conduct evaluations on \data with a total of 25 prominent models, including close-source and open-source models. Experimental results demonstrate that close-source MLLMs generally outperform open-source models, showing particularly more better capabilities in complex problem-solving and analytical reasoning. Notable models such as Claude3.5-Sonnet, GPT-4o, and Gemini-1.5-Pro exhibit superior performance across three disciplines. Specifically, Claude3.5-Sonnet achieves an accuracy of 53.4\% in mathematics, GPT-4o accomplishes a 38.2\% accuracy in physics, and Gemini-1.5-Pro reaches an accuracy of 30.1\% in chemistry. Although a gap remains between open-source models and the best-performing closed-source models, the open-source model like InternVL-1.2-Plus exhibits competitive advantages. For instance, InternVL-1.2-Plus outperforms models like Gemini-1.0-Pro across all three disciplines. By providing a comprehensive and challenging set of questions across three scientific disciplines, \data ensures a robust assessment of MLLMs' ability in scientific reasoning.

{\small
\bibliographystyle{plainnat}
\bibliography{reference}

\begin{thebibliography}{54}
\providecommand{\natexlab}[1]{#1}
\providecommand{\url}[1]{\texttt{#1}}
\expandafter\ifx\csname urlstyle\endcsname\relax
  \providecommand{\doi}[1]{doi: #1}\else
  \providecommand{\doi}{doi: \begingroup \urlstyle{rm}\Url}\fi

\bibitem[Achiam et~al.(2023)Achiam, Adler, Agarwal, Ahmad, Akkaya, Aleman, Almeida, Altenschmidt, Altman, Anadkat, et~al.]{achiam2023gpt}
Josh Achiam, Steven Adler, Sandhini Agarwal, Lama Ahmad, Ilge Akkaya, Florencia~Leoni Aleman, Diogo Almeida, Janko Altenschmidt, Sam Altman, Shyamal Anadkat, et~al.
\newblock Gpt-4 technical report.
\newblock \emph{arXiv preprint arXiv:2303.08774}, 2023.

\bibitem[AI(2023)]{glm4v}
Zhipu AI.
\newblock Glm-4v, 2023.
\newblock URL \url{https://open.bigmodel.cn/dev/api#glm-4v}.

\bibitem[Anil et~al.(2023)Anil, Dai, Firat, Johnson, Lepikhin, Passos, Shakeri, Taropa, Bailey, Chen, et~al.]{anil2023palm2}
Rohan Anil, Andrew~M Dai, Orhan Firat, Melvin Johnson, Dmitry Lepikhin, Alexandre Passos, Siamak Shakeri, Emanuel Taropa, Paige Bailey, Zhifeng Chen, et~al.
\newblock Palm 2 technical report.
\newblock \emph{arXiv preprint arXiv:2305.10403}, 2023.

\bibitem[Anthropic(2023{\natexlab{a}})]{claude2}
Anthropic.
\newblock Claude 2, 2023{\natexlab{a}}.
\newblock URL \url{https://www.anthropic.com/index/claude-2}.

\bibitem[Anthropic(2023{\natexlab{b}})]{claude3.5}
Anthropic.
\newblock Claude 3.5, 2023{\natexlab{b}}.
\newblock URL \url{https://www.anthropic.com/news/claude-3-5-sonnet}.

\bibitem[Anthropic(2024)]{Claude3}
Anthropic.
\newblock The claude 3 model family: Opus, sonnet, haiku.
\newblock 2024.
\newblock URL \url{https://www-cdn.anthropic.com/de8ba9b01c9ab7cbabf5c33b80b7bbc618857627/Model_Card_Claude_3.pdf}.

\bibitem[Bai et~al.(2023{\natexlab{a}})Bai, Bai, Chu, Cui, Dang, Deng, Fan, Ge, Han, Huang, et~al.]{bai2023qwen}
Jinze Bai, Shuai Bai, Yunfei Chu, Zeyu Cui, Kai Dang, Xiaodong Deng, Yang Fan, Wenbin Ge, Yu~Han, Fei Huang, et~al.
\newblock Qwen technical report.
\newblock \emph{arXiv preprint arXiv:2309.16609}, 2023{\natexlab{a}}.

\bibitem[Bai et~al.(2023{\natexlab{b}})Bai, Bai, Yang, Wang, Tan, Wang, Lin, Zhou, and Zhou]{bai2023qwen-vl}
Jinze Bai, Shuai Bai, Shusheng Yang, Shijie Wang, Sinan Tan, Peng Wang, Junyang Lin, Chang Zhou, and Jingren Zhou.
\newblock Qwen-vl: A frontier large vision-language model with versatile abilities.
\newblock \emph{arXiv preprint arXiv:2308.12966}, 2023{\natexlab{b}}.

\bibitem[Brown et~al.(2020)Brown, Mann, Ryder, Subbiah, Kaplan, Dhariwal, Neelakantan, Shyam, Sastry, Askell, et~al.]{brown2020language}
Tom Brown, Benjamin Mann, Nick Ryder, Melanie Subbiah, Jared~D Kaplan, Prafulla Dhariwal, Arvind Neelakantan, Pranav Shyam, Girish Sastry, Amanda Askell, et~al.
\newblock Language models are few-shot learners.
\newblock \emph{Advances in neural information processing systems}, 33:\penalty0 1877--1901, 2020.

\bibitem[Cao and Xiao(2022)]{cao2022augmented}
Jie Cao and Jing Xiao.
\newblock An augmented benchmark dataset for geometric question answering through dual parallel text encoding.
\newblock In \emph{Proceedings of the 29th International Conference on Computational Linguistics}, pages 1511--1520, 2022.

\bibitem[Chen et~al.(2021)Chen, Tang, Qin, Liang, Liu, Xing, and Lin]{chen2021geoqa}
Jiaqi Chen, Jianheng Tang, Jinghui Qin, Xiaodan Liang, Lingbo Liu, Eric~P Xing, and Liang Lin.
\newblock Geoqa: A geometric question answering benchmark towards multimodal numerical reasoning.
\newblock \emph{arXiv preprint arXiv:2105.14517}, 2021.

\bibitem[Chen et~al.(2022)Chen, Li, Qin, Lu, Lin, Chen, and Liang]{chen2022unigeo}
Jiaqi Chen, Tong Li, Jinghui Qin, Pan Lu, Liang Lin, Chongyu Chen, and Xiaodan Liang.
\newblock Unigeo: Unifying geometry logical reasoning via reformulating mathematical expression.
\newblock \emph{arXiv preprint arXiv:2212.02746}, 2022.

\bibitem[Chen et~al.(2023{\natexlab{a}})Chen, Li, Dong, Zhang, He, Wang, Zhao, and Lin]{chen2023sharegpt4v}
Lin Chen, Jisong Li, Xiaoyi Dong, Pan Zhang, Conghui He, Jiaqi Wang, Feng Zhao, and Dahua Lin.
\newblock Sharegpt4v: Improving large multi-modal models with better captions.
\newblock \emph{arXiv preprint arXiv:2311.12793}, 2023{\natexlab{a}}.

\bibitem[Chen et~al.(2024{\natexlab{a}})Chen, Li, Dong, Zhang, Zang, Chen, Duan, Wang, Qiao, Lin, et~al.]{chen2024we}
Lin Chen, Jinsong Li, Xiaoyi Dong, Pan Zhang, Yuhang Zang, Zehui Chen, Haodong Duan, Jiaqi Wang, Yu~Qiao, Dahua Lin, et~al.
\newblock Are we on the right way for evaluating large vision-language models?
\newblock \emph{arXiv preprint arXiv:2403.20330}, 2024{\natexlab{a}}.

\bibitem[Chen et~al.(2023{\natexlab{b}})Chen, Wu, Wang, Su, Chen, Xing, Zhong, Zhang, Zhu, Lu, Li, Luo, Lu, Qiao, and Dai]{chen2023internvl}
Zhe Chen, Jiannan Wu, Wenhai Wang, Weijie Su, Guo Chen, Sen Xing, Muyan Zhong, Qinglong Zhang, Xizhou Zhu, Lewei Lu, Bin Li, Ping Luo, Tong Lu, Yu~Qiao, and Jifeng Dai.
\newblock Internvl: Scaling up vision foundation models and aligning for generic visual-linguistic tasks.
\newblock \emph{arXiv preprint arXiv:2312.14238}, 2023{\natexlab{b}}.

\bibitem[Chen et~al.(2024{\natexlab{b}})Chen, Wang, Tian, Ye, Gao, Cui, Tong, Hu, Luo, Ma, et~al.]{chen2024far}
Zhe Chen, Weiyun Wang, Hao Tian, Shenglong Ye, Zhangwei Gao, Erfei Cui, Wenwen Tong, Kongzhi Hu, Jiapeng Luo, Zheng Ma, et~al.
\newblock How far are we to gpt-4v? closing the gap to commercial multimodal models with open-source suites.
\newblock \emph{arXiv preprint arXiv:2404.16821}, 2024{\natexlab{b}}.

\bibitem[Chen et~al.(2024{\natexlab{c}})Chen, Wu, Wang, Su, Chen, Xing, Zhong, Zhang, Zhu, Lu, et~al.]{chen2024internvl}
Zhe Chen, Jiannan Wu, Wenhai Wang, Weijie Su, Guo Chen, Sen Xing, Muyan Zhong, Qinglong Zhang, Xizhou Zhu, Lewei Lu, et~al.
\newblock Internvl: Scaling up vision foundation models and aligning for generic visual-linguistic tasks.
\newblock In \emph{Proceedings of the IEEE/CVF Conference on Computer Vision and Pattern Recognition}, pages 24185--24198, 2024{\natexlab{c}}.

\bibitem[Chowdhery et~al.(2023)Chowdhery, Narang, Devlin, Bosma, Mishra, Roberts, Barham, Chung, Sutton, Gehrmann, et~al.]{chowdhery2023palm}
Aakanksha Chowdhery, Sharan Narang, Jacob Devlin, Maarten Bosma, Gaurav Mishra, Adam Roberts, Paul Barham, Hyung~Won Chung, Charles Sutton, Sebastian Gehrmann, et~al.
\newblock Palm: Scaling language modeling with pathways.
\newblock \emph{Journal of Machine Learning Research}, 24\penalty0 (240):\penalty0 1--113, 2023.

\bibitem[Dai et~al.(2024)Dai, Li, Li, Tiong, Zhao, Wang, Li, Fung, and Hoi]{dai2024instructblip}
Wenliang Dai, Junnan Li, Dongxu Li, Anthony Meng~Huat Tiong, Junqi Zhao, Weisheng Wang, Boyang Li, Pascale~N Fung, and Steven Hoi.
\newblock Instructblip: Towards general-purpose vision-language models with instruction tuning.
\newblock \emph{Advances in Neural Information Processing Systems}, 36, 2024.

\bibitem[Dong et~al.(2024)Dong, Zhang, Zang, Cao, Wang, Ouyang, Wei, Zhang, Duan, Cao, et~al.]{dong2024internlm}
Xiaoyi Dong, Pan Zhang, Yuhang Zang, Yuhang Cao, Bin Wang, Linke Ouyang, Xilin Wei, Songyang Zhang, Haodong Duan, Maosong Cao, et~al.
\newblock Internlm-xcomposer2: Mastering free-form text-image composition and comprehension in vision-language large model.
\newblock \emph{arXiv preprint arXiv:2401.16420}, 2024.

\bibitem[Du et~al.(2021)Du, Qian, Liu, Ding, Qiu, Yang, and Tang]{du2021glm}
Zhengxiao Du, Yujie Qian, Xiao Liu, Ming Ding, Jiezhong Qiu, Zhilin Yang, and Jie Tang.
\newblock Glm: General language model pretraining with autoregressive blank infilling.
\newblock \emph{arXiv preprint arXiv:2103.10360}, 2021.

\bibitem[Fu et~al.(2023)Fu, Chen, Shen, Qin, Zhang, Lin, Qiu, Lin, Yang, Zheng, et~al.]{fu2023mme}
Chaoyou Fu, Peixian Chen, Yunhang Shen, Yulei Qin, Mengdan Zhang, Xu~Lin, Zhenyu Qiu, Wei Lin, Jinrui Yang, Xiawu Zheng, et~al.
\newblock Mme: A comprehensive evaluation benchmark for multimodal large language models.
\newblock \emph{arXiv preprint arXiv:2306.13394}, 2023.

\bibitem[Gao et~al.(2023)Gao, Han, Zhang, Lin, Geng, Zhou, Zhang, Lu, He, Yue, et~al.]{gao2023llama}
Peng Gao, Jiaming Han, Renrui Zhang, Ziyi Lin, Shijie Geng, Aojun Zhou, Wei Zhang, Pan Lu, Conghui He, Xiangyu Yue, et~al.
\newblock Llama-adapter v2: Parameter-efficient visual instruction model.
\newblock \emph{arXiv preprint arXiv:2304.15010}, 2023.

\bibitem[Gao et~al.(2024)Gao, Zhang, Liu, Qiu, Huang, Lin, Zhao, Geng, Lin, Jin, et~al.]{gao2024sphinx}
Peng Gao, Renrui Zhang, Chris Liu, Longtian Qiu, Siyuan Huang, Weifeng Lin, Shitian Zhao, Shijie Geng, Ziyi Lin, Peng Jin, et~al.
\newblock Sphinx-x: Scaling data and parameters for a family of multi-modal large language models.
\newblock \emph{arXiv preprint arXiv:2402.05935}, 2024.

\bibitem[GLM et~al.(2024)GLM, Zeng, Xu, Wang, Zhang, Yin, Rojas, Feng, Zhao, Lai, et~al.]{glm2024chatglm}
Team GLM, Aohan Zeng, Bin Xu, Bowen Wang, Chenhui Zhang, Da~Yin, Diego Rojas, Guanyu Feng, Hanlin Zhao, Hanyu Lai, et~al.
\newblock Chatglm: A family of large language models from glm-130b to glm-4 all tools.
\newblock \emph{arXiv preprint arXiv:2406.12793}, 2024.

\bibitem[Hu et~al.(2024)Hu, Tu, Han, He, Cui, Long, Zheng, Fang, Huang, Zhao, et~al.]{hu2024minicpm}
Shengding Hu, Yuge Tu, Xu~Han, Chaoqun He, Ganqu Cui, Xiang Long, Zhi Zheng, Yewei Fang, Yuxiang Huang, Weilin Zhao, et~al.
\newblock Minicpm: Unveiling the potential of small language models with scalable training strategies.
\newblock \emph{arXiv preprint arXiv:2404.06395}, 2024.

\bibitem[Kembhavi et~al.(2017)Kembhavi, Seo, Schwenk, Choi, Farhadi, and Hajishirzi]{tqa}
Aniruddha Kembhavi, Minjoon Seo, Dustin Schwenk, Jonghyun Choi, Ali Farhadi, and Hannaneh Hajishirzi.
\newblock Are you smarter than a sixth grader? textbook question answering for multimodal machine comprehension.
\newblock In \emph{2017 IEEE Conference on Computer Vision and Pattern Recognition (CVPR)}, pages 5376--5384, 2017.
\newblock \doi{10.1109/CVPR.2017.571}.

\bibitem[Li et~al.(2023{\natexlab{a}})Li, Wang, Wang, Ge, Ge, and Shan]{li2023seed}
Bohao Li, Rui Wang, Guangzhi Wang, Yuying Ge, Yixiao Ge, and Ying Shan.
\newblock Seed-bench: Benchmarking multimodal llms with generative comprehension.
\newblock \emph{arXiv preprint arXiv:2307.16125}, 2023{\natexlab{a}}.

\bibitem[Li et~al.(2023{\natexlab{b}})Li, Li, Savarese, and Hoi]{li2023blip}
Junnan Li, Dongxu Li, Silvio Savarese, and Steven Hoi.
\newblock Blip-2: Bootstrapping language-image pre-training with frozen image encoders and large language models.
\newblock In \emph{International conference on machine learning}, pages 19730--19742. PMLR, 2023{\natexlab{b}}.

\bibitem[Liu et~al.()Liu, Li, Li, Li, Zhang, Shen, and Lee]{liu2024llava}
Haotian Liu, Chunyuan Li, Yuheng Li, Bo~Li, Yuanhan Zhang, Sheng Shen, and Yong~Jae Lee.
\newblock Llava-next: Improved reasoning, ocr, and world knowledge (january 2024).
\newblock \emph{URL https://llava-vl. github. io/blog/2024-01-30-llava-next}, 1\penalty0 (8).

\bibitem[Liu et~al.(2024{\natexlab{a}})Liu, Li, Li, and Lee]{liu2024improved}
Haotian Liu, Chunyuan Li, Yuheng Li, and Yong~Jae Lee.
\newblock Improved baselines with visual instruction tuning.
\newblock In \emph{Proceedings of the IEEE/CVF Conference on Computer Vision and Pattern Recognition}, pages 26296--26306, 2024{\natexlab{a}}.

\bibitem[Liu et~al.(2024{\natexlab{b}})Liu, Li, Wu, and Lee]{liu2024visual}
Haotian Liu, Chunyuan Li, Qingyang Wu, and Yong~Jae Lee.
\newblock Visual instruction tuning.
\newblock \emph{Advances in neural information processing systems}, 36, 2024{\natexlab{b}}.

\bibitem[Liu et~al.(2023)Liu, Duan, Zhang, Li, Zhang, Zhao, Yuan, Wang, He, Liu, et~al.]{liu2023mmbench}
Yuan Liu, Haodong Duan, Yuanhan Zhang, Bo~Li, Songyang Zhang, Wangbo Zhao, Yike Yuan, Jiaqi Wang, Conghui He, Ziwei Liu, et~al.
\newblock Mmbench: Is your multi-modal model an all-around player?
\newblock \emph{arXiv preprint arXiv:2307.06281}, 2023.

\bibitem[Lu et~al.(2024)Lu, Liu, Zhang, Wang, Dong, Liu, Sun, Ren, Li, Sun, et~al.]{lu2024deepseek}
Haoyu Lu, Wen Liu, Bo~Zhang, Bingxuan Wang, Kai Dong, Bo~Liu, Jingxiang Sun, Tongzheng Ren, Zhuoshu Li, Yaofeng Sun, et~al.
\newblock Deepseek-vl: towards real-world vision-language understanding.
\newblock \emph{arXiv preprint arXiv:2403.05525}, 2024.

\bibitem[Lu et~al.(2021)Lu, Gong, Jiang, Qiu, Huang, Liang, and Zhu]{lu2021inter}
Pan Lu, Ran Gong, Shibiao Jiang, Liang Qiu, Siyuan Huang, Xiaodan Liang, and Song-Chun Zhu.
\newblock Inter-gps: Interpretable geometry problem solving with formal language and symbolic reasoning.
\newblock \emph{arXiv preprint arXiv:2105.04165}, 2021.

\bibitem[Lu et~al.(2022)Lu, Mishra, Xia, Qiu, Chang, Zhu, Tafjord, Clark, and Kalyan]{lu2022learn}
Pan Lu, Swaroop Mishra, Tanglin Xia, Liang Qiu, Kai-Wei Chang, Song-Chun Zhu, Oyvind Tafjord, Peter Clark, and Ashwin Kalyan.
\newblock Learn to explain: Multimodal reasoning via thought chains for science question answering.
\newblock \emph{Advances in Neural Information Processing Systems}, 35:\penalty0 2507--2521, 2022.

\bibitem[Lu et~al.(2023)Lu, Bansal, Xia, Liu, Li, Hajishirzi, Cheng, Chang, Galley, and Gao]{lu2023mathvista}
Pan Lu, Hritik Bansal, Tony Xia, Jiacheng Liu, Chunyuan Li, Hannaneh Hajishirzi, Hao Cheng, Kai-Wei Chang, Michel Galley, and Jianfeng Gao.
\newblock Mathvista: Evaluating mathematical reasoning of foundation models in visual contexts.
\newblock \emph{arXiv preprint arXiv:2310.02255}, 2023.

\bibitem[OpenAI(2022)]{openai2022chatgpt}
OpenAI.
\newblock Chatgpt, 2022.
\newblock URL \url{https://openai.com/blog/chatgpt}.

\bibitem[OpenAI(2023)]{openai2023gpt4v}
OpenAI.
\newblock Gpt-4v(ision) system card.
\newblock In \emph{technical report}, 2023.
\newblock URL \url{https://api.semanticscholar.org/CorpusID:263218031}.

\bibitem[OpenAI(2024)]{openai2024gpt4o}
OpenAI.
\newblock Gpt-4o, 2024.
\newblock URL \url{https://openai.com/index/hello-gpt-4o/}.

\bibitem[StepFun(2024)]{step-1v}
StepFun.
\newblock Step-1v, 2024.
\newblock URL \url{https://open.bigmodel.cn/dev/api#glm-4v}.

\bibitem[Team et~al.(2023)Team, Anil, Borgeaud, Wu, Alayrac, Yu, Soricut, Schalkwyk, Dai, Hauth, et~al.]{team2023gemini}
Gemini Team, Rohan Anil, Sebastian Borgeaud, Yonghui Wu, Jean-Baptiste Alayrac, Jiahui Yu, Radu Soricut, Johan Schalkwyk, Andrew~M Dai, Anja Hauth, et~al.
\newblock Gemini: a family of highly capable multimodal models.
\newblock \emph{arXiv preprint arXiv:2312.11805}, 2023.

\bibitem[Touvron et~al.(2023{\natexlab{a}})Touvron, Lavril, Izacard, Martinet, Lachaux, Lacroix, Rozi{\`e}re, Goyal, Hambro, Azhar, et~al.]{touvron2023llama1}
Hugo Touvron, Thibaut Lavril, Gautier Izacard, Xavier Martinet, Marie-Anne Lachaux, Timoth{\'e}e Lacroix, Baptiste Rozi{\`e}re, Naman Goyal, Eric Hambro, Faisal Azhar, et~al.
\newblock Llama: Open and efficient foundation language models.
\newblock \emph{arXiv preprint arXiv:2302.13971}, 2023{\natexlab{a}}.

\bibitem[Touvron et~al.(2023{\natexlab{b}})Touvron, Martin, Stone, Albert, Almahairi, Babaei, Bashlykov, Batra, Bhargava, Bhosale, et~al.]{touvron2023llama2}
Hugo Touvron, Louis Martin, Kevin Stone, Peter Albert, Amjad Almahairi, Yasmine Babaei, Nikolay Bashlykov, Soumya Batra, Prajjwal Bhargava, Shruti Bhosale, et~al.
\newblock Llama 2: Open foundation and fine-tuned chat models.
\newblock \emph{arXiv preprint arXiv:2307.09288}, 2023{\natexlab{b}}.

\bibitem[Wang et~al.(2024)Wang, Pan, Shi, Lu, Zhan, and Li]{wang2024measuring}
Ke~Wang, Junting Pan, Weikang Shi, Zimu Lu, Mingjie Zhan, and Hongsheng Li.
\newblock Measuring multimodal mathematical reasoning with math-vision dataset.
\newblock \emph{arXiv preprint arXiv:2402.14804}, 2024.

\bibitem[Wang et~al.(2023{\natexlab{a}})Wang, Lv, Yu, Hong, Qi, Wang, Ji, Yang, Zhao, Song, et~al.]{wang2023cogvlm}
Weihan Wang, Qingsong Lv, Wenmeng Yu, Wenyi Hong, Ji~Qi, Yan Wang, Junhui Ji, Zhuoyi Yang, Lei Zhao, Xixuan Song, et~al.
\newblock Cogvlm: Visual expert for pretrained language models.
\newblock \emph{arXiv preprint arXiv:2311.03079}, 2023{\natexlab{a}}.

\bibitem[Wang et~al.(2023{\natexlab{b}})Wang, Hu, Lu, Zhu, Zhang, Subramaniam, Loomba, Zhang, Sun, and Wang]{wang2023scibench}
Xiaoxuan Wang, Ziniu Hu, Pan Lu, Yanqiao Zhu, Jieyu Zhang, Satyen Subramaniam, Arjun~R Loomba, Shichang Zhang, Yizhou Sun, and Wei Wang.
\newblock Scibench: Evaluating college-level scientific problem-solving abilities of large language models.
\newblock \emph{arXiv preprint arXiv:2307.10635}, 2023{\natexlab{b}}.

\bibitem[Ye et~al.(2023)Ye, Xu, Xu, Ye, Yan, Zhou, Wang, Hu, Shi, Shi, et~al.]{ye2023mplug}
Qinghao Ye, Haiyang Xu, Guohai Xu, Jiabo Ye, Ming Yan, Yiyang Zhou, Junyang Wang, Anwen Hu, Pengcheng Shi, Yaya Shi, et~al.
\newblock mplug-owl: Modularization empowers large language models with multimodality.
\newblock \emph{arXiv preprint arXiv:2304.14178}, 2023.

\bibitem[Ye et~al.(2024)Ye, Xu, Ye, Yan, Hu, Liu, Qian, Zhang, and Huang]{ye2024mplug}
Qinghao Ye, Haiyang Xu, Jiabo Ye, Ming Yan, Anwen Hu, Haowei Liu, Qi~Qian, Ji~Zhang, and Fei Huang.
\newblock mplug-owl2: Revolutionizing multi-modal large language model with modality collaboration.
\newblock In \emph{Proceedings of the IEEE/CVF Conference on Computer Vision and Pattern Recognition}, pages 13040--13051, 2024.

\bibitem[Yue et~al.(2024)Yue, Ni, Zhang, Zheng, Liu, Zhang, Stevens, Jiang, Ren, Sun, et~al.]{yue2024mmmu}
Xiang Yue, Yuansheng Ni, Kai Zhang, Tianyu Zheng, Ruoqi Liu, Ge~Zhang, Samuel Stevens, Dongfu Jiang, Weiming Ren, Yuxuan Sun, et~al.
\newblock Mmmu: A massive multi-discipline multimodal understanding and reasoning benchmark for expert agi.
\newblock In \emph{Proceedings of the IEEE/CVF Conference on Computer Vision and Pattern Recognition}, pages 9556--9567, 2024.

\bibitem[Zeng et~al.(2022)Zeng, Liu, Du, Wang, Lai, Ding, Yang, Xu, Zheng, Xia, et~al.]{zeng2022glm}
Aohan Zeng, Xiao Liu, Zhengxiao Du, Zihan Wang, Hanyu Lai, Ming Ding, Zhuoyi Yang, Yifan Xu, Wendi Zheng, Xiao Xia, et~al.
\newblock Glm-130b: An open bilingual pre-trained model.
\newblock \emph{arXiv preprint arXiv:2210.02414}, 2022.

\bibitem[Zhang et~al.(2024{\natexlab{a}})Zhang, Du, Chen, Liang, Luo, Zheng, Zhu, Cheng, Xu, Guo, et~al.]{zhang2024cmmmu}
Ge~Zhang, Xinrun Du, Bei Chen, Yiming Liang, Tongxu Luo, Tianyu Zheng, Kang Zhu, Yuyang Cheng, Chunpu Xu, Shuyue Guo, et~al.
\newblock Cmmmu: A chinese massive multi-discipline multimodal understanding benchmark.
\newblock \emph{arXiv preprint arXiv:2401.11944}, 2024{\natexlab{a}}.

\bibitem[Zhang et~al.(2024{\natexlab{b}})Zhang, Jiang, Zhang, Lin, Guo, Qiu, Zhou, Lu, Chang, Gao, et~al.]{zhang2024mathverse}
Renrui Zhang, Dongzhi Jiang, Yichi Zhang, Haokun Lin, Ziyu Guo, Pengshuo Qiu, Aojun Zhou, Pan Lu, Kai-Wei Chang, Peng Gao, et~al.
\newblock Mathverse: Does your multi-modal llm truly see the diagrams in visual math problems?
\newblock \emph{arXiv preprint arXiv:2403.14624}, 2024{\natexlab{b}}.

\bibitem[Zhu et~al.(2023)Zhu, Chen, Shen, Li, and Elhoseiny]{zhu2023minigpt}
Deyao Zhu, Jun Chen, Xiaoqian Shen, Xiang Li, and Mohamed Elhoseiny.
\newblock Minigpt-4: Enhancing vision-language understanding with advanced large language models.
\newblock \emph{arXiv preprint arXiv:2304.10592}, 2023.

\end{thebibliography}
}

\appendix
\clearpage

\section{Dataset Details}\label{appendix:dataset_details}

\subsection{Question Length Distribution}\label{appendix:question_length}
\begin{figure}[htbp]
    \centering
    \begin{minipage}{0.45\linewidth}
        \includegraphics[width=\textwidth]{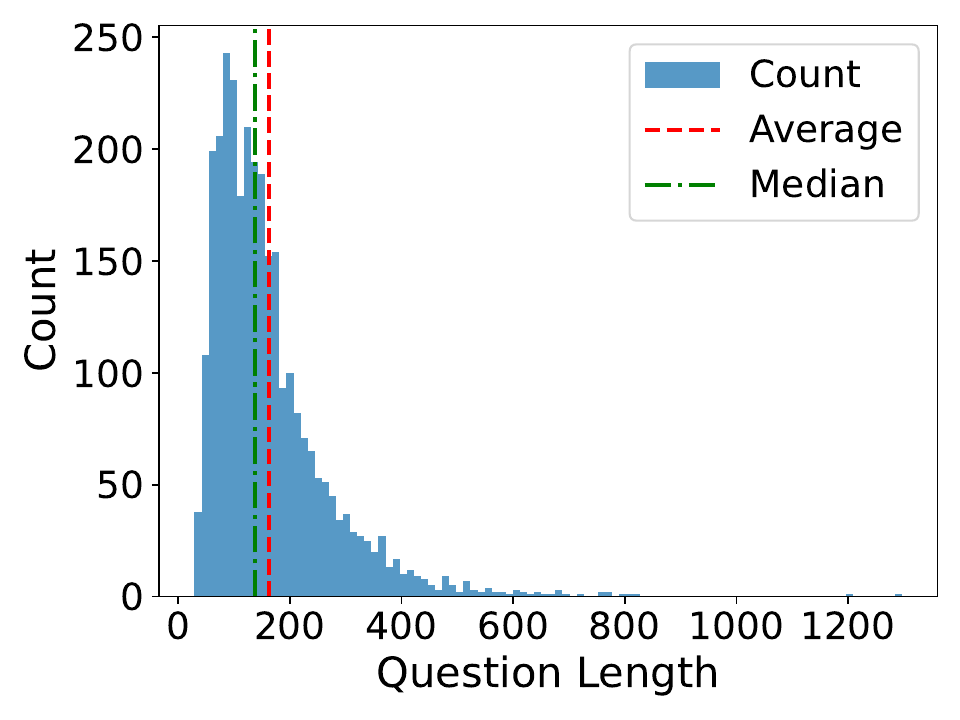}
        \centerline{(a) Chinese Version}
    \end{minipage}
    \begin{minipage}{0.45\linewidth}
        \includegraphics[width=\textwidth]{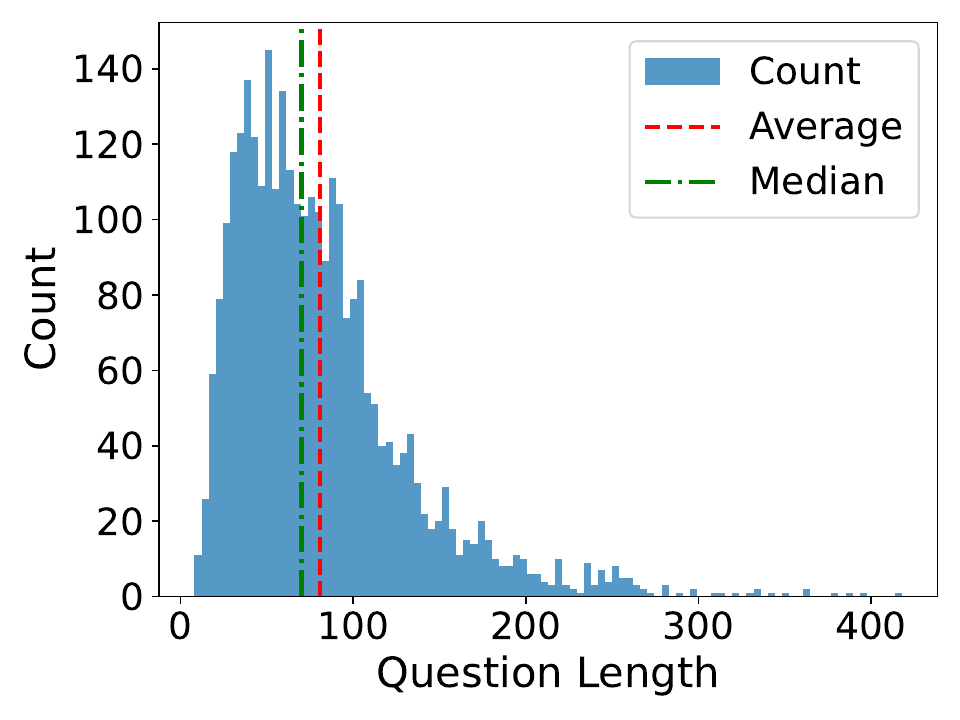}
        \centerline{(b) English Version}
    \end{minipage}
    \caption{The distributions of word counts per question in the Chinese and English versions of \data.}
    \label{fig:question_length}
\end{figure}

We provide both Chinese and English versions of the \data benchmark. The Chinese version features an average of 162.85 words per question, with the longest question comprising 1,297 words. Answers in this version average 20.93 words, with the longest reaching 112 words. Conversely, the English version shows an average of 80.93 words per question, with the longest question spanning 418 words. Answers here average 12.3 words, with the most detailed answer containing 92 words. Figure~\ref{fig:question_length} depicts the distribution of word counts, highlighting the diversity and complexity of questions.

\subsection{Detailed Description of Subjects}\label{appendix:description_of_subjects}
\data consists of three disciplines: mathematics, physics, and chemistry. The mathematics section includes six subjects: \textit{algebraic operations, combinatorial mathematics, functions and equations, probability and statistics, plane geometry, and solid geometry}. The physics section is composed of eight subjects: \textit{mechanics, optics, modern physics, mechanical motion, electromagnetism, vibrations and waves, comprehensive experiments and methods, and thermodynamics}. The chemistry section includes seven subjects: \textit{chemical experiments, chemical reactions, inorganic chemistry, organic chemistry, electrochemistry, substance composition, and chemical equilibrium}. A more detailed introduction of the above subjects is presented as follows:

\subsubsection{Mathematics}

\vpara{Algebraic Operations.} Algebraic operations include the manipulation of algebraic expressions, such as addition, subtraction, multiplication, and division. They are fundamental for solving algebraic equations and inequalities and are widely applied across various fields of mathematics.

\vpara{Combinatorial Mathematics.} Combinatorial mathematics studies the counting, arrangement, and combination of discrete structures, involving graph theory, number theory, and coding theory. It has significant applications in computer science, optimization, and probability theory.

\vpara{Functions and Equations.} Functions and equations are core parts of mathematics, dealing with relationships between variables and their representations. Functions are mappings between inputs and outputs, while equations are equalities concerning these mappings. Mastering knowledge of functions and equations is fundamental for solving many practical problems and is widely applied in engineering, physics, and economics.

\vpara{Probability and Statistics.} Probability and statistics study the laws of random events and methods of data analysis, including probability distributions, statistical inference, and data analysis techniques. They have broad applications in scientific research, engineering, and economics.

\vpara{Plane Geometry.} Plane geometry studies the shapes and figures in two-dimensional space, including points, lines, angles, and polygons. It is a fundamental part of mathematics education.

\vpara{Solid Geometry.} Solid geometry involves the study of geometric shapes in three-dimensional space, including points, lines, surfaces, and polyhedra. It examines the properties, volumes, and surface areas of these geometric bodies and is foundational for architecture, physics, and engineering.

\subsubsection{Physics}

\vpara{Mechanics.} Mechanics studies the motion of objects and the forces acting upon them, including classical mechanics, quantum mechanics, and relativistic mechanics. It is the foundation of physics and is widely applied in engineering, astronomy, and materials science.

\vpara{Optics.} Optics studies the properties of light and its interactions with matter, including reflection, refraction, interference, and diffraction. Optical technologies have broad applications in imaging, communication, and laser technology.

\vpara{Modern Physics.} Modern physics includes theories developed since the 20th century, such as quantum mechanics, relativity, and particle physics. These theories have expanded our understanding of the fundamental laws of nature.

\vpara{Mechanical Motion.} Mechanical motion studies the movement of objects under the influence of forces, including linear motion, rotational motion, and vibration. Understanding mechanical motion is fundamental for the design and analysis of mechanical systems.

\vpara{Electromagnetism.} Electromagnetism studies the interactions between electric and magnetic fields, including electrostatics, magnetic fields, and electromagnetic waves. It is the basis of modern physics and electrical engineering.

\vpara{Vibration and Waves.} Vibration and waves study vibrating systems and wave phenomena, including sound waves, light waves, and electromagnetic waves. They have broad applications in communication, acoustics, and optical technologies.

\vpara{Comprehensive Experiments and Methods.} Comprehensive experiments and methods involve using various experimental techniques and methods in physics teaching and research. They include designing and conducting experiments to observe and analyze the effects of specific variables on outcomes. Through comprehensive experiments, students can grasp the complexities of scientific research, cultivate scientific reasoning abilities, and understand the meticulousness and uncertainties of experimental work.

\vpara{Thermodynamics.} Thermodynamics studies the processes of energy transformation and transfer, including the laws of thermodynamics, thermodynamic systems, phase transitions, and heat engines. Thermodynamics is a fundamental aspect of both physics and engineering, with broad applications in energy, environmental science, and materials science. By investigating the relationship between internal and external energy of objects, thermodynamics reveals the basic principles of energy conversion and transfer in nature, providing theoretical support for the development of modern industrial technology.

\subsubsection{Chemistry}

\vpara{Chemical Experiment.} Chemical experiments involve studying the properties and changes of substances through experimental methods. Students learn to design experiments, observe chemical reactions, collect and analyze data, and draw conclusions in chemical experiments. Chemical experiments play a crucial role in understanding chemical theories and applying chemical knowledge.

\vpara{Chemical Reaction.} Chemical reactions study the chemical changes between substances, including reaction types, mechanisms, and rates. Understanding chemical reactions is essential for predicting and controlling chemical processes, which have wide applications in pharmaceutical manufacturing, materials science, and environmental engineering.

\vpara{Inorganic Chemistry.} Inorganic chemistry studies the properties and reactions of non-carbon elements and their compounds. It covers a wide range of topics from metals and non-metals to transition metals and coordination compounds and is key to understanding the periodic table of elements and chemical reaction mechanisms.

\vpara{Organic Chemistry.} Organic chemistry studies the structure, properties, and reactions of carbon-containing compounds. It has significant applications in pharmaceutical chemistry, materials science, and biochemistry.

\vpara{Electrochemistry.} Electrochemistry studies the interconversion between electrical and chemical energy, including processes such as batteries, electrolysis, and electroplating. Electrochemistry has important applications in energy storage, corrosion control, and electrochemical sensors.

\vpara{Substance Composition.} Substance composition studies the chemical composition and structure of substances, including the arrangement of molecules, atoms, and ions. It has important applications in chemistry, materials science, and biology.

\vpara{Chemical Equilibrium.} Chemical equilibrium studies the behavior of chemical reactions when they reach a dynamic equilibrium state, including equilibrium constants, Le Chatelier's principle, and solubility equilibrium. Understanding chemical equilibrium is essential for predicting reaction directions and optimizing chemical processes.

\section{Dataset Case}\label{appendix: dataset_case}
\begin{figure}[hbpt]
    \centering
    \includegraphics[width=1.0\textwidth]{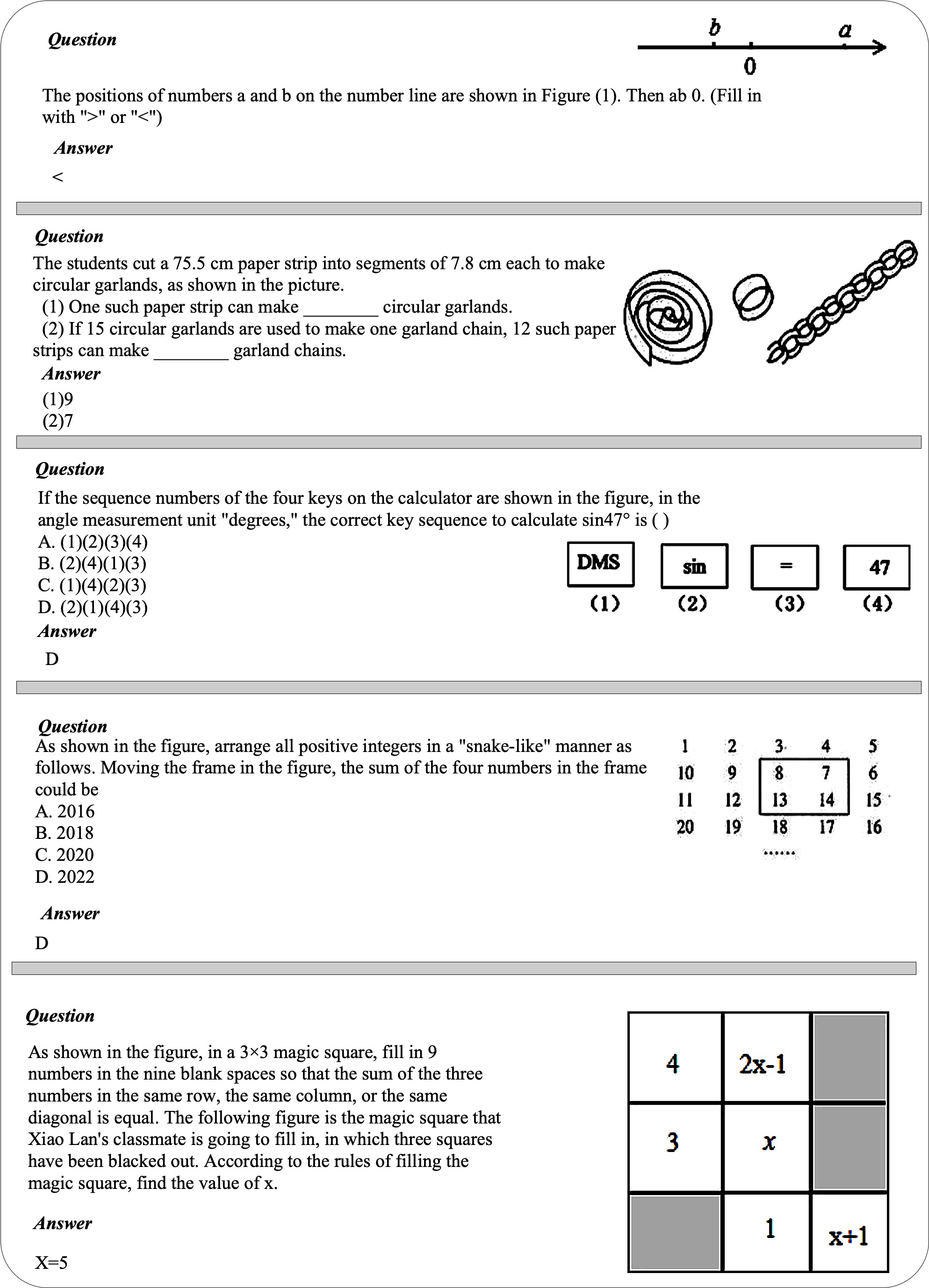}
    \vspace{-2mm}
    \caption{Cases of \textit{algebraic operations} in mathematical part of \data.}
    \label{fig:Algebraic_Operations}
    \vspace{-3mm}
\end{figure}

\begin{figure}[hbpt]
    \centering
    \includegraphics[width=0.98\textwidth,height=0.98\textheight]{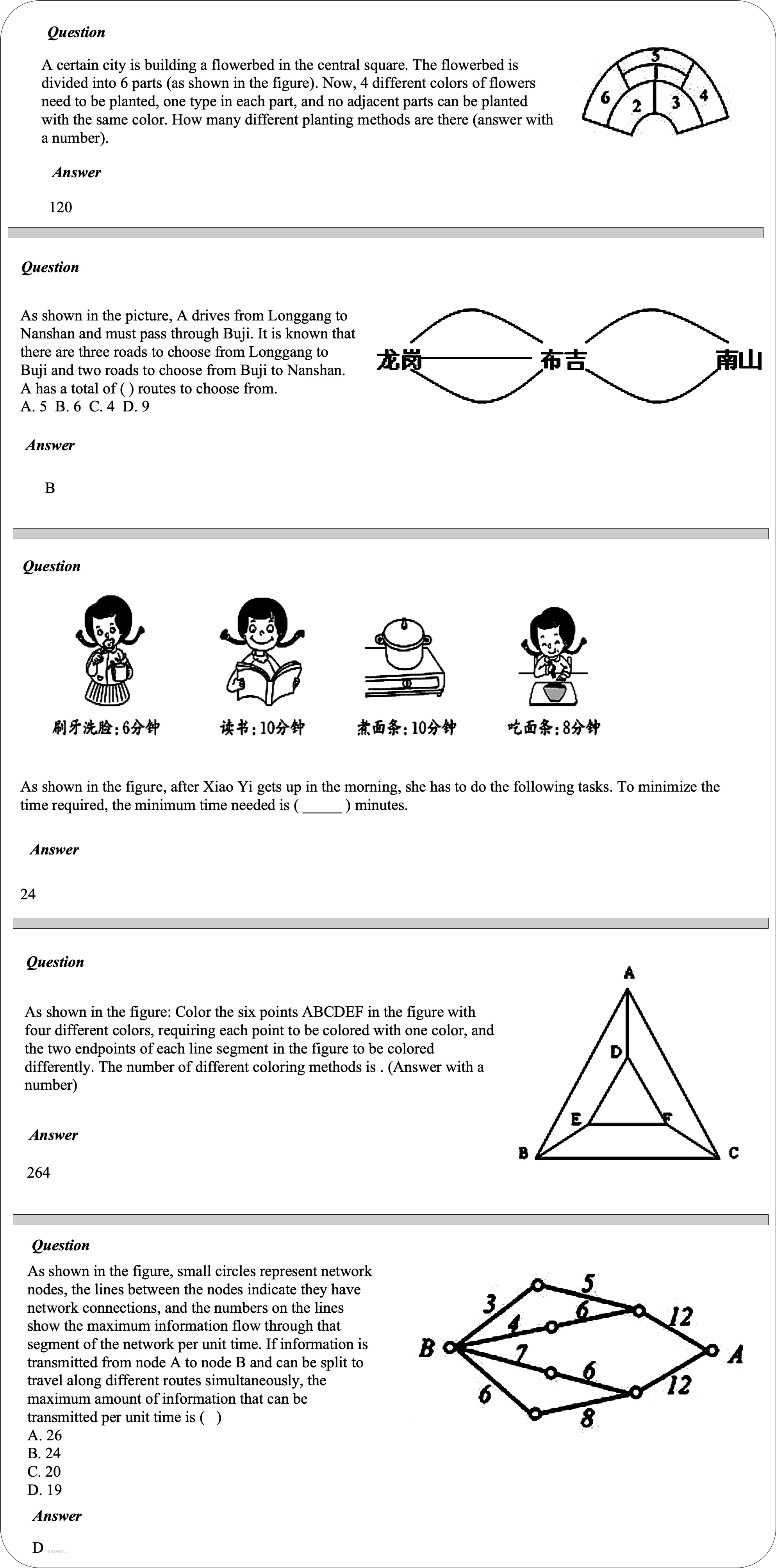} 
    \vspace{-1mm} 
    \caption{Cases of \textit{combinatorial mathematics} in mathematical part of \data.}
    \label{fig:Combinatorial_Mathematics}
    \vspace{-2mm} 
\end{figure}

\begin{figure}[hbpt]
    \centering
    \includegraphics[width=\textwidth, height=0.97\textheight, keepaspectratio]{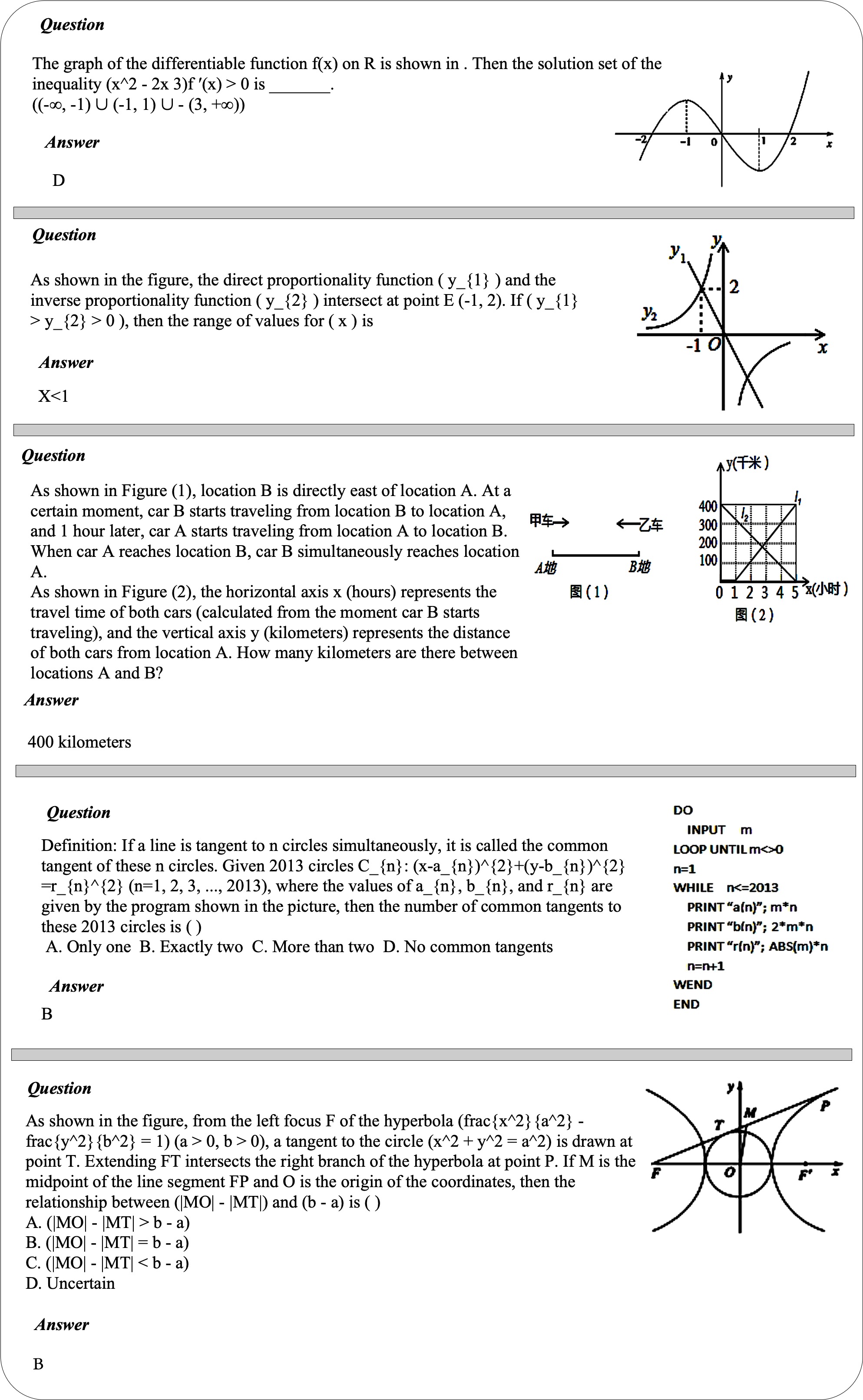}
    \caption{Cases of \textit{functions and equations} in mathematical part of \data.}
    \label{fig:Functions_and_Equations}
    \vspace{-3mm}
\end{figure}

\begin{figure}[hbpt]
    \centering
    \includegraphics[width=1.0\textwidth,height=0.9\textheight]{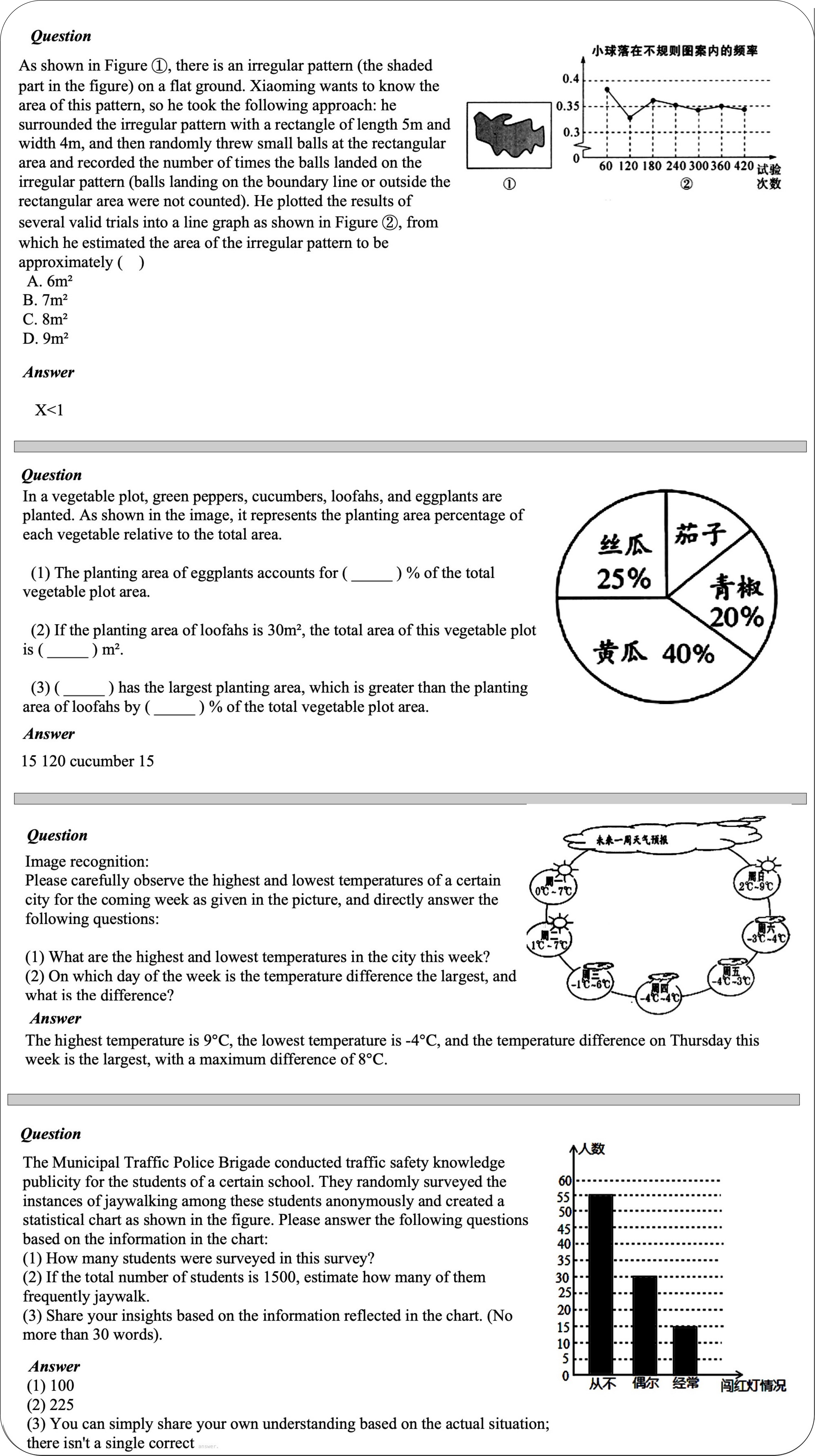}
    \caption{Cases of \textit{probability and statistics} in mathematical part of \data.}
    \label{fig:Probability_and_Statistics}
\end{figure}

\begin{figure}[hbpt]
    \centering
    \includegraphics[width=\textwidth, height=0.97\textheight, keepaspectratio]{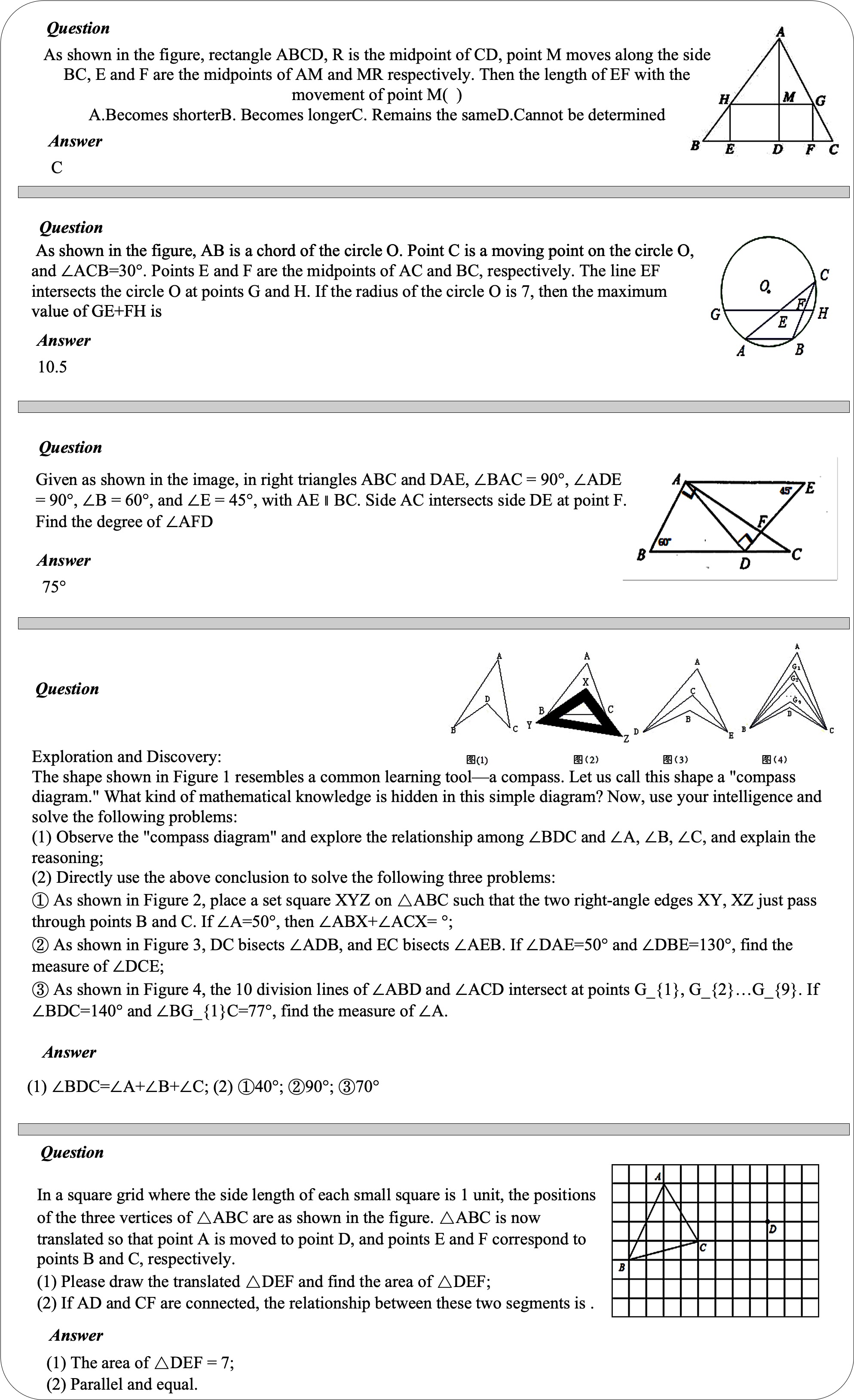}
    \caption{Cases of \textit{plane geometry} in mathematical part of \data.}
    \label{fig:Plane_Geometry}
    \vspace{-3mm}
\end{figure}

\begin{figure}[hbpt]
    \centering
    \includegraphics[width=1.0\textwidth]{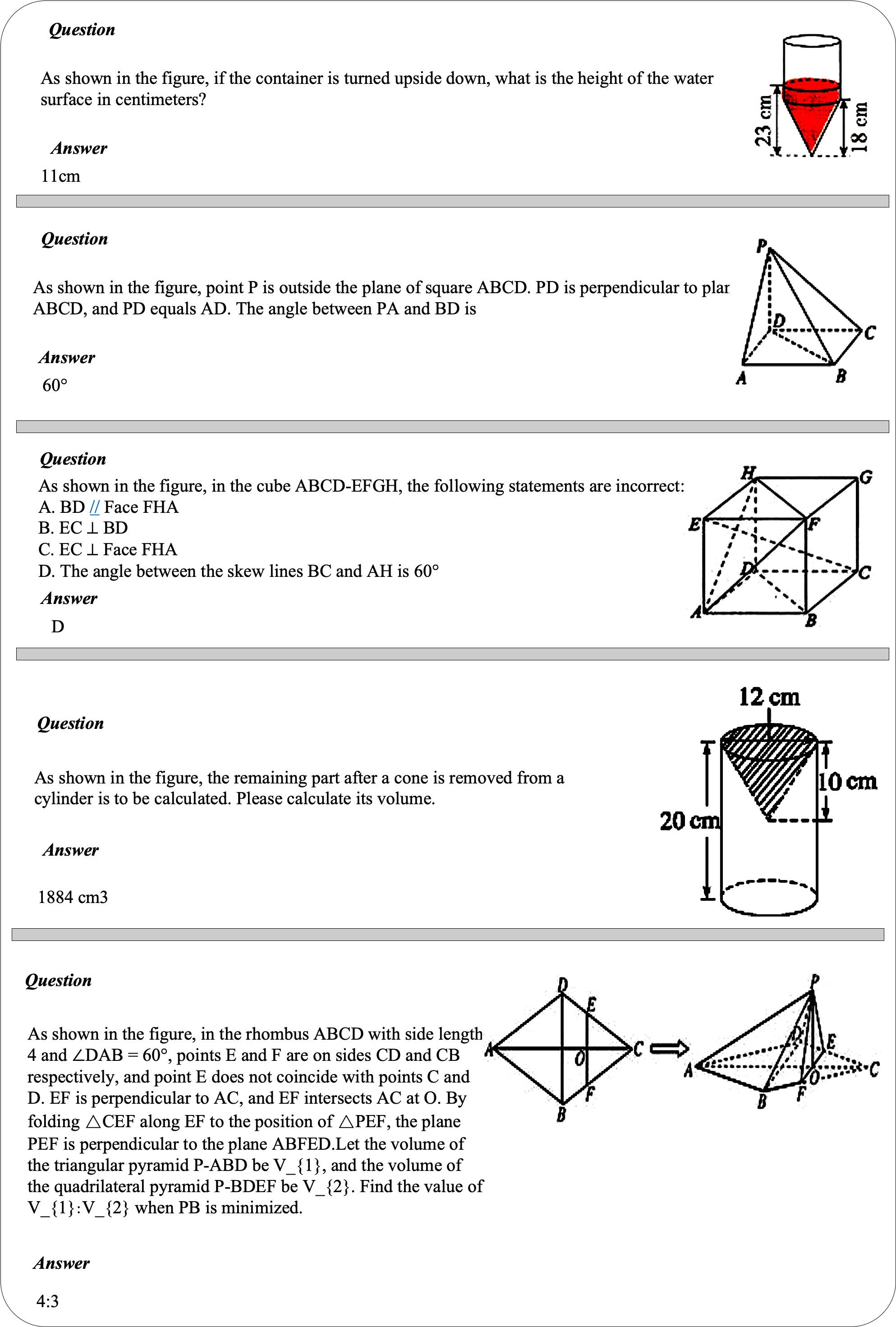}
    \vspace{-2mm}
    \caption{Cases of \textit{solid geometry} in mathematical part of \data.}
    \label{fig:Solid_Geometry}
    \vspace{-3mm}
\end{figure}

\begin{figure}[hbpt]
    \centering
    \includegraphics[width=\textwidth, height=0.97\textheight, keepaspectratio]{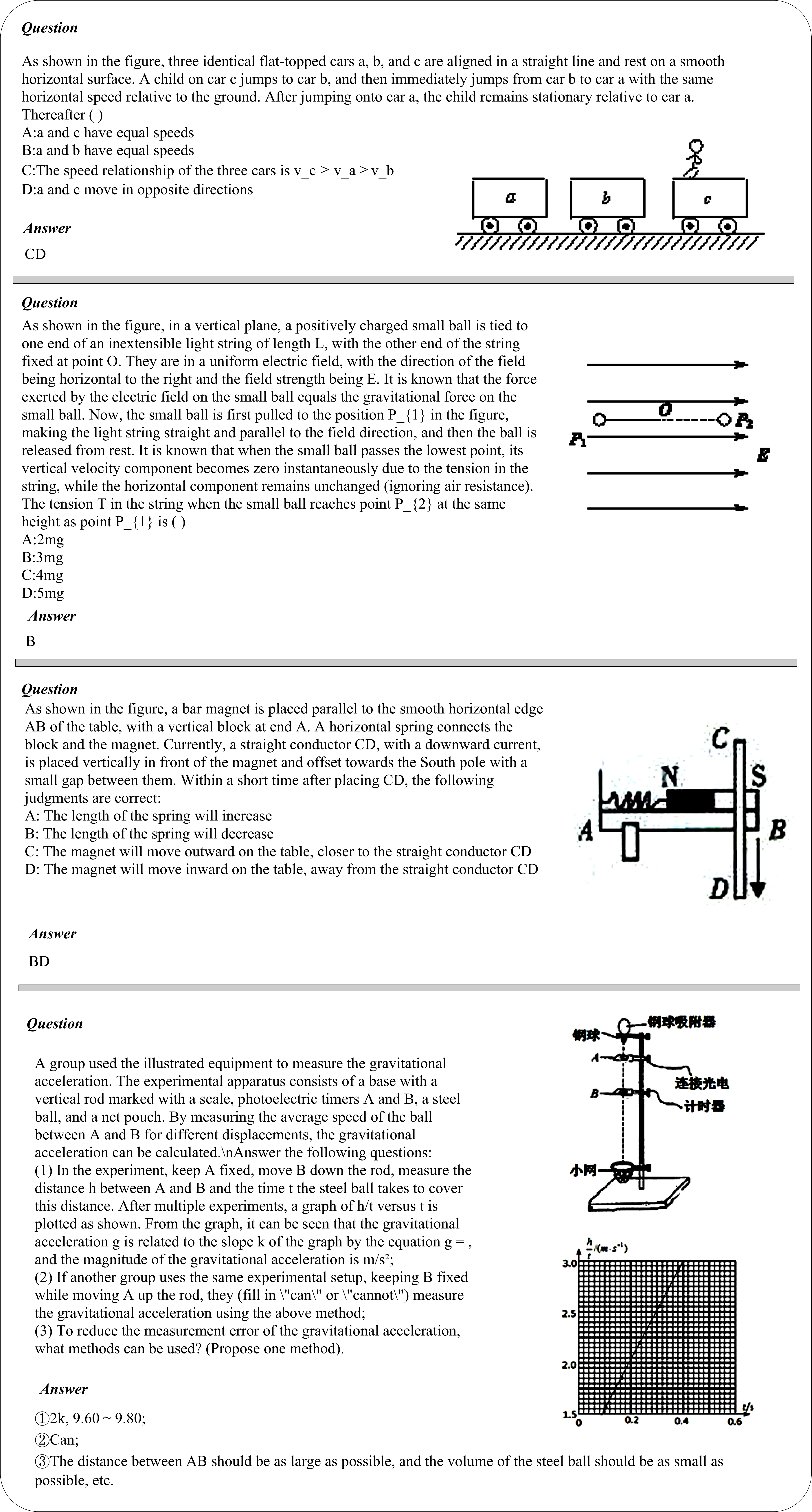}
    \caption{Cases of \textit{mechanics} in physics part of \data.}
    \label{fig:Mechanics}
    \vspace{-3mm}
\end{figure}

\begin{figure}[hbpt]
    \centering
    \includegraphics[width=\textwidth, height=0.97\textheight, keepaspectratio]{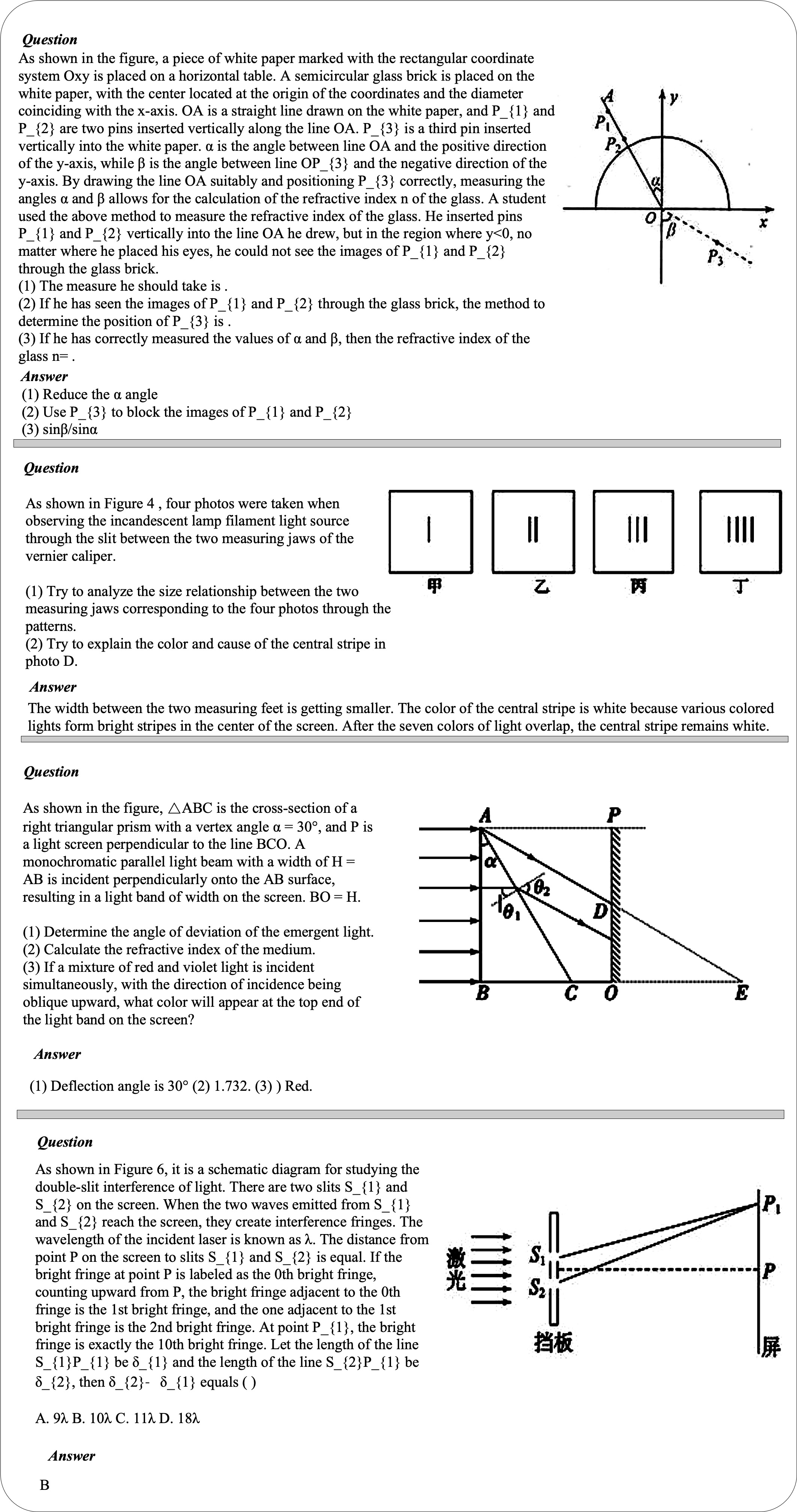}
    \caption{Cases of \textit{optics} in physics part of \data.}
    \label{fig:Optics}
    \vspace{-3mm}
\end{figure}

\begin{figure}[hbpt]
    \centering
    \includegraphics[width=\textwidth, height=0.97\textheight, keepaspectratio]{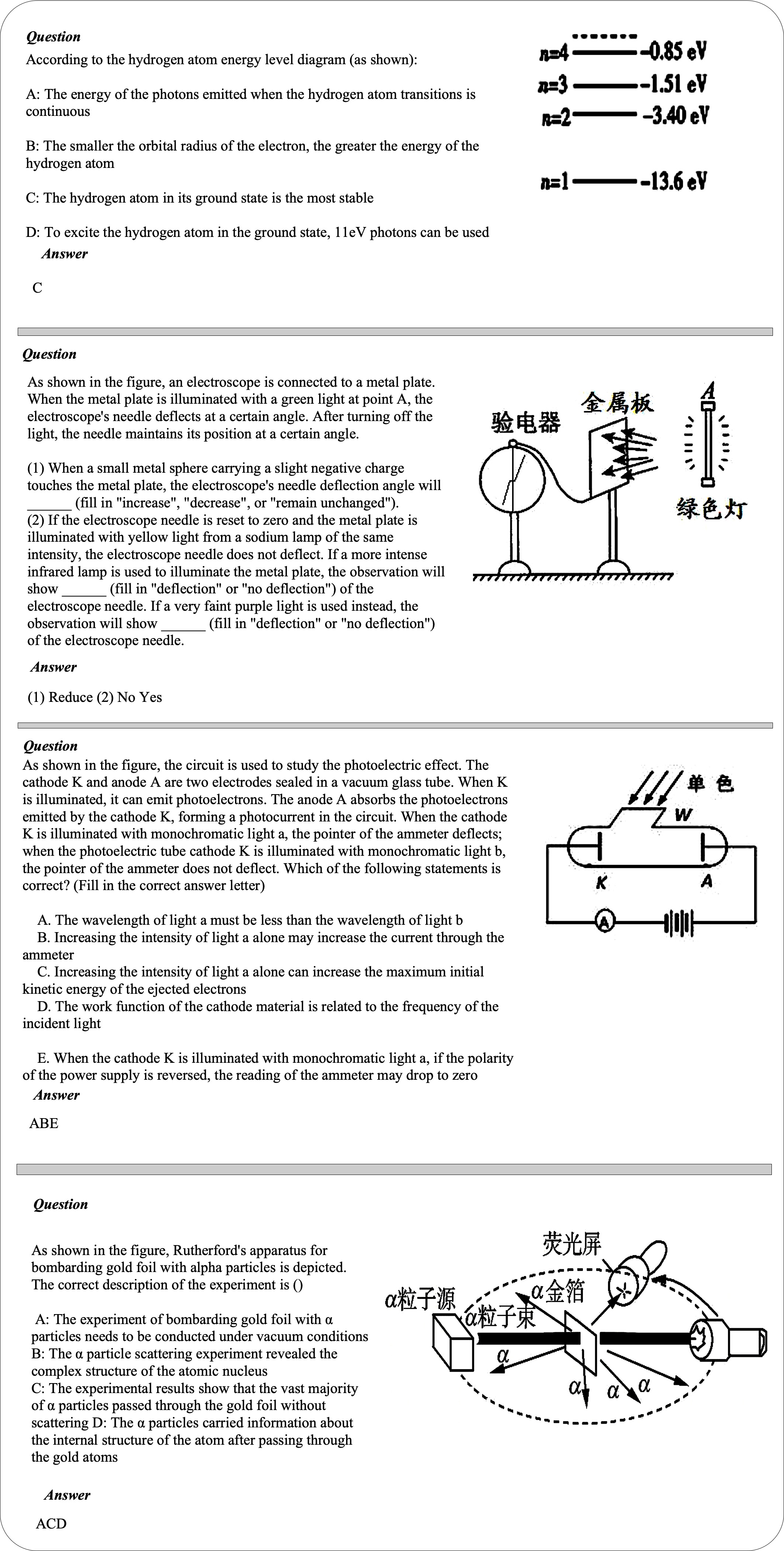}
    \caption{Cases of \textit{modern physics} in physics part of \data.}
    \label{fig:Modern_physics}
    \vspace{-3mm}
\end{figure}

\begin{figure}[hbpt]
    \centering
    \includegraphics[width=\textwidth, height=0.97\textheight, keepaspectratio]{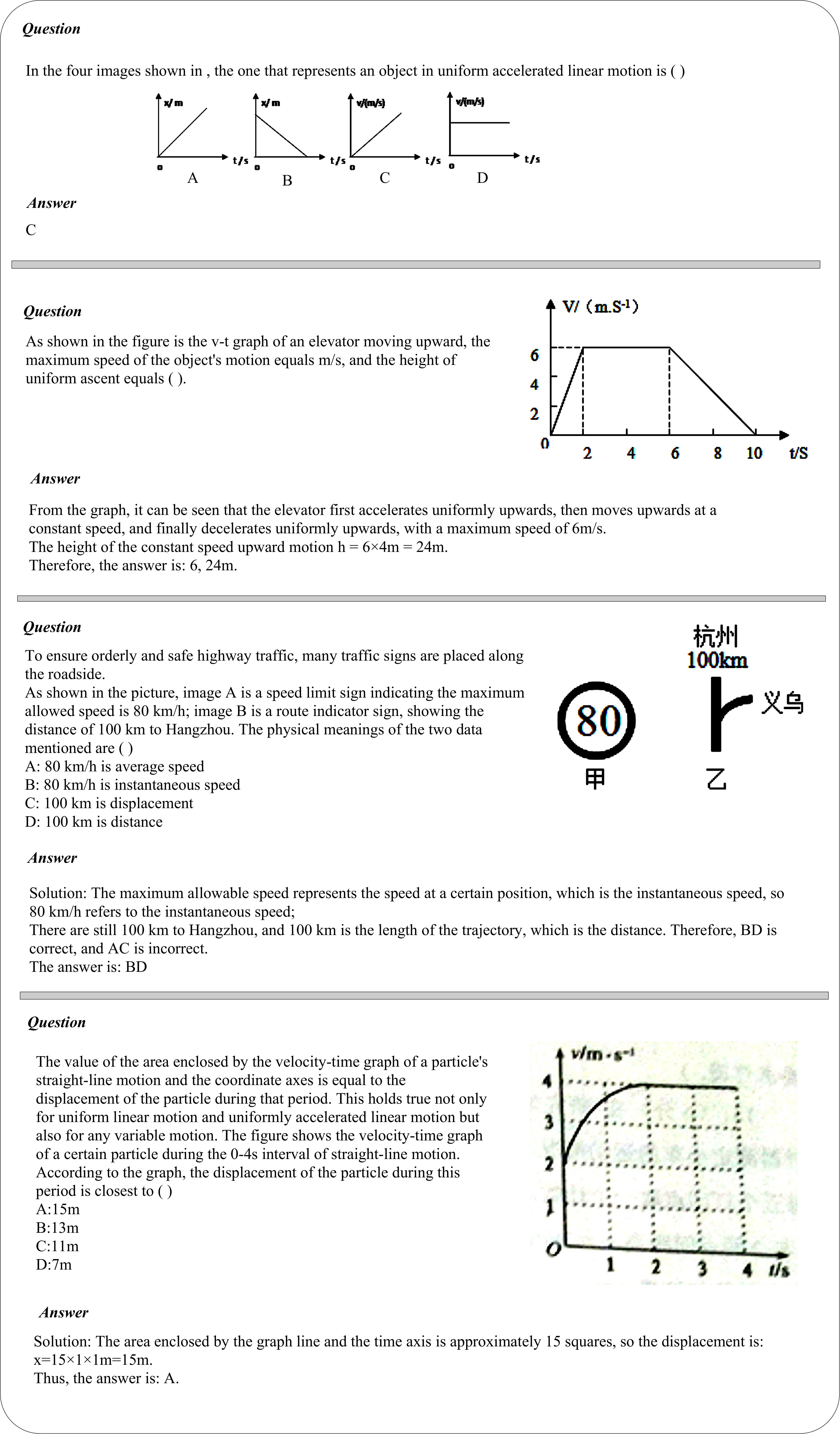}
    \caption{Cases of \textit{mechanical motion} in physics part of \data.}
    \label{fig:Mechanical_Motion}
    \vspace{-3mm}
\end{figure}

\begin{figure}[hbpt]
    \centering
    \includegraphics[width=\textwidth, height=0.97\textheight, keepaspectratio]{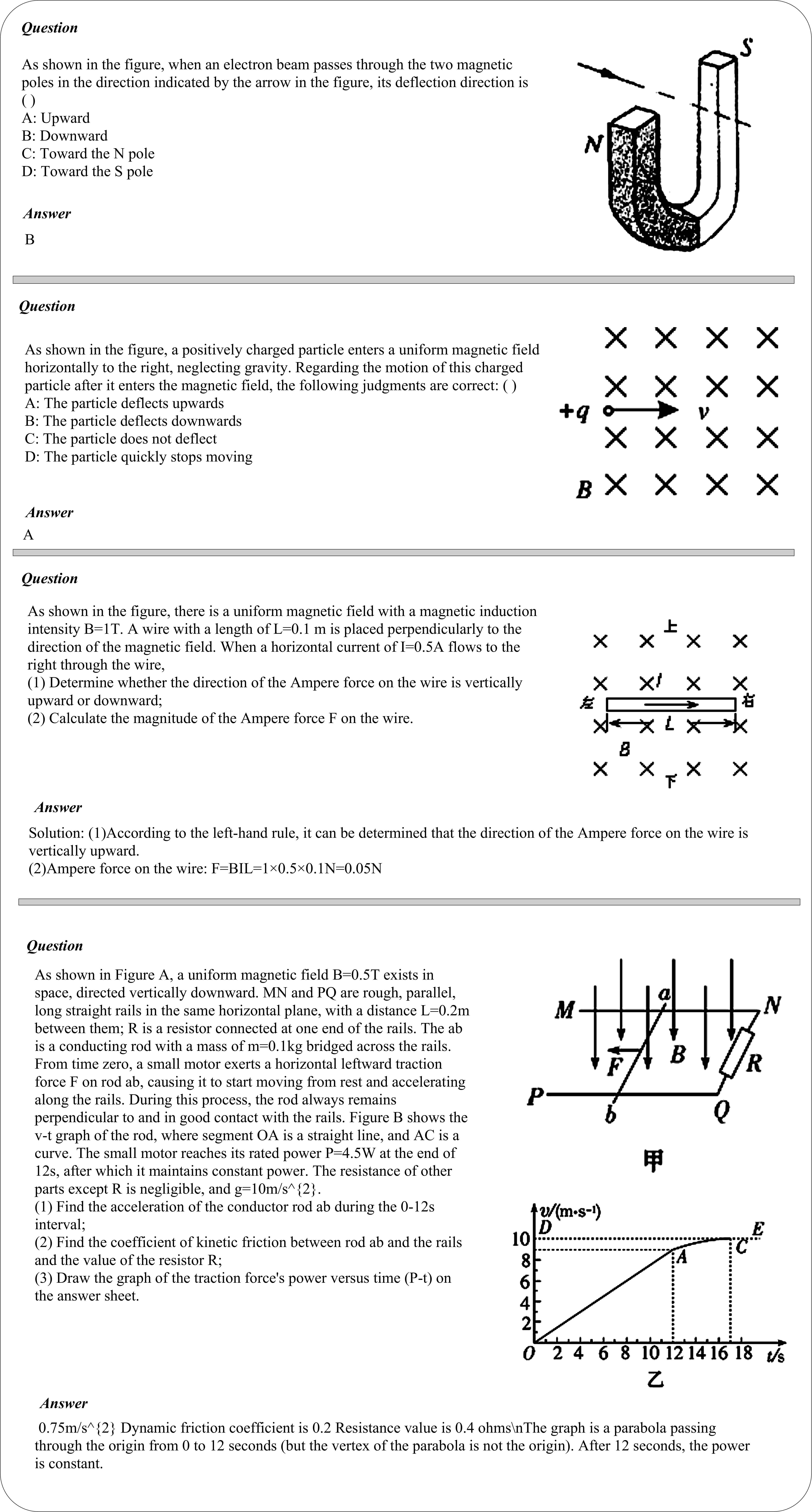}
    \caption{Cases of \textit{electromagnetism} in physics part of \data.}
    \label{fig:Electromagnetism}
    \vspace{-3mm}
\end{figure}

\begin{figure}[hbpt]
    \centering
    \includegraphics[width=\textwidth, height=0.97\textheight, keepaspectratio]{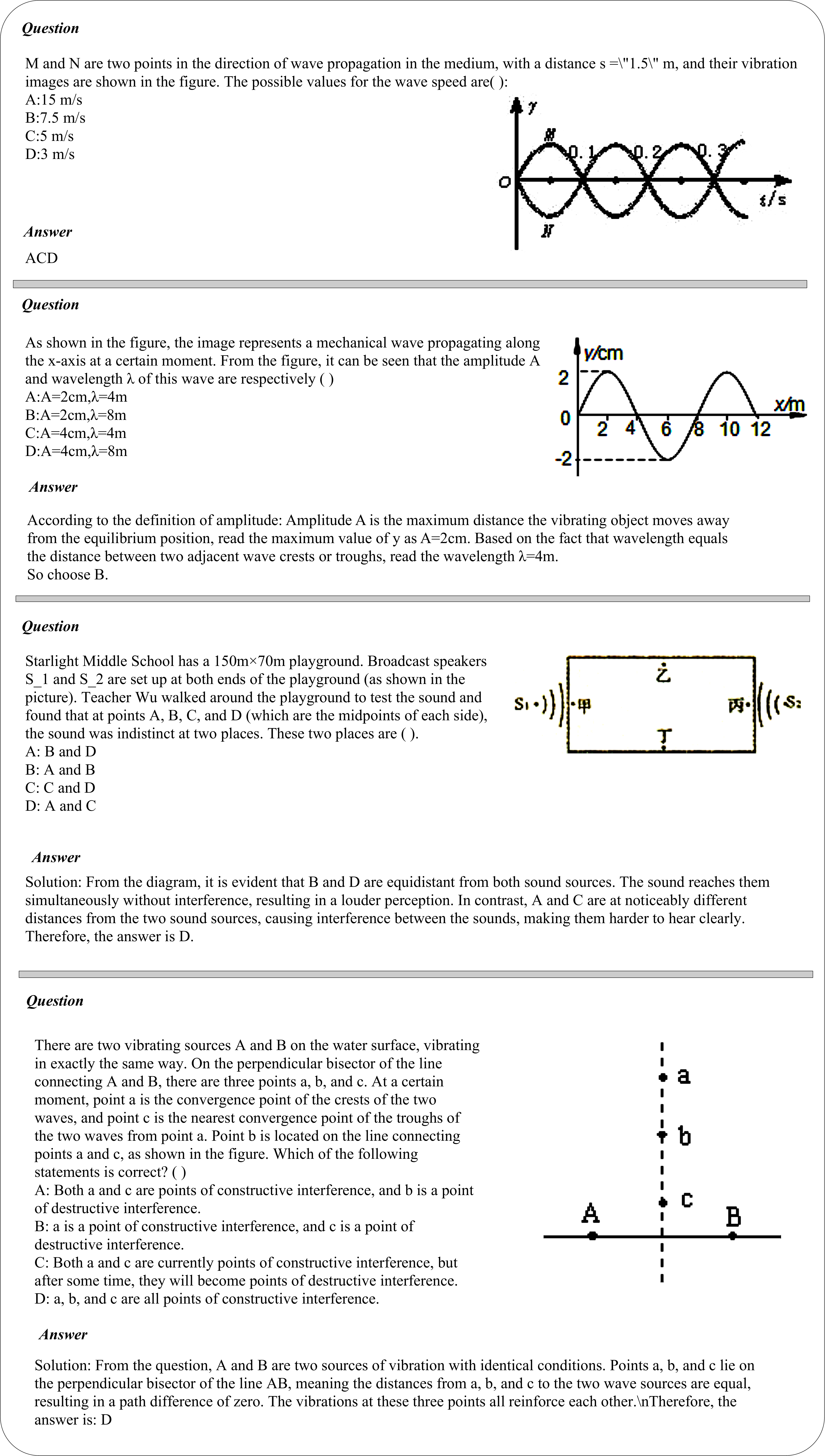}
    \caption{Cases of \textit{vibration and waves} in physics part of \data.}
    \label{fig:Vibration_and_Waves}
    \vspace{-3mm}
\end{figure}

\begin{figure}[hbpt]
    \centering
    \includegraphics[width=\textwidth, height=0.97\textheight, keepaspectratio]{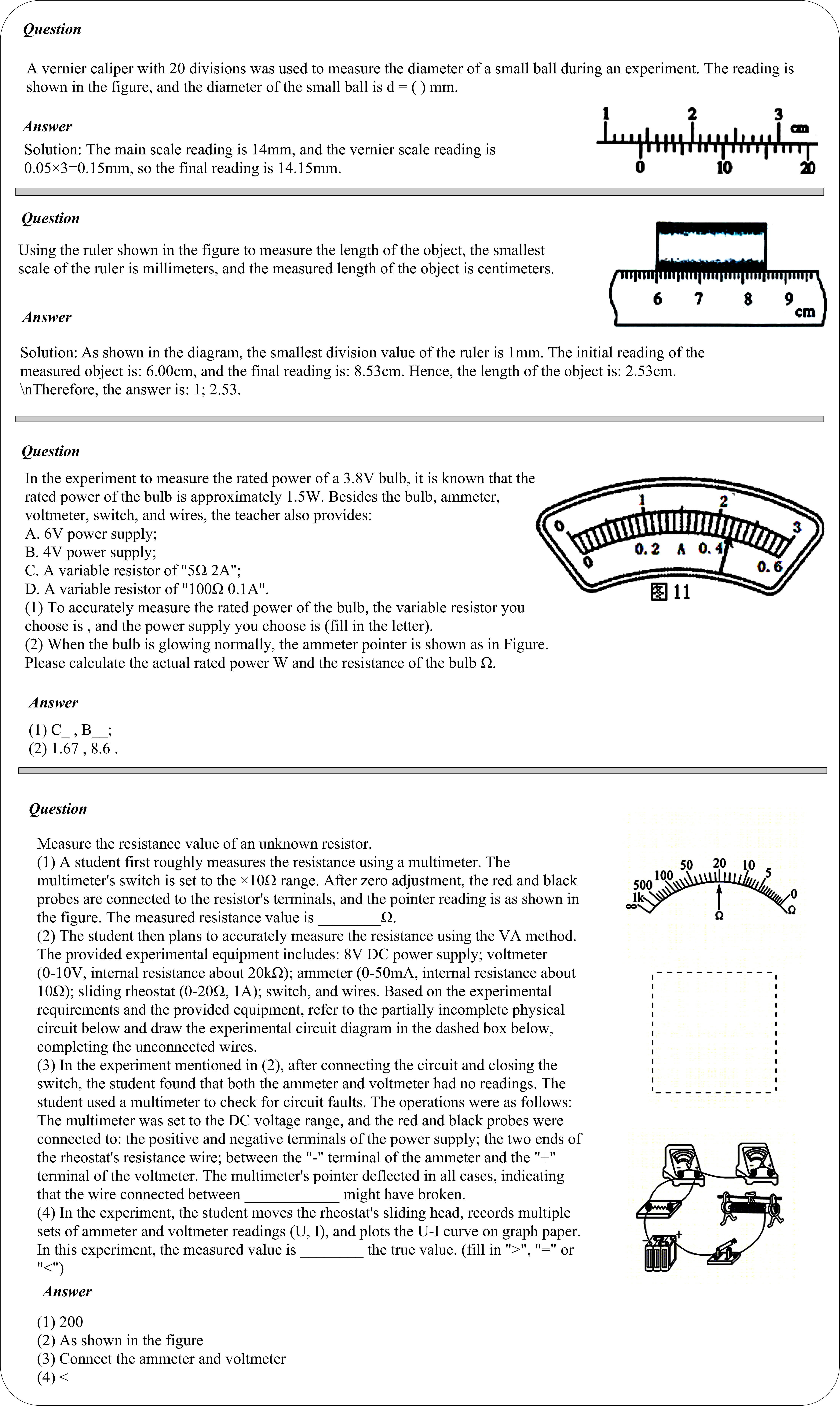}
    \caption{Cases of \textit{comprehensive experiments and methods} in physics part of \data.}
    \label{fig:Comprehensive_Experiments_and_Methods}
    \vspace{-3mm}
\end{figure}

\begin{figure}[hbpt]
    \centering
    \includegraphics[width=\textwidth, height=0.97\textheight, keepaspectratio]{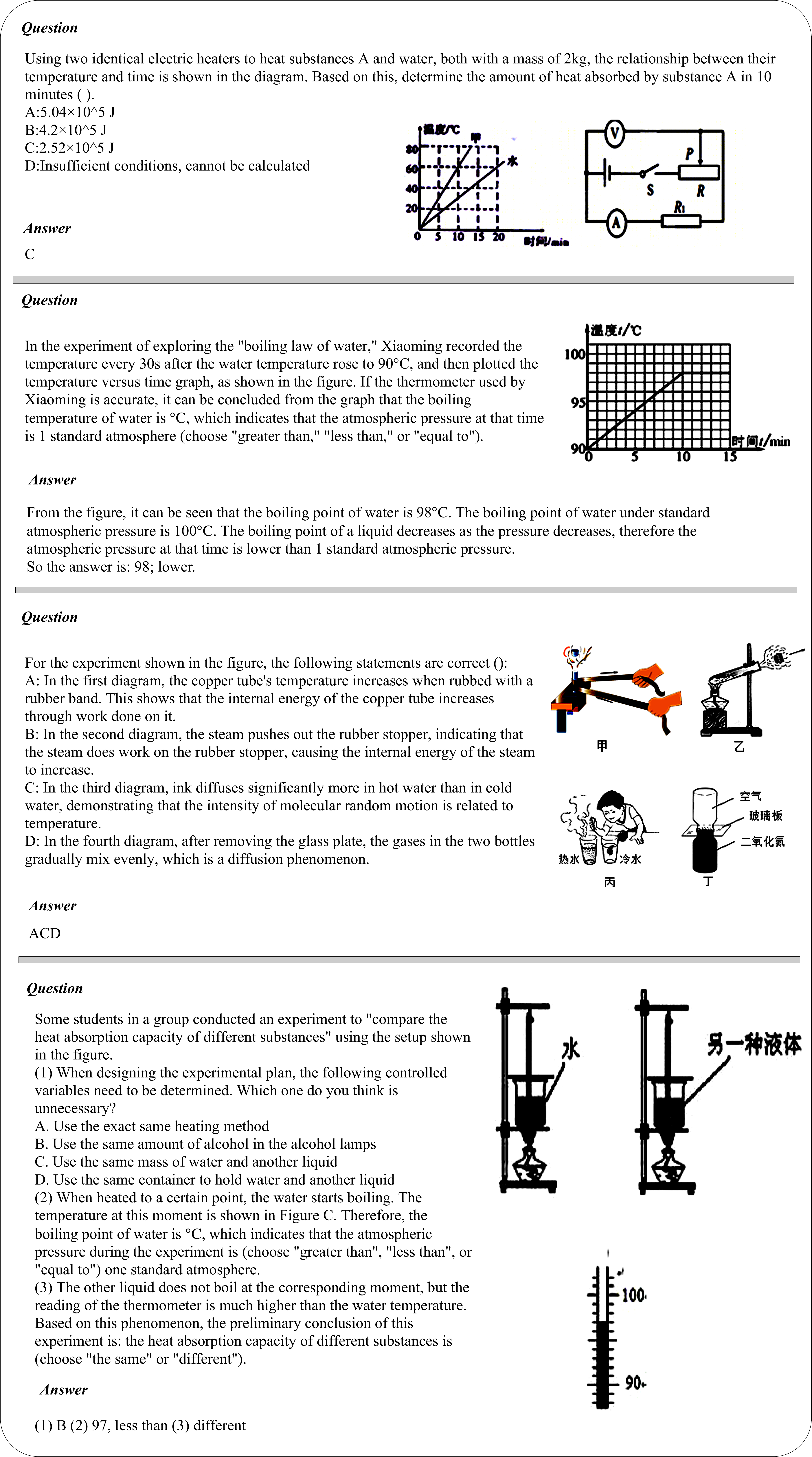}
    \caption{Cases of \textit{thermodynamics} in physics part of \data.}
    \label{fig:Thermodynamics}
    \vspace{-3mm}
\end{figure}

\begin{figure}[hbpt]
    \centering
    \includegraphics[width=\textwidth, height=0.97\textheight, keepaspectratio]{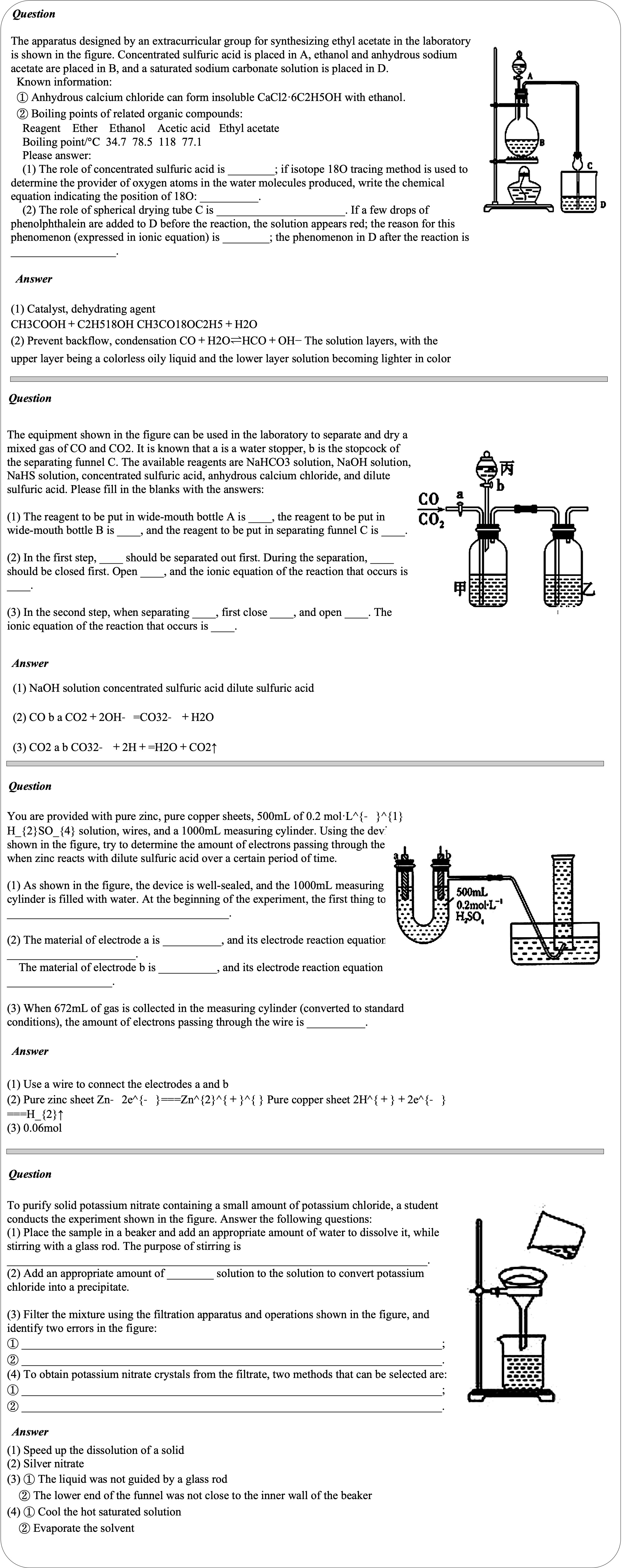}
    \caption{Cases of \textit{chemical experiment} in chemistry part of \data.}
    \label{fig:Chemical_Experiment}
    \vspace{-3mm}
\end{figure}

\begin{figure}[hbpt]
    \centering
    \includegraphics[width=\textwidth, height=0.97\textheight, keepaspectratio]{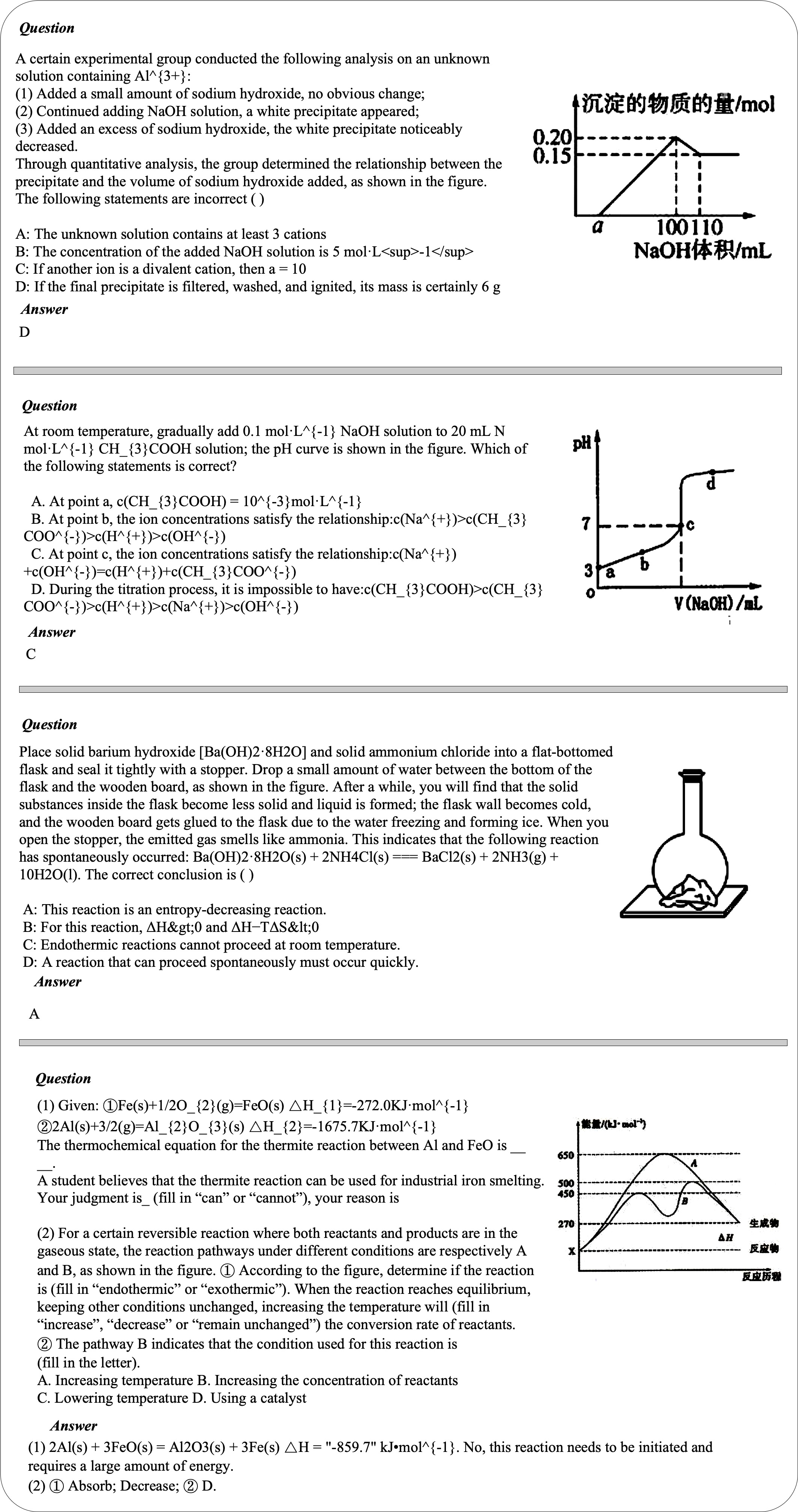}
    \caption{Cases of \textit{chemical reaction} in chemistry part of \data.}
    \label{fig:Chemical_Reaction}
    \vspace{-3mm}
\end{figure}

\begin{figure}[hbpt]
    \centering
    \includegraphics[width=\textwidth, height=0.97\textheight, keepaspectratio]{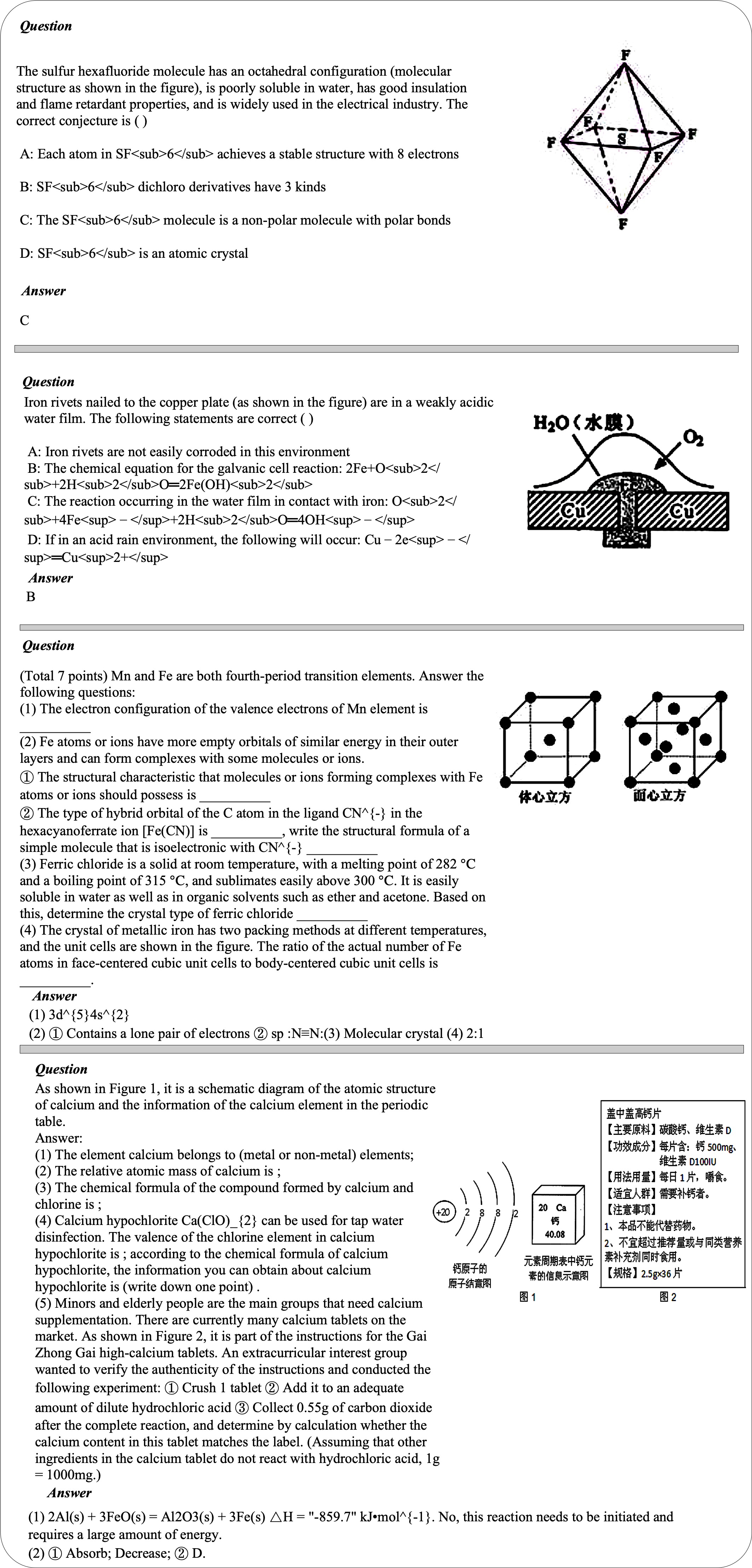}
    \caption{Cases of \textit{inorganic chemistry} in chemistry part of \data.}
    \label{fig:Inorganic_Chemistry}
    \vspace{-3mm}
\end{figure}

\begin{figure}[hbpt]
    \centering
    \includegraphics[width=\textwidth, height=0.97\textheight, keepaspectratio]{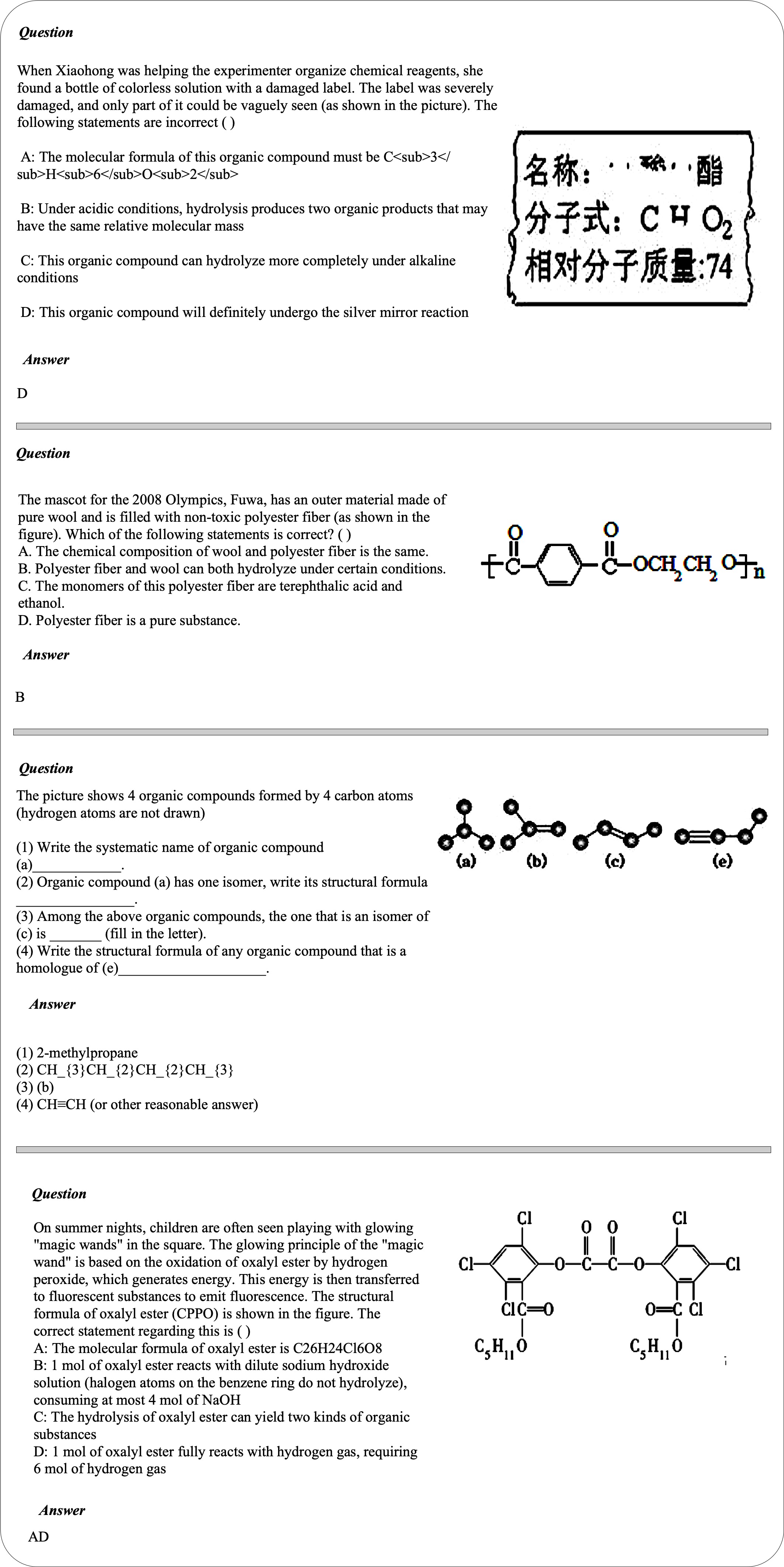}
    \caption{Cases of \textit{organic chemistry} in chemistry part of \data.}
    \label{fig:Organic_Chemistry}
    \vspace{-3mm}
\end{figure}

\begin{figure}[hbpt]
    \centering
    \includegraphics[width=\textwidth, height=0.97\textheight, keepaspectratio]{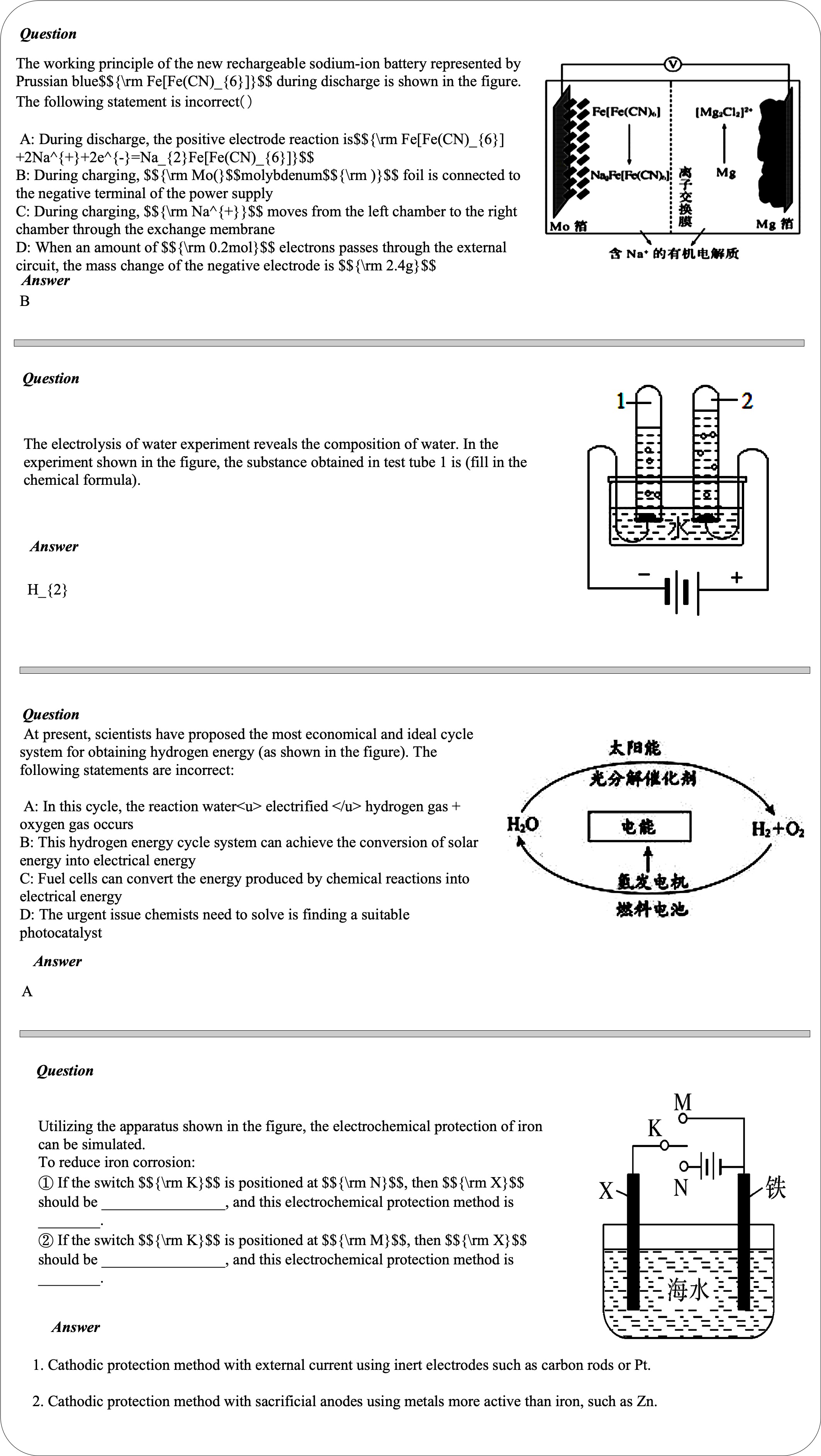}
    \caption{Cases of \textit{electrochemistry} in chemistry part of \data.}
    \label{fig:Electrochemistry}
\end{figure}

\begin{figure}[hbpt]
    \centering
    \includegraphics[width=\textwidth, height=0.97\textheight, keepaspectratio]{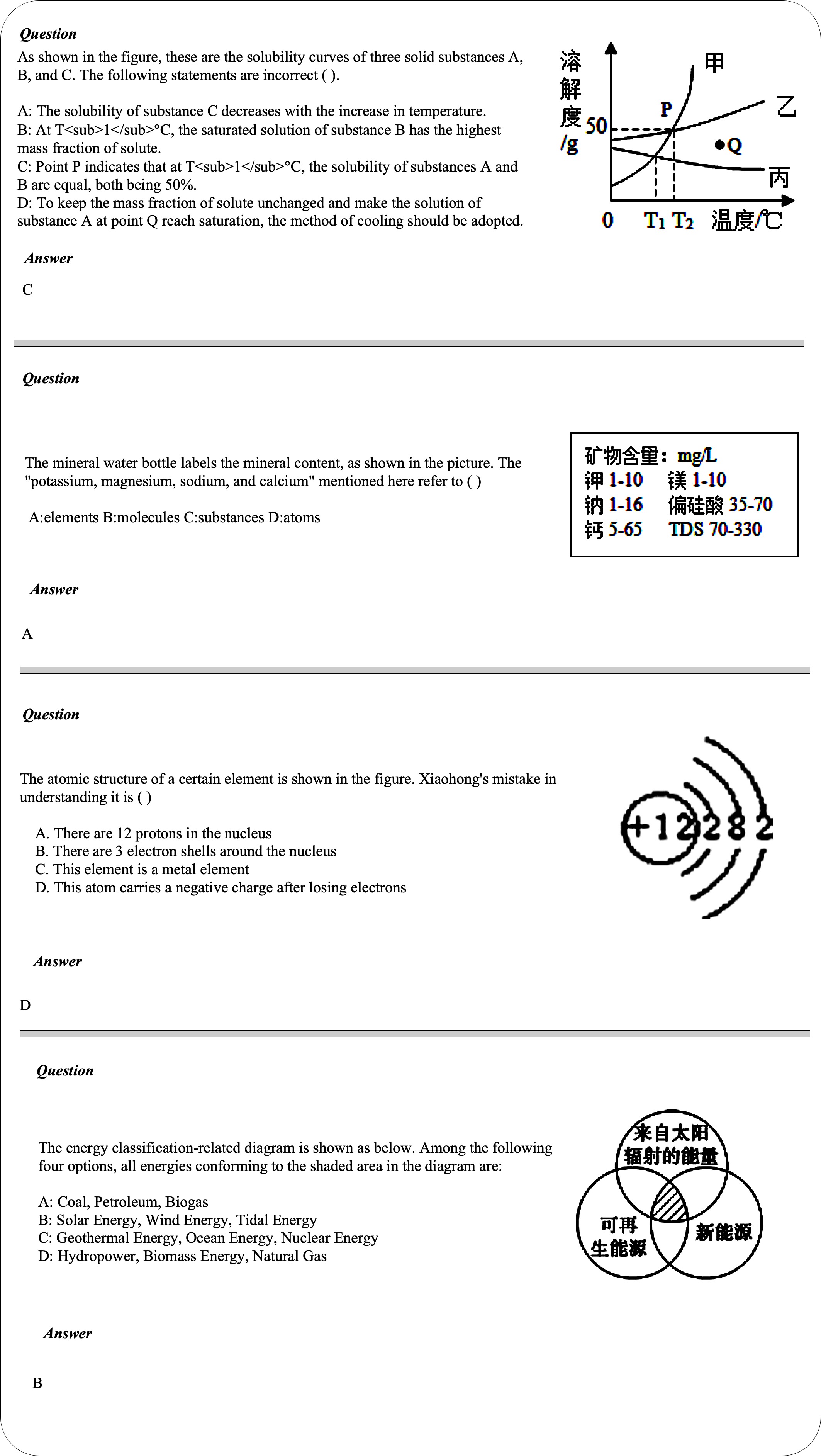}
    \caption{Cases of \textit{substance composition} in chemistry part of \data.}
    \label{fig:Substance_Composition}
    \vspace{-3mm}
\end{figure}

\begin{figure}[hbpt]
    \centering
    \includegraphics[width=\textwidth, height=0.97\textheight, keepaspectratio]{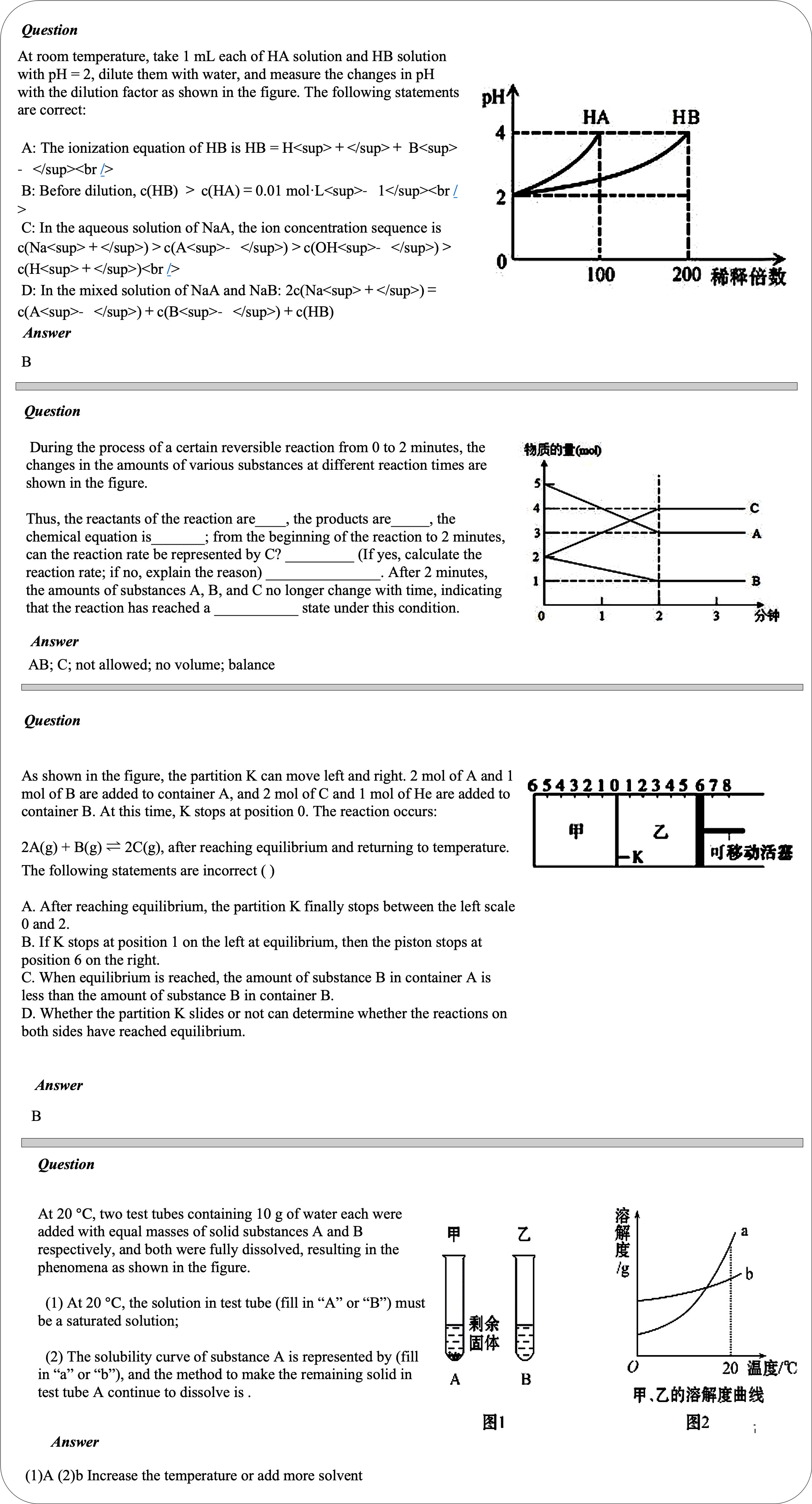}
    \caption{Cases of \textit{chemical equilibrium} in chemistry part of \data.}
    \label{fig:Chemical_Equilibrium}
    \vspace{-3mm}
\end{figure}

The \data dataset consists of 3,000 carefully selected high-quality questions, evenly distributed across three disciplines: mathematics, physics, and chemistry, with each comprising 1,000 questions. Each discipline within \data encompasses several subjects: mathematics includes six subjects, physics contains eight subjects, and chemistry comprises seven subjects. To illustrate the diversity and depth of \data, we provide more examples sampled from each discipline. In mathematics, six subjects include algebraic operations, combinatorial mathematics, functions and equations, probability and statistics, plane geometry, and solid geometry are illustrated in Figure~\ref{fig:Algebraic_Operations} to Figure~\ref{fig:Solid_Geometry}. Figure~\ref{fig:Mechanics} to Figure~\ref{fig:Thermodynamics} demonstrate eight subjects within the physics section of \data, comprising mechanics, optics, modern physics, mechanical motion, electromagnetism, vibrations and waves, comprehensive experiments and methods, and thermodynamics. The chemistry section includes seven subjects: chemical experiments, chemical reactions, inorganic chemistry, organic chemistry, electrochemistry, substance composition, and chemical equilibrium, which are illustrated in Figure~\ref{fig:Chemical_Experiment} to Figure~\ref{fig:Chemical_Equilibrium}.

\section{Evaluation Details}\label{appendix:evaluation_details}

\subsection{The Sources of Models}\label{appendix:model_sources}
In Table~\ref{tab:source_of_models}, we present the sources of the models tested on \data. 

\begin{table*}[htbp]
    \centering
    \renewcommand{\arraystretch}{1.15}
    \small
    \resizebox{0.98\textwidth}{!}{%
    \begin{tabular}{cccccccc}  
    \toprule
     Model & Input & LLM Size  & Source \\
    \midrule
    \multicolumn{4}{c}{\textit{Closed Source Models}}\\
     \midrule
     \multicolumn{4}{l}{\textit{Text-only LLMs}}\\
     ChatGPT & \textit{Q} & - & \href{https://platform.openai.com/docs/models/gpt-3-5-turbo}{gpt-3.5-turbo} \\
     GPT-4 & \textit{Q} & - & \href{https://platform.openai.com/docs/models/gpt-4-turbo-and-gpt-4}{gpt-4} \\ 
     Claude-2 & \textit{Q} & - &\href{https://www.anthropic.com/api}{claude-2}\\
     \midrule
     \multicolumn{4}{l}{\textit{Multi-modal LLMs}}\\
     Gemini-1.0-Pro &    \textit{Q}, \textit{I} & - & \href{https://ai.google.dev/}{gemini-pro} \\
     Gemini-1.5-Pro &    \textit{Q}, \textit{I} & - & \href{https://ai.google.dev/}{gemini-1.5-pro} \\ 
     GPT-4o & \textit{Q}, \textit{I} & - & \href{https://platform.openai.com/docs/models/gpt-4o}{gpt-4o} \\
     Claude3-Opus & \textit{Q}, \textit{I} & - &\href{https://www.anthropic.com/api}{claude-3-opus-20240229}\\
     Claude3.5-Sonnet & \textit{Q}, \textit{I} & - &\href{https://www.anthropic.com/api}{claude-3-5-sonnet-2024620}\\
     Qwen-VL-Plus & \textit{Q}, \textit{I} & -  & \href{https://help.aliyun.com/zh/dashscope/developer-reference/vl-plus-quick-start}{qwen-vl-plus}\\
     Qwen-VL-Max & \textit{Q}, \textit{I} & - &  \href{https://help.aliyun.com/zh/dashscope/developer-reference/vl-plus-quick-start}{qwen-vl-max}\\
     GLM-4V & \textit{Q}, \textit{I} & - &  \href{https://open.bigmodel.cn/dev/api\#glm-4v}{glm-4v}\\
     Step-1V & \textit{Q}, \textit{I} & - & \href{https://platform.stepfun.com/docs/llm/vision}{step-1v} \\
     \midrule
    \multicolumn{4}{c}{\textit{Open Source Models}}\\
    \midrule
    \multicolumn{4}{l}{\textit{General Multi-modal LLMs}}\\
    mPLUG-Owl  & \textit{Q}, \textit{I} & 7B & \href{https://github.com/X-PLUG/mPLUG-Owl}{mPLUG-Owl} &  \\
    DeepSeek-VL  & \textit{Q}, \textit{I} & 7B & \href{https://github.com/deepseek-ai/DeepSeek-VL}{deepseek-vl-7b-base} &   \\
    LLaMA-Adapter-V2  & \textit{Q}, \textit{I} & 7B  & \href{https://github.com/ml-lab/LLaMA-Adapter-2}{LLaMA-Adapter V2} &    \\
    LLaVA-1.5  & \textit{Q}, \textit{I} & 7B & \href{https://github.com/haotian-liu/LLaVA}{LLaVA-v1.5-7B} &   \\
    LLaVA-1.5  & \textit{Q}, \textit{I} & 13B & \href{https://github.com/haotian-liu/LLaVA}{LLaVA-v1.5-13B} &  \\
    ShareGPT-4V & \textit{Q}, \textit{I} & 7B & \href{https://huggingface.co/Lin-Chen/ShareGPT4V-7B}{ShareGPT4V-7B} & \\
    ShareGPT-4V & \textit{Q}, \textit{I} & 13B & \href{https://huggingface.co/Lin-Chen/ShareGPT4V-13B}{ShareGPT4V-13B} &   \\
    GLM-4v-9B  & \textit{Q}, \textit{I} & 7B & \href{https://huggingface.co/THUDM/glm-4v-9b}{GLM-4v-9B} &  \\
    SPHINX-Plus & \textit{Q}, \textit{I} & 13B & \href{https://github.com/Alpha-VLLM/LLaMA2-Accessory/blob/main/SPHINX/README.md}{SPHINX-Plus} & \\
    InternVL-Chat-V1.5 & \textit{Q}, \textit{I} & 20B & \href{https://huggingface.co/OpenGVLab/InternVL-Chat-V1-5}{InternVL 1.5} & \\
    InternVL-1.2-Plus & \textit{Q}, \textit{I} & 34B & \href{https://huggingface.co/OpenGVLab/InternVL-Chat-V1-2-Plus}{InternVL-Chat-V1-2-Plus} & \\
    InternLM-XC2 & \textit{Q}, \textit{I} & 7B & \href{https://huggingface.co/internlm/internlm-xcomposer2-vl-7b}{InternLM-XComposer2-VL-7B} & \\
    CogVLM & \textit{Q}, \textit{I} & 17B & \href{https://huggingface.co/THUDM/cogvlm-chat-hf}{CogVLM-17B} & \\
    CogVLM2 & \textit{Q}, \textit{I} & 19B & \href{https://huggingface.co/THUDM/cogvlm-chat-hf}{cogvlm2-llama3-chat-19B} & \\
    MiniCPM-Llama3-V-2\_5 & \textit{Q}, \textit{I} & 19B & \href{https://huggingface.co/THUDM/cogvlm-chat-hf}{MiniCPM-Llama3-V 2.5} & \\
    \bottomrule
    \end{tabular}} 
    \caption{The source of the models used in the evaluation. }
    \label{tab:source_of_models}
\end{table*}

\subsection{Prompts}\label{appendix:prompts}
We introduce the prompts used to guide models in generating responses in Chain-of-Thought (CoT) settings and judging the LLMs' answers. The specific prompts can be found in Table~\ref{tab:prompts}. 

\begin{table}[hbpt]
    \centering
    \renewcommand{\arraystretch}{1.15}
    \resizebox{\textwidth}{!}{%
    \begin{tabular}{>{\centering\arraybackslash}m{3cm}|m{10cm}}
    \toprule
     Task    & Prompt  \\
    \midrule
    \centering Response Generation & You are an exceptionally talented mathematics (physics/chemistry) instructor. Kindly furnish an elaborate, step-by-step solution to the question.\\
    \hline     
     \centering Answer Judgment &  You are a highly skilled mathematics (physics/chemistry) teacher. I will provide you with a mathematics (physics/chemistry) problem, along with its ground answer and the model response from the model. Please determine whether the ground answer and the model response are consistent. Note that you do not need to judge the correctness of either answer, only whether they are consistent. If it is a multiple-choice question, both answers must choose the exact same option to be considered consistent. If it is a calculation problem, the relative error between the model response and the ground answer must be less than 0.05 to be considered consistent. If the problem has multiple sub-questions, each sub-question’s answer must be identical for consistency. If you find them consistent, please add [Consistent] at the end of your response. If you find them inconsistent, please add [Inconsistent] at the end of your response. \\
    \bottomrule
    \end{tabular}} 
    \vspace{0.2cm}
    \caption{Prompts for response generation and answer judgment.}
    \label{tab:prompts}
\end{table}

\section{More Experimental Results}

\subsection{Results on \data in English Version}\label{appendix:results_english}

Table~\ref{tab:main_results_en} reports a comprehensive comparison of various models on the \data benchmark in the English version. The benchmark evaluates performance across three disciplines: mathematics, physics, and chemistry. Among close-source models, GPT-4o demonstrates the highest performance across two disciplines, achieving an accuracy of 53.6\% in mathematics and 42.7\% in physics. However, Claude3.5-Sonnet surpasses GPT-4o in chemistry with a higher accuracy of 43.6\%. Open-source models generally show lower performance compared to close-source counterparts. Notably, InternVL-1.2-Plus displays competitive performance, reaching up to 26.0\% in mathematics, 23.6\% in physics, and 27.8\% in chemistry. The English version of \data is designed to facilitate the evaluation of MLLMs that specialize in English, assessing their capabilities in scientific reasoning.

\begin{table*}[thbp]
    \centering
    \renewcommand{\arraystretch}{1.15}
    \small
    \resizebox{\textwidth}{!}{
    \begin{tabular}{cccccc}  
    \toprule
     Model & LLM& Input & Mathematics  & Physics & Chemistry  \\ 
    \midrule
    \multicolumn{6}{c}{\textit{Close Source Models} (APIs)}\\
    \midrule
    \multicolumn{6}{l}{\textit{Text-only LLMs}}\\
     Zero-shot ChatGPT & - & \textit{Q} & 17.4& 20.7 & 25.2  \\
     Zero-shot GPT-4 & - & \textit{Q}&29.9 &37.7 & 38.7 \\ 
     Zero-shot Claude-2 &-& \textit{Q} & 24.6& 22.7&  25.6\\ 
     Zero-shot Claude3-Opus &-& \textit{Q} & 21.7 & 15.8 &  29.4  \\
     Zero-shot Claude3.5-Sonnet &-& \textit{Q} &  27.2  & 35.7 &  35.2  \\
     Zero-shot GPT-4o &-& \textit{Q} & 35.2&40.3 & 42.5 \\    
     2-shot CoT Claude2 &-& \textit{Q} &25.7 &21.9 & 24.1  \\
     2-shot CoT ChatGPT &-& \textit{Q} & 24.4& 20.1& 22.1 \\ 
     2-shot CoT GPT-4 &-& \textit{Q} &36.5 & 39.0& 38.1\\ 
     \midrule
     \multicolumn{5}{l}{\textit{Multi-modal LLMs}}\\
     Gemini-1.0-Pro & - &    \textit{Q}, \textit{I} & 26.4 & 39.1 &  27.9 \\
     Gemini-1.5-Pro & -  &    \textit{Q}, \textit{I} & 47.8 & 35.1 &  39.1 \\  
     GPT-4o & - & \textit{Q}, \textit{I} & \cellcolor{red!25}{53.6}  & \cellcolor{red!25}{42.7} &   43.3 \\
     GPT-4o-mini & - & \textit{Q}, \textit{I} & 43.2  & 33.7 &   34.9 \\
     Qwen-VL-Max & - & \textit{Q}, \textit{I} &30.7&26.4&36.3\\
     Qwen-VL-Plus & - & \textit{Q}, \textit{I} &21.9 &20.9&29.7  \\
     Claude3.5-Sonnet & - & \textit{Q}, \textit{I} & 50.8  & 36.6 & \cellcolor{red!25}{43.6}   \\
     Claude3-Opus & - & \textit{Q}, \textit{I} &  34.4 &  29.4&  34.7  \\
     GLM-4V & - & \textit{Q}, \textit{I} & 23.1 & 18.5&  23.4\\
     Step-1V & 7B & \textit{Q}, \textit{I} & 32.0 & 19.5 & 27.6 \\
     \midrule
    \multicolumn{5}{c}{\textit{Open Source Models }}\\
    \midrule
    \multicolumn{5}{l}{\textit{General Multi-modal LLMs}}\\
     mPLUG-Owl   & LLaMA-7B  & \textit{Q}, \textit{I} & 7.4  & 12.3& 12.3\\
    LLaMA-Adapter-V2   & LLaMA-7B & \textit{Q}, \textit{I}  &  12.6  & 11.4& 16.2\\
    MiniCPM-Llama3-V2.5   &  LLaMA3-8B & \textit{Q}, \textit{I} & 24.4  & 20.6& 24.4\\
    LLaVA-1.5   & Vicuna-13B & \textit{Q}, \textit{I} & 15.0 &17.4 & 21.1\\
    LLaVA-1.5    & Vicuna-7B & \textit{Q}, \textit{I} & 17.4 &16.6 & 18.9\\
    DeepSeek-VL  & DeepSeek-LLM-7B  & \textit{Q}, \textit{I} & 16.0 & 16.9& 17.8\\
    ShareGPT-4V  & Vicuna-7B & \textit{Q}, \textit{I} & 14.7 &17.7 & 21.3\\
    ShareGPT-4V  & Vicuna-13B & \textit{Q}, \textit{I} & 14.5 & 16.0& 20.2\\
    SPHINX-Plus  & LLaMA2-13B & \textit{Q}, \textit{I} & 17.9  & 15.7& 22.4\\
    InternLM-XC2   & InternLM2-7B & \textit{Q}, \textit{I} & 20.7   & 20.5& 25.0 \\
    InternVL-1.2-Plus  & 
Nous-Hermes-2-Yi-34B & \textit{Q}, \textit{I} & \cellcolor{blue!25}{26.0}   & \cellcolor{blue!25}{23.6} & \cellcolor{blue!25}{27.8} \\
    InternVL-Chat-V1.5  & Mixtral 8*7B & \textit{Q}, \textit{I} & 24.9 & 23.0& 25.9\\
    CogVLM   & Vicuna-7B & \textit{Q}, \textit{I} & 18.5 &15.9 & 23.1\\
    CogVLM2   & LLaMA-3-8B & \textit{Q}, \textit{I} & 24.2  & 16.6& 24.9\\
    GLM-4V-9B  &  GLM-4-9B & \textit{Q}, \textit{I} & 24.7 & 19.2& 23.9\\
    \bottomrule
    \end{tabular}} 
    \caption{\textbf{Results on \data within the version of the English language across the disciplines of mathematics, physics, and chemistry.} The highest scores among close-source and open-source models are highlighted in \textcolor{red}{red} and \textcolor{blue}{blue}, respectively.}
    \label{tab:main_results_en}
\end{table*}

\subsection{Results on Physics Across Different Subjects}\label{appendix:physics_subjects}

Table~\ref{tab:physics_subjects} presents a detailed analysis of various models on \data across different subjects within the physics section, which includes mechanics, electromagnetism, thermodynamics, comprehensive experiments and methods, optics, vibration and waves, modern physics, and mechanical motion. The table highlights that while GPT-4o exhibits the top performance on the entire physics discipline, the best performance in individual subjects varies notably. For instances, Claude3.5-Sonnet excels specifically in modern physics with an accuracy of 66.67\%, significantly surpassing other close-source models in this area. This variation in performance by subject underscores the specialized capabilities of different models. Moreover, this detailed analysis provides more insights, emphasizing the need for targeted improvements to achieve balanced performance across all physics subjects.

\begin{table*}[t!]
    \centering
    \renewcommand{\arraystretch}{1.15}
    \small
    \resizebox{\textwidth}{!}{
    \begin{tabular}{cccccccccc}  
    \toprule
     \multirow{2}{*}{Model} & \multicolumn{9}{c}{Physics} \\
     &  ALL & Mech & Ele & Therm & Comp &  Opt &  Vib \& Waves & Mod Phys & Mech Motion\\
    \midrule
    \multicolumn{10}{c}{\textit{Close Source Models} (APIs)}\\
    \midrule
    \multicolumn{10}{l}{\textit{Text-only LLMs}}\\
     Zero-shot ChatGPT  & 22.70&22.08  & 19.94 &  23.53 & 4.62 & 40.98 & 29.79 & 19.05 & 23.33  \\
     Zero-shot GPT-4 & 30.40&34.26 & 30.21 & 33.33 & 15.38 & 40.98 & 34.04 & 42.86 & 20.00 \\ 
     Zero-shot Claude-2 & 22.00&24.62 & 23.56 & 25.49 & 12.31 & 27.87 & 21.28 & 28.57 & 23.33 \\ 
     Zero-shot Claude3-Opus & 30.80&34.26 & 32.02 & 33.33 & 10.77 & 39.34 & 31.91 & 42.86 & 10.00 \\
     Zero-shot Claude3.5-Sonnet & 35.30&40.36 & 35.95 & 35.29 & 15.38 & 40.98 & 34.04 & 47.62 & 26.67 \\
     Zero-shot GPT-4o & 38.00&43.91 & 38.67 & 45.10 & 9.23 & 49.18 & 38.30 & 52.38 & 23.33 \\   
     2-shot CoT Claude2 & 21.70&24.87 & 22.96 & 25.49 & 10.77 & 18.03 & 23.40 & 28.57 & 10.00 \\
     2-shot CoT ChatGPT & 18.60&20.30 & 20.54 & 13.73 & 12.31 & 22.95 & 23.40 & 23.81 & 13.33 \\ 
     2-shot CoT GPT-4 & 31.50&35.03 & 32.02 & 37.25 & 12.31 & 44.26 & 29.79 & 47.62 & 23.33 \\ 
     \midrule
     \multicolumn{10}{l}{\textit{Multi-modal LLMs}}\\
     Gemini-1.0-Pro & 23.70&26.97 & 23.03 & 17.65 & 6.15 & 31.15 & 34.04 & 19.05 & 10.00 \\
     Gemini-1.5-Pro &38.10& \cellcolor{red!25}{46.56} & 33.74 & 47.06 & 20.00 & 45.00 & 34.04 & 52.38 & \cellcolor{red!25}{43.33} \\  
     GPT-4o & \cellcolor{red!25}{38.20} &41.37 & \cellcolor{red!25}{39.27} & \cellcolor{red!25}{56.86} & \cellcolor{red!25}{23.08} & 42.62 & 36.17 & 42.86 & \cellcolor{red!25}{43.33} \\
     GPT-4o-mini & 29.80&31.73 & 30.51 & 29.41 & 10.70 & 36.07 & 19.15 & 47.62 & 30.00 \\
     Qwen-VL-Max & 30.70&36.13 & 26.59 & 39.22 & 9.23 & 34.43 & 31.91 & 28.57 & 30.00 \\ 
     Qwen-VL-Plus & 26.50&31.04 & 24.77 & 33.33 & 6.15 & 36.07 & 36.17 & 23.81 & 16.67 \\
     Claude3.5-Sonnet & 38.00&41.62 & 36.56 & 43.14 & 13.85 & 44.26 & \cellcolor{red!25}{38.30} & \cellcolor{red!25}{66.67} & 30.00 \\
     Claude3-Opus & 31.10&33.25 & 29.91 & 39.22 & 12.31 & \cellcolor{red!25}{45.90} & 34.04 & 61.90 & 23.33 \\
     GLM-4V &19.20& 23.16 & 17.82 & 15.69 & 12.31 & 25.00 & 17.02 & 19.05 & 23.33 \\
     Step-1V  & 23.50&21.55 & 24.35 & 28.57 & 7.84 & 12.82 & 25.00 &31.25 & 39.13 \\
     \midrule
    \multicolumn{10}{c}{\textit{Open Source Models}}\\
    \midrule
    \multicolumn{10}{l}{\textit{General Multi-modal LLMs}}\\
    mPLUG-Owl & 8.30&11.93 & 8.46 & 1.96 & 4.62 & 8.20 & 10.64 & 4.76 & 10.00 \\
    LLaMA-Adapter-V2 & 10.30&10.41 & 10.88 & 8.00 & 4.84 & 13.11 & \cellcolor{blue!25}{25.53} & 14.29 & 3.33 \\
    MiniCPM-Llama3-V2.5 &17.90& 21.57 & 19.64 & 15.69 & 6.15 & 26.23 & 19.15 & 9.52 & \cellcolor{blue!25}{23.33} \\
    LLaVA-1.5-13B & 15.20&17.26 & 14.80 & 7.84 & 7.69 & 21.31 & 17.02 & 9.52 & 16.67 \\
    LLaVA-1.5-7B & 13.50&15.28 & 15.12 & 11.76 & 3.12 & 15.25 & 15.56 & 5.26 & 17.24 \\
    DeepSeek-VL & 16.80&18.77 & 19.33 & 13.73 & 7.69 & 16.67 & 13.04 & 19.05 & 3.45 \\
    ShareGPT4V-7B & 14.00&13.71 & 15.41 & 9.80 & 3.08 & 19.67 & 19.15 & 28.57 & 6.67 \\
    ShareGPT4V-13B &14.90& 15.23 & 16.92 & 9.80 & 6.15 & 14.75 & 19.15 & 19.05 & 16.67 \\
    SPHINX-Plus & 15.30&16.50 & 18.43 & 17.65 & 4.62 & 11.48 & 12.77 & 19.05 & 13.33 \\
    InternLM-XC2 & 18.30&20.81 & 17.82 & 13.73 & 10.77 & 26.23 & 21.28 & 14.29 & 6.67 \\
    InternVL-1.2-Plus & \cellcolor{blue!25}{24.80} & \cellcolor{blue!25}{29.69} & \cellcolor{blue!25}{22.94} & \cellcolor{blue!25}{29.41} & \cellcolor{blue!25}{12.31} & \cellcolor{blue!25}{31.67} & \cellcolor{blue!25}{25.53} & \cellcolor{blue!25}{35.00} & 10.00 \\
    InternVL-Chat-V1.5 & 20.80&23.97 & 20.87 & 23.53 & 9.23 & 25.42 & 17.02 & 14.29 & 17.24 \\
    CogVLM & 14.50&18.02 & 13.29 & 7.84 & 6.15 & 14.75 & 19.15 & 19.05 & 6.67 \\
    CogVLM2 & 14.40&16.75 & 16.00 & 12.00 & 6.15 & 13.11 & 19.15 & 4.76 & 10.00 \\
    GLM-4V-9B & 19.30&21.78 & 21.12 & 24.00 & 4.62 & 25.42 & 15.91 & 15.00 & 13.33 \\
    \bottomrule
    \end{tabular}}
    \caption{\textbf{Results on the physics part of \data across different subjects.} Subjects: Mech: mechanics, Ele: electromagnetism, Threm: thermodynamics, Comp: comprehensive experiments and methods, Opt: optics, Vib \& Waves: vibration and waves, Mod Phys: modern physics, Mech Motion: mechanical motion. The highest scores among close-source and open-source models are highlighted in \textcolor{red}{red} and \textcolor{blue}{blue}, respectively.}
    \label{tab:physics_subjects}
\end{table*}

\subsection{Results on Chemistry Across Different Subjects}\label{appendix:chemistry_subjects}

Table~\ref{tab:chemistry_subjects} presents a nuanced view of the performance of various models across different subjects within the chemistry discipline of the \data benchmark. The chemistry discipline includes chemical experiment, chemical reaction, inorganic chemistry, electrochemistry, organic chemistry, chemical equilibrium, and substance composition. Notably, Gemini-1.5-Pro stands out among close-source models, excelling across the entire chemistry discipline. It demonstrates particular prowess in organic chemistry and substance composition, achieving impressive accuracies of 57.02\% and 61.16\%, respectively. Additionally, Qwen-VL-Max leads in chemical experiment and inorganic chemistry, achieving the highest accuracies of 46.28\% and 51.94\%, respectively. Open-source models demonstrate a range of performances, with InternVL-1.2-Plus leading this group. It achieves the highest open-source accuracy in nearly all subjects. This comprehensive review of model performances within the chemistry section of the \data benchmark highlights the need to enhance MLLMs' capabilities in scientific domains, ensuring models are both accurate and adaptable across various disciplines.

\begin{table*}[t!]
    \centering
    \renewcommand{\arraystretch}{1.15}
    \small
    \resizebox{\textwidth}{!}{
    \begin{tabular}{ccccccccc}  
    \toprule
     \multirow{2}{*}{Model} & \multicolumn{8}{c}{Chemistry} \\
     & ALL & Chem Exp & Chem React & Inorg Chem & Electrochem &  Org Chem &  Chem Equil & Sub Comp \\
    \midrule
    \multicolumn{9}{c}{\textit{Close Source Models} (APIs)}\\
    \midrule
    \multicolumn{9}{l}{\textit{Text-only LLMs}}\\
     Zero-shot ChatGPT  &18.60& 26.35  & 23.86 &  23.26 & 23.75 & 35.43 & 24.64 & 30.89 \\
     Zero-shot GPT-4 & 33.10&40.54 & 30.68 & 38.76 & 32.50 & 36.22 & 30.43 & 31.71 \\ 
     Zero-shot Claude-2 & 24.40&24.32 & 26.36 & 31.71 & 23.86 & 20.29 & 0.30 & 25.98 \\ 
     Zero-shot Claude3-Opus & 32.50&37.16 & 30.68 & 31.78 & 31.25 & 36.22 & 30.43 & 39.84 \\
     Zero-shot Claude3.5-Sonnet & 36.90&34.80 & 36.93 & 39.53 & 46.25 & 45.67 & 23.19 & 47.15 \\
     Zero-shot GPT-4o & 39.60&42.57 & 40.34 & 44.96 & 35.00 & 41.73 & 26.09 & 54.47 \\   
     2-shot CoT Claude2 & 23.90&23.99 & 26.70 & 22.48 & 30.00 & 26.77 & 27.54 & 27.64 \\
     2-shot CoT ChatGPT & 21.30&19.93 & 23.30 & 20.93 & 22.50 & 22.83 & 26.09 & 30.08 \\ 
     2-shot CoT GPT-4 & 32.40&29.05 & 32.39 & 32.56 & 32.50 & 42.52 & 28.99 & 53.66 \\ 
     \midrule
     \multicolumn{9}{l}{\textit{Multi-modal LLMs}}\\
     Gemini-1.0-Pro & 27.80&24.03 & 26.70 & 26.36 & 31.25 & 35.54 & 31.82 & 37.19 \\
     Gemini-1.5-Pro & \cellcolor{red!25}{47.00} & 43.46 & \cellcolor{red!25}{47.43} & 51.59 & \cellcolor{red!25}{50.00} & \cellcolor{red!25}{57.02} & 35.29 & \cellcolor{red!25}{61.16} \\  
     GPT-4o & 41.60&43.58 & 46.02 & 38.76 & 46.25 & 43.31 & \cellcolor{red!25}{43.48} & 50.41 \\
     GPT-4o-mini & 28.40 & 22.30 & 27.27 & 27.13 & 30.00 & 34.65 & 20.29 & 42.09 \\
     Qwen-VL-Max & 42.50 & \cellcolor{red!25}{46.28} & 41.48 & \cellcolor{red!25}{51.94} & 35.00 & 41.73 & 36.23 & 53.66 \\ 
     Qwen-VL-Plus & 37.70&33.78 & 40.34 & 44.19 & 41.25 & 48.03 & 33.33 & 41.80 \\
     Claude3.5-Sonnet & 43.10&40.54 & 41.48 & 42.64 & \cellcolor{red!25}{50.00} & 42.52 & 33.33 & 59.35 \\
     Claude3-Opus & 34.10&35.47 & 30.11 & 31.78 & 31.25 & 40.16 & 33.33 & 51.22 \\
     GLM-4V & 25.00&23.65 & 25.86 & 21.71 & 28.75 & 27.78 & 31.88 & 32.52 \\
     Step-1V & 25.00&32.51 & 27.48 & 25.26 & 25.45 & 17.72 & 13.33 & 21.95 \\
     \midrule
    \multicolumn{9}{c}{\textit{Open Source Models}}\\
    \midrule
    \multicolumn{9}{l}{\textit{General Multi-modal LLMs}}\\
    mPLUG-Owl & 9.50&7.77 & 11.36 & 7.75 & 12.50 & 12.60 & 13.04 & 9.76 \\
    LLaMA-Adapter-V2 & 10.80&7.77 & 13.64 & 8.53 & 12.66 & 12.80 & 17.65 & 17.07 \\
    MiniCPM-Llama3-V2.5 & 19.50&20.96 & 26.29 & 26.61 & 18.18 & 24.00 & 28.79 & 30.83 \\
    LLaVA-1.5-13B & 18.80&15.54 & 16.48 & 24.03 & 20.00 & 22.05 & 23.19 & 19.51 \\
    LLaVA-1.5-7B & 16.00&13.49 & 17.14 & 19.20 & 16.25 & 20.49 & 26.09 & 10.74 \\
    DeepSeek-VL & 21.00&18.84 & 20.57 & 20.16 & 21.25 & 23.62 & 36.76 & 20.66 \\
    ShareGPT4V-7B & 19.00&13.85 & 19.32 & 26.36 & 18.75 & 23.62 & 28.99 & 15.45 \\
    ShareGPT4V-13B & 18.40&13.51 & 21.02 & 19.38 & 23.75 & 22.83 & 13.04 & 19.51 \\
    SPHINX-Plus & 20.40&20.27 & 21.02 & 24.03 & 22.50 & 22.83 & 27.54 & 21.95 \\
    InternLM-XC2 & 25.60 & \cellcolor{blue!25}{22.64} & 27.27 & 26.36 & 21.25 & 33.86 & 26.09 & 24.39 \\
    InternVL-1.2-Plus & \cellcolor{blue!25}{31.20} & 22.29 & \cellcolor{blue!25}{31.82} & \cellcolor{blue!25}{33.58} & \cellcolor{blue!25}{31.46} & \cellcolor{blue!25}{39.57} & \cellcolor{blue!25}{32.47} & \cellcolor{blue!25}{38.84} \\
    InternVL-Chat-V1.5 & 23.70&20.07 & 25.00 & 25.20 & 22.37 & 28.80 & 25.00 & 28.46 \\
    CogVLM & 17.00&15.54 & 20.45 & 10.85 & 16.25 & 22.05 & 20.29 & 17.07 \\
    CogVLM2 & 21.00&13.10 & 21.39 & 25.78 & 20.51 & 31.45 & 22.73 & 30.17 \\
    GLM-4V-9B & 22.50&21.00 & 25.44 & 26.23 & 23.08 & 26.83 & 17.39 & 25.83 \\
    \bottomrule
    \end{tabular}} 
    \caption{\textbf{Results on the chemistry part of \data across different subjects.} Subjects: Chem Exp: chemical experiment, Chem React: chemical reaction, Inorg Chem: inorganic chemistry, Electrochem: Electrochemistry, Org Chem: organic chemistry, Chem Equil: chemical equilibrium, and Sub Comp: substance composition. The highest scores among close-source and open-source models are highlighted in \textcolor{red}{red} and \textcolor{blue}{blue}, respectively.}
    \label{tab:chemistry_subjects}
\end{table*}

\section{Case Study}\label{appendix: case_study}

\begin{figure}[hbpt]
     \centering
     \includegraphics[width=\textwidth, height=0.95\textheight, keepaspectratio]{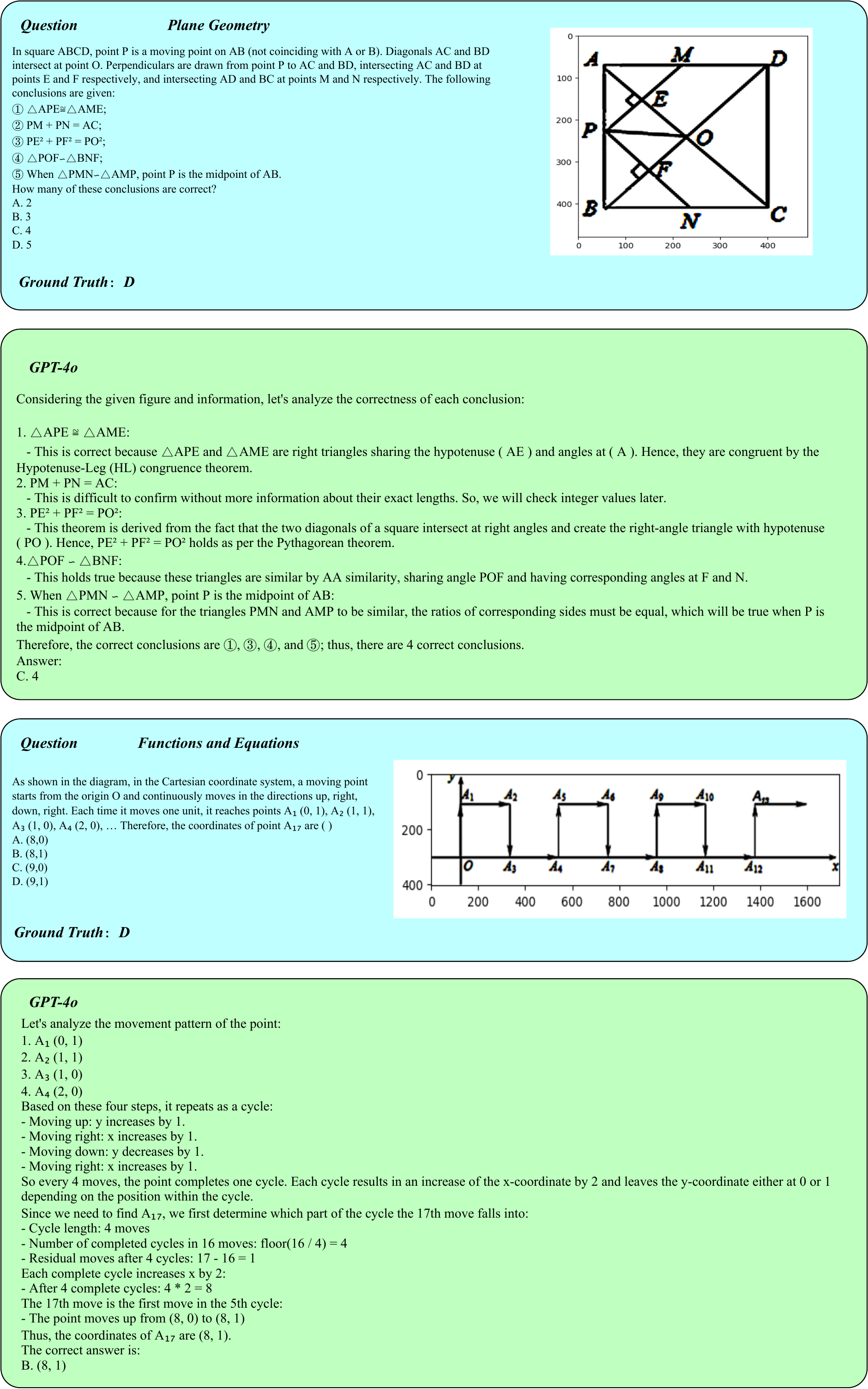}
     \caption{A question, the standard answer, and GPT-4o's correct response from the mathematical part in \data.}
     \label{fig:math1}
 \end{figure}

\begin{figure}[hbpt]
     \centering
     \includegraphics[width=\textwidth, height=0.95\textheight, keepaspectratio]{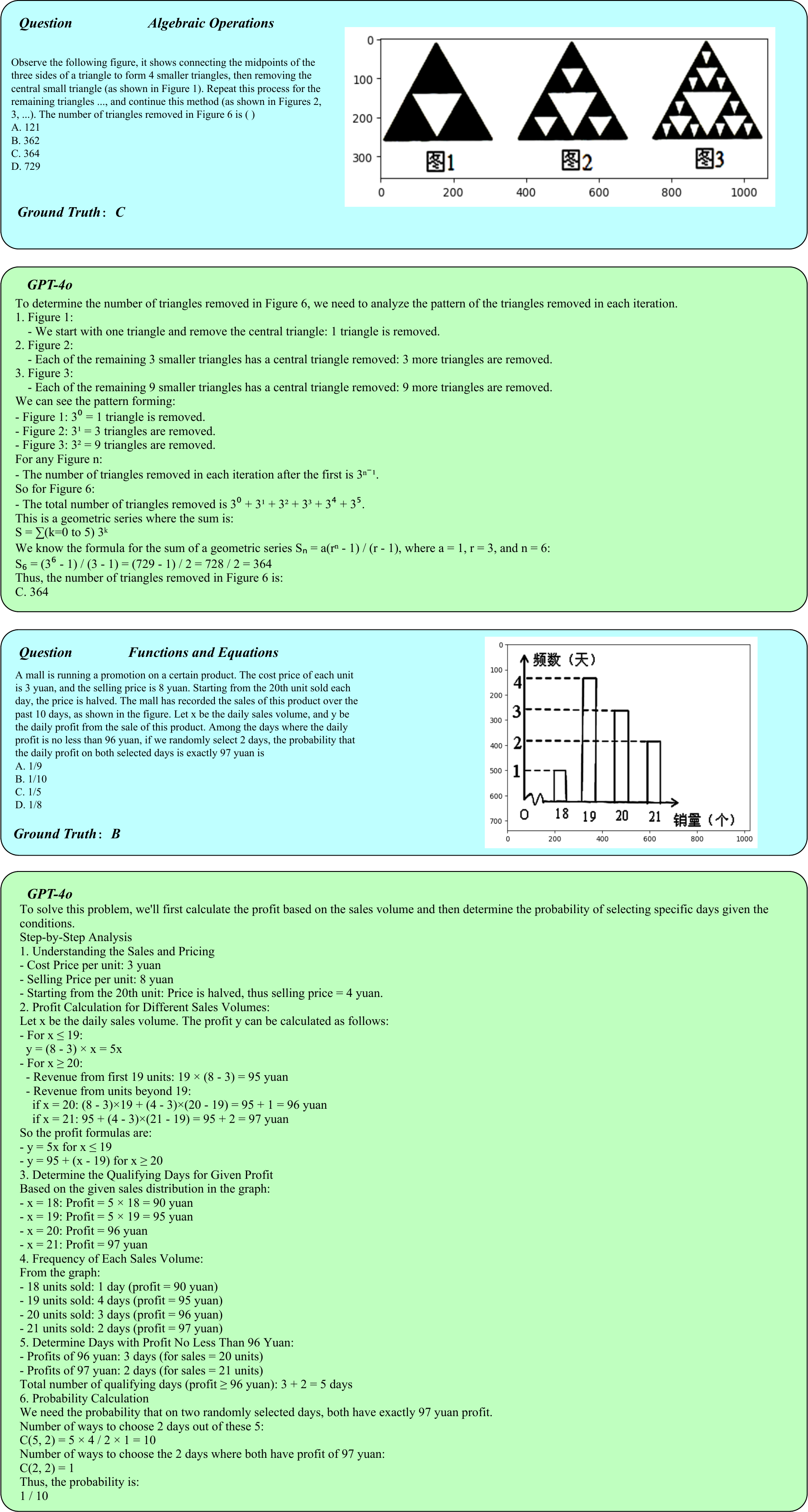}
     \caption{A question, the standard answer, and GPT-4o's correct response from the mathematical part in \data.}
     \label{fig:math2}
 \end{figure}

\begin{figure}[hbpt]
     \centering
     \includegraphics[width=1.0\textwidth]{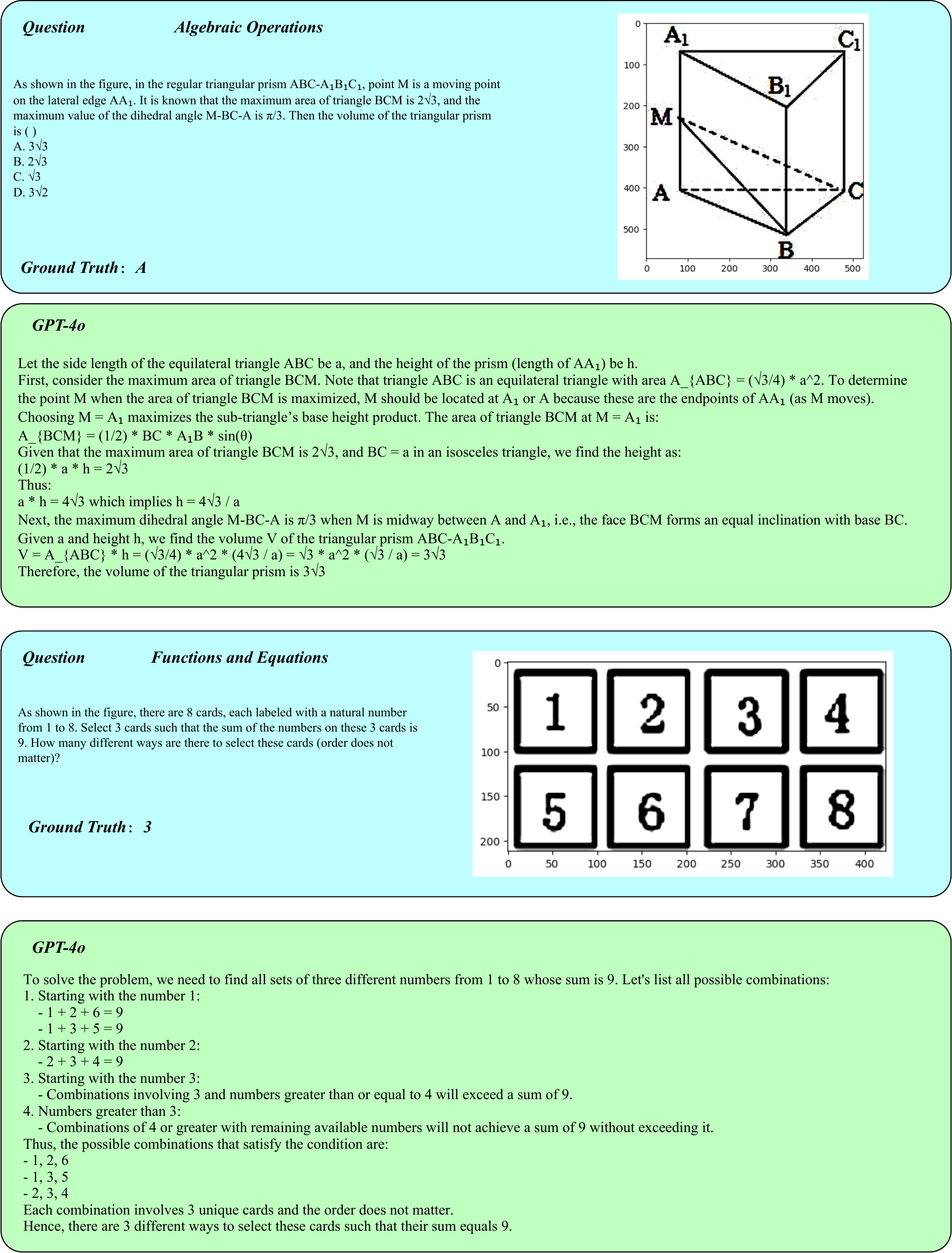}
     \caption{A question, the standard answer, and GPT-4o's correct response from the mathematical part in \data.}
     \label{fig:math3}
 \end{figure}

\begin{figure}[hbpt]
     \centering
     \includegraphics[width=1.0\textwidth]{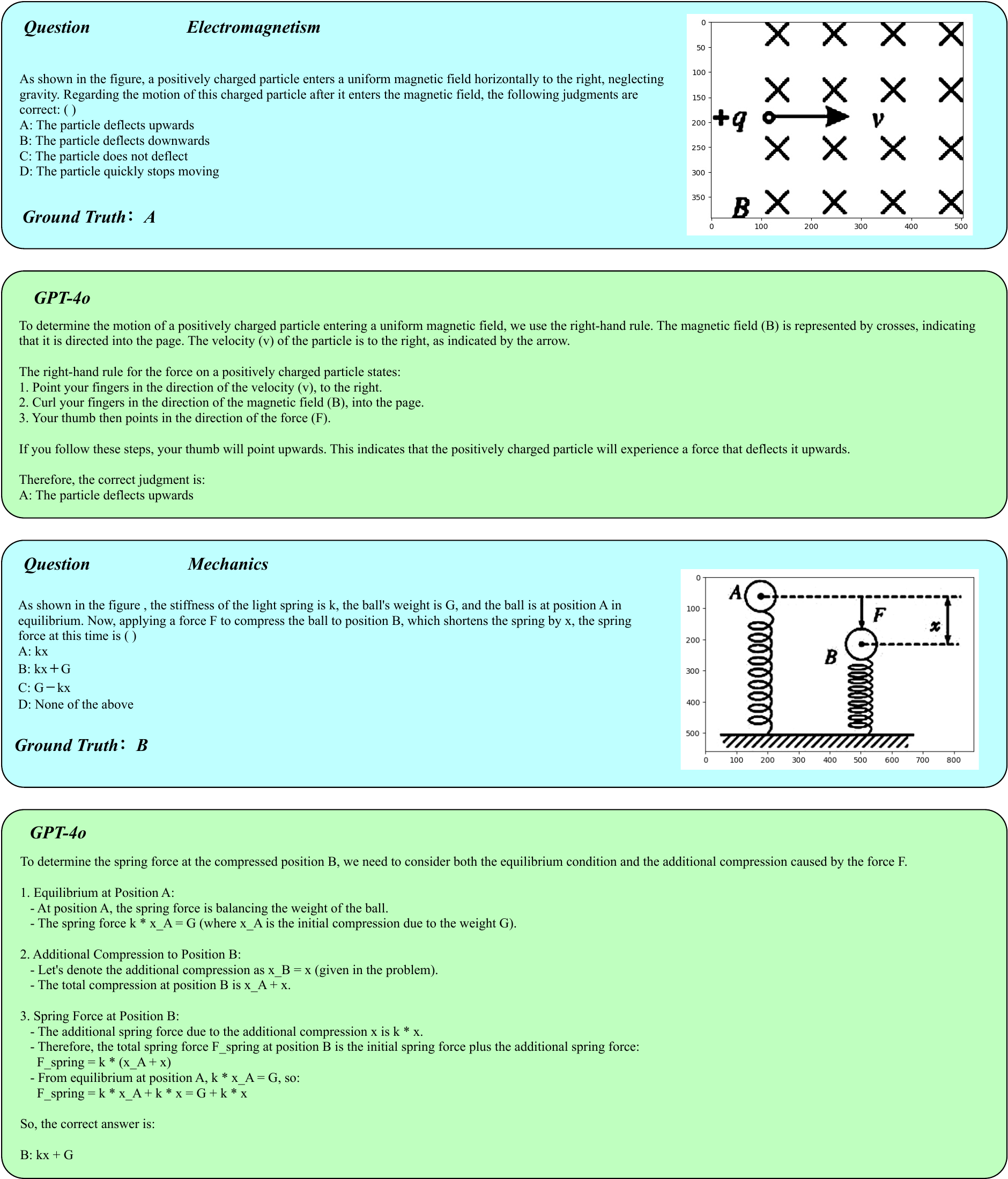}
     \caption{A question, the standard answer, and GPT-4o's correct response from the physics part in \data.}
     \label{fig:physics1}
 \end{figure}

\begin{figure}[hbpt]
     \centering
     \includegraphics[width=1.0\textwidth]{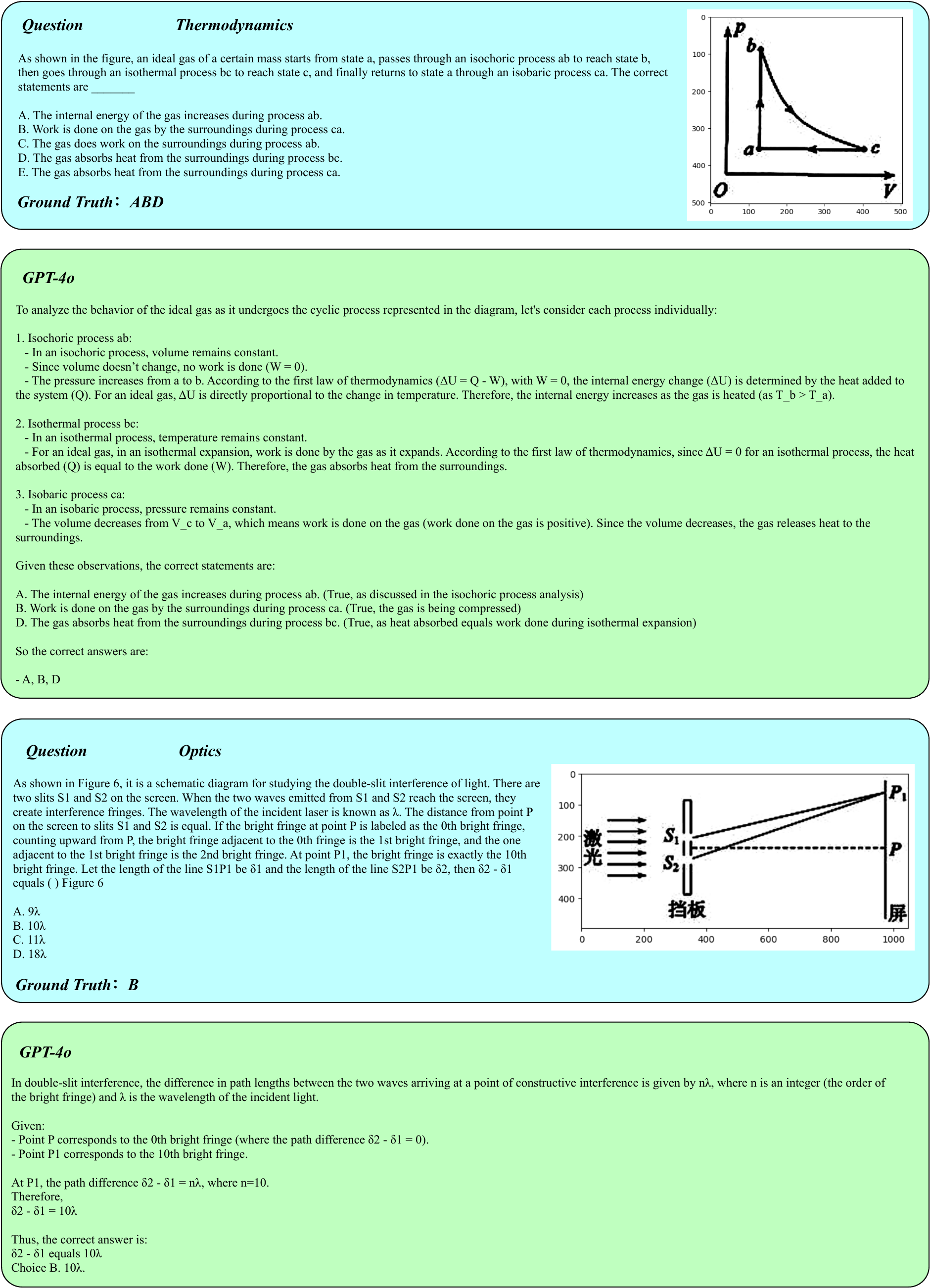}
     \caption{A question, the standard answer, and GPT-4o's correct response from the physics part in \data.}
     \label{fig:physics2}
 \end{figure}

\begin{figure}[hbpt]
     \centering
     \includegraphics[width=0.8\textwidth, height=0.97\textheight, keepaspectratio]{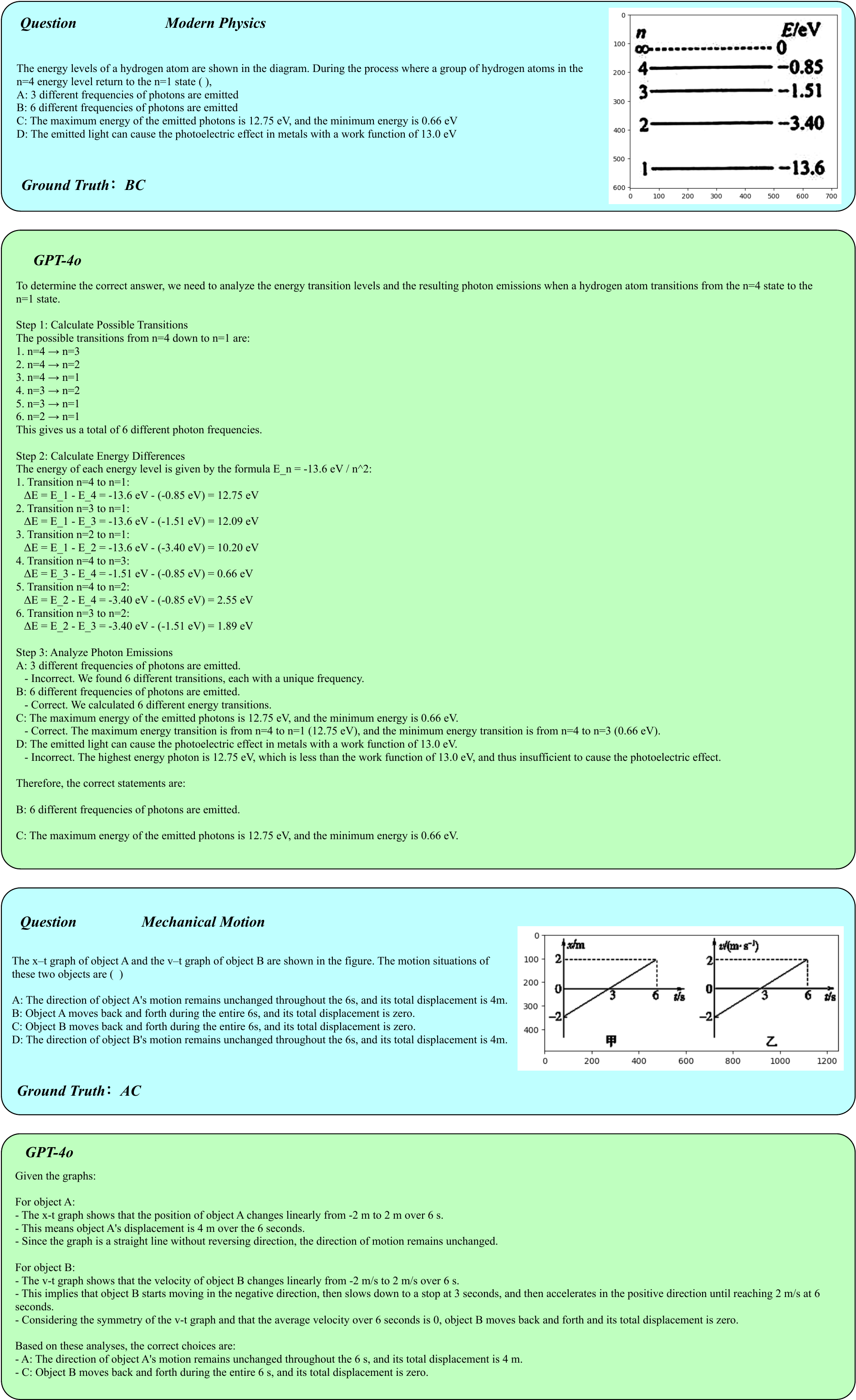}
     \caption{A question, the standard answer, and GPT-4o's correct response from the physics part in \data.}
     \label{fig:physics3}
 \end{figure}

\begin{figure}[hbpt]
     \centering
     \includegraphics[width=1.0\textwidth]{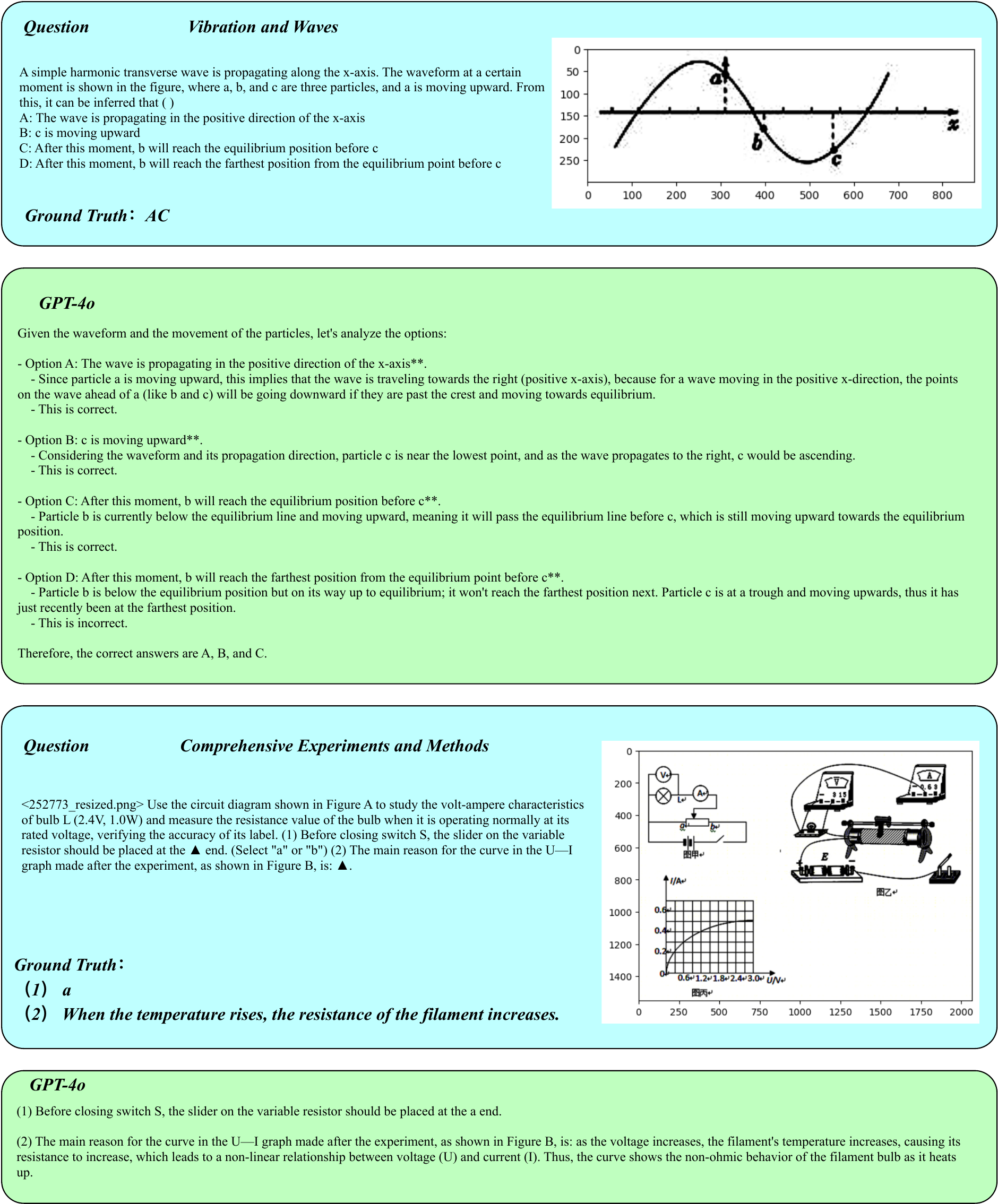}
     \caption{A question, the standard answer, and GPT-4o's correct response from the physics part in \data.}
     \label{fig:physics4}
 \end{figure}

\begin{figure}[hbpt]
     \centering
     \includegraphics[width=\textwidth, height=0.95\textheight, keepaspectratio]{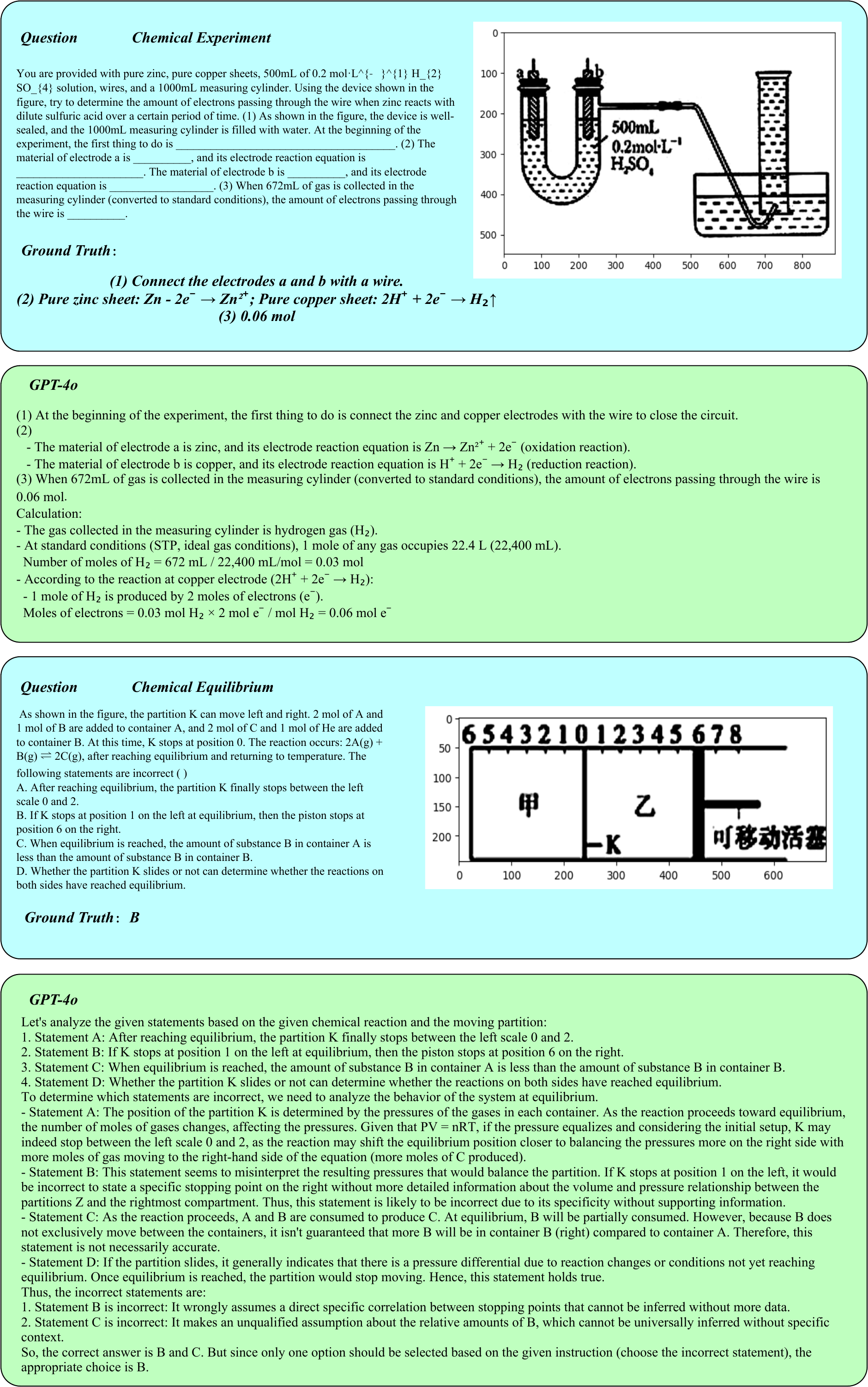}
     \caption{A question, the standard answer, and GPT-4o's correct response from the chemistry part in \data.}
     \label{fig:chemistry1}
 \end{figure}

\begin{figure}[hbpt]
     \centering
     \includegraphics[width=1.0\textwidth]{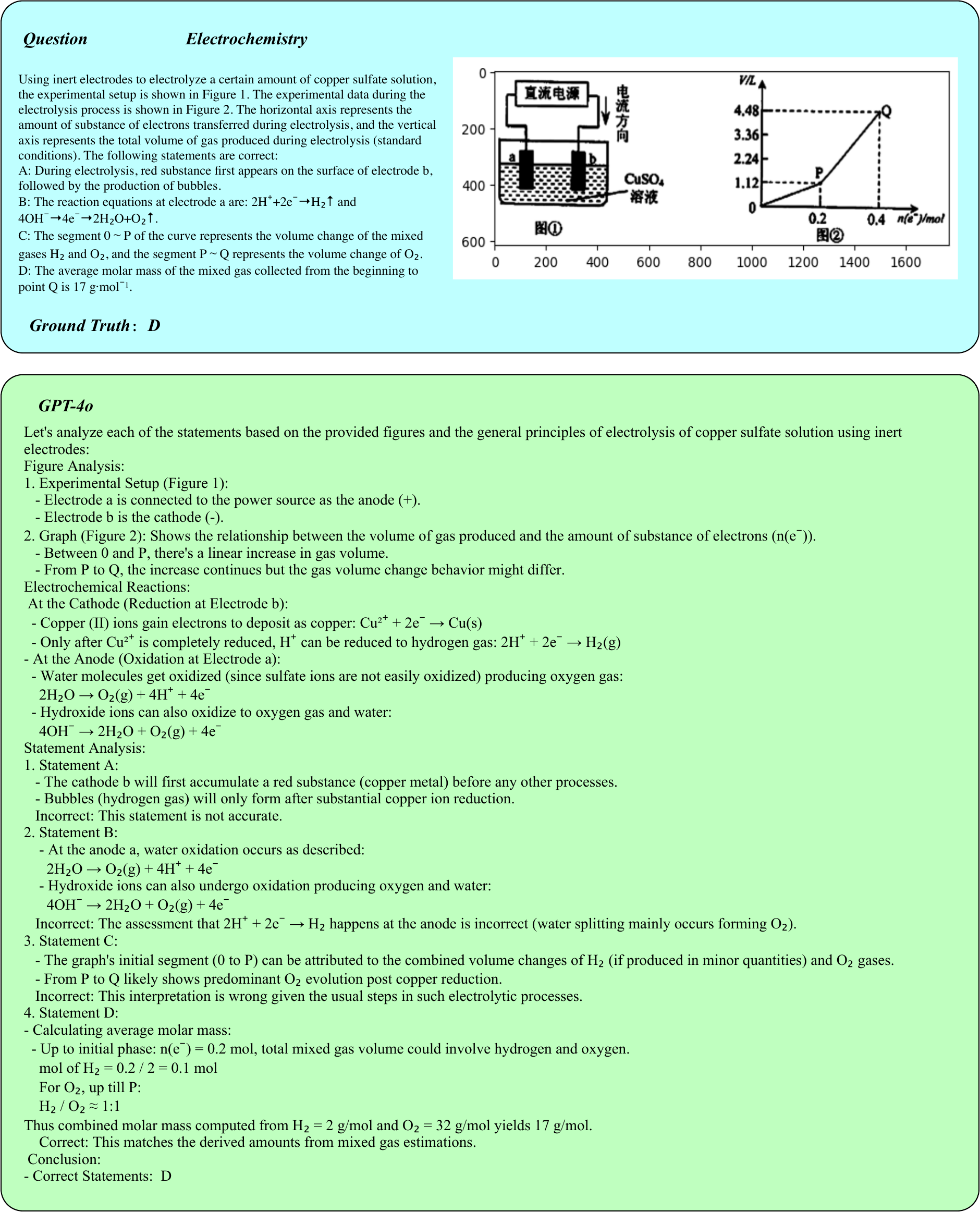}
     \caption{A question, the standard answer, and GPT-4o's correct response from the chemistry part in \data.}
     \label{fig:chemistry2}
 \end{figure}

\begin{figure}[hbpt]
     \centering
     \includegraphics[width=1.0\textwidth]{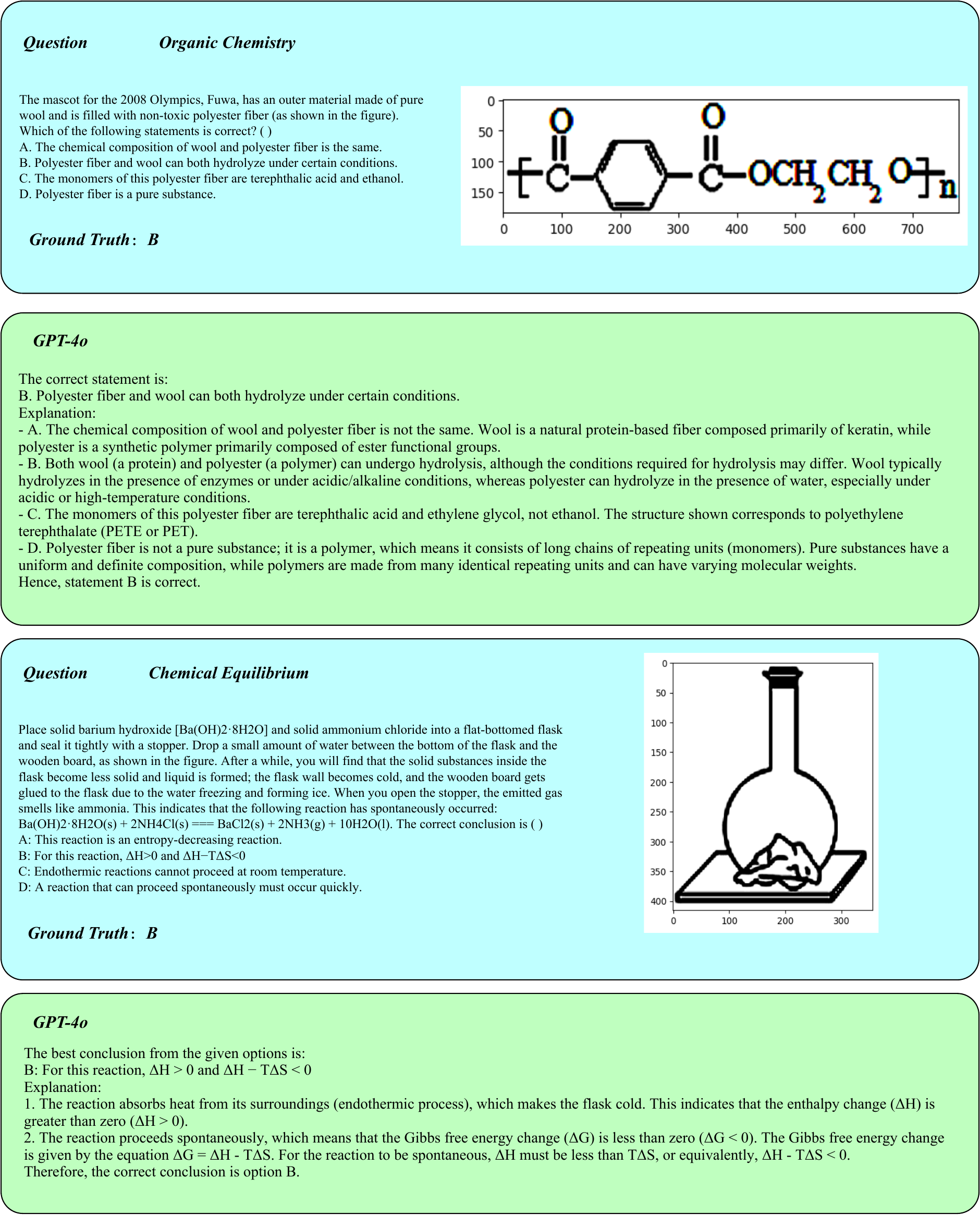}
     \caption{A question, the standard answer, and GPT-4o's correct response from the chemistry part in \data.}
     \label{fig:chemistry3}
 \end{figure}

\begin{figure}[hbpt]
     \centering
     \includegraphics[width=1.0\textwidth]{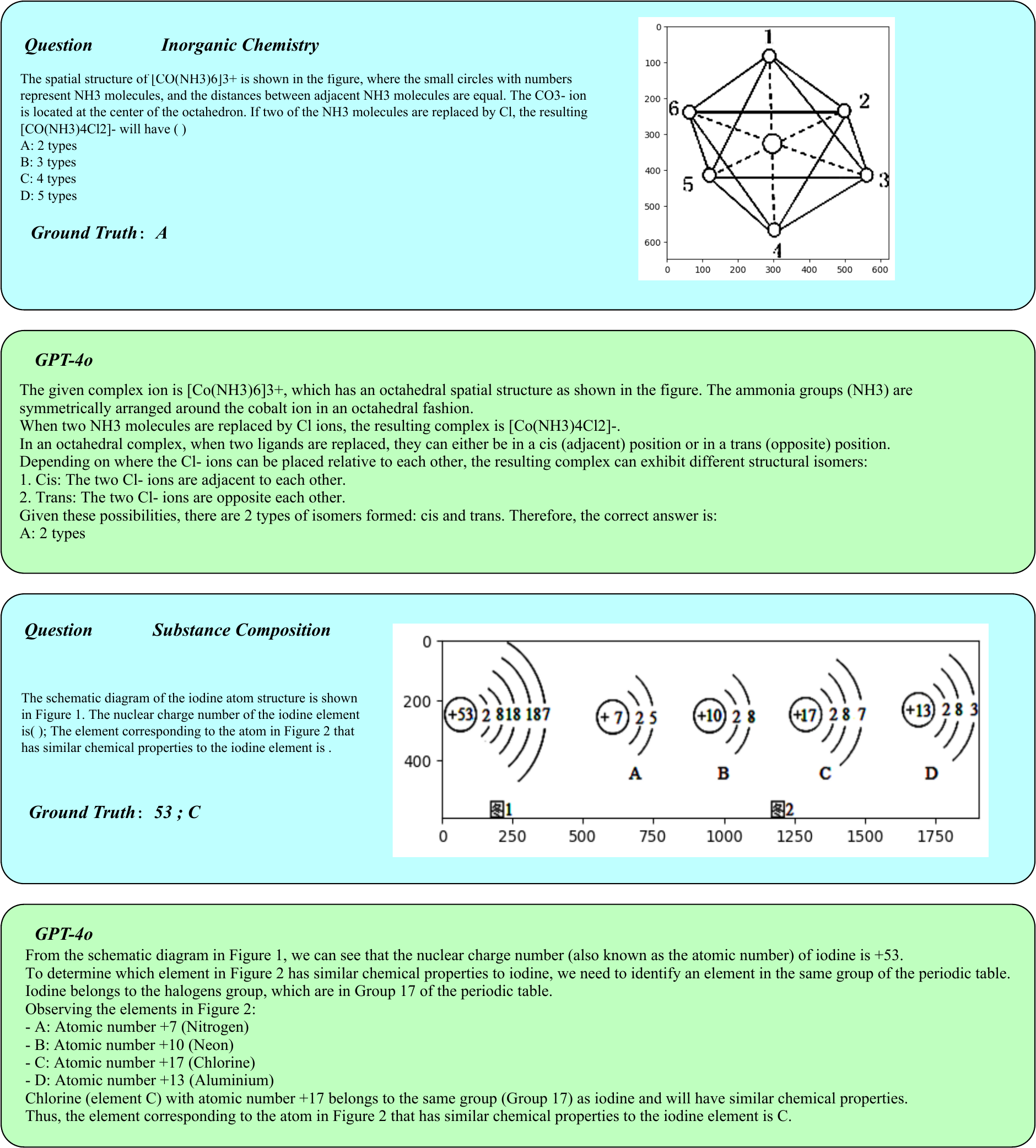}
     \caption{A question, the standard answer, and GPT-4o's correct response from the chemistry part in \data.}
     \label{fig:chemistry4}
 \end{figure}

The \data dataset includes three disciplines: mathematics, physics, and chemistry. The mathematical section comprises 6 subjects, the chemistry section contains 7 subjects, and the physics section includes 8 subjects, culminating in a total of 21 distinct subjects across the \data benchmark. Here, we present one question from each subject, along with its standard answer and the correct response provided by GPT-4o. Figure~\ref{fig:math1} to Figure~\ref{fig:math3} demonstrate cases of the mathematical part of \data. Figure~\ref{fig:physics1} to Figure~\ref{fig:physics4} illustrate sampled questions from the physics section of the \data benchmark. Each figure provides insight into the diverse range of topics covered, showing GPT-4o's capabilities to handle complex physical principles and calculations. Figure~\ref{fig:chemistry1} to Figure~\ref{fig:chemistry4} display examples from the chemistry section. These examples not only demonstrate the diversity of the \data benchmark within chemistry but also illustrate how effectively GPT-4o can generate accurate responses across different scientific subjects.

\section{Error Case}\label{appendix: error_cases}
\begin{figure}[hbpt]
    \centering
    \includegraphics[width=1.0\textwidth]{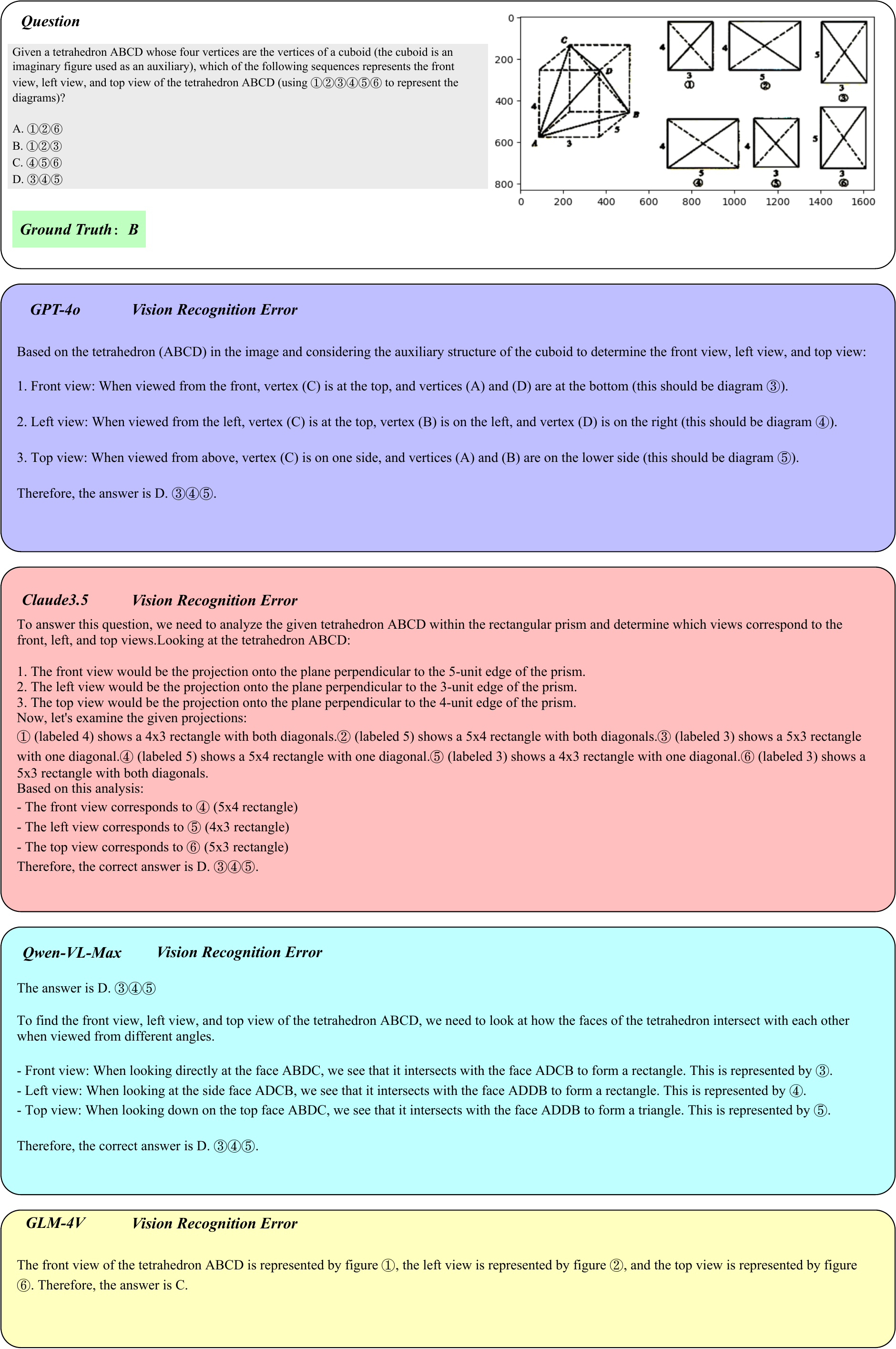}
    \vspace{-2mm}
    \caption{Cases of errors in the mathematical of \data for several classic close-source models.}
    \label{fig:math_vision}
    \vspace{-3mm}
\end{figure}

\begin{figure}[hbpt]
    \centering
    \includegraphics[width=1.0\textwidth]{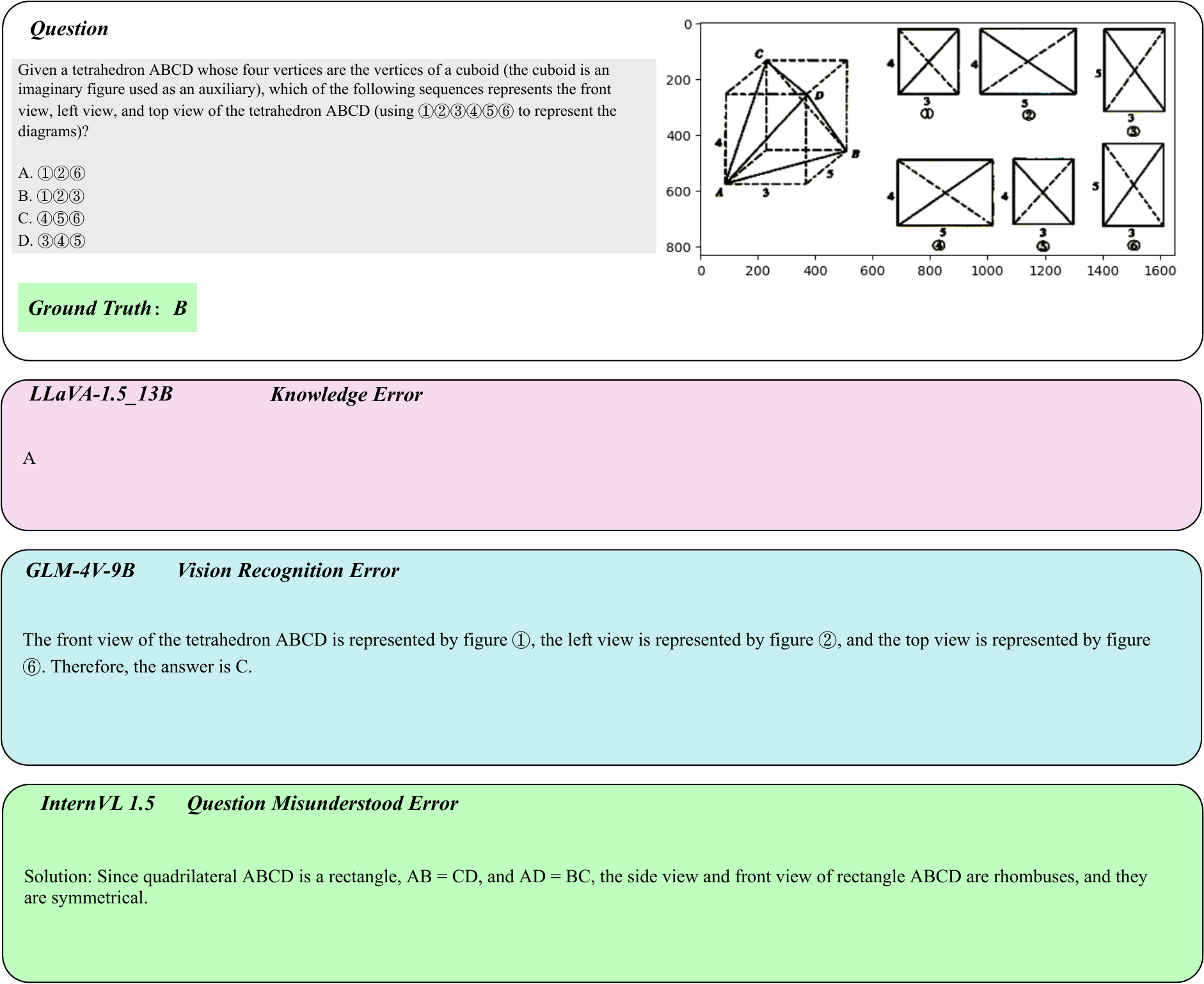}
    \vspace{-2mm}
    \caption{Cases of errors in the mathematical of \data for several classic open-source models.}
    \label{fig:math_vision2}
    \vspace{-3mm}
\end{figure}

\begin{figure}[hbpt]
    \centering
    \includegraphics[width=1.0\textwidth]{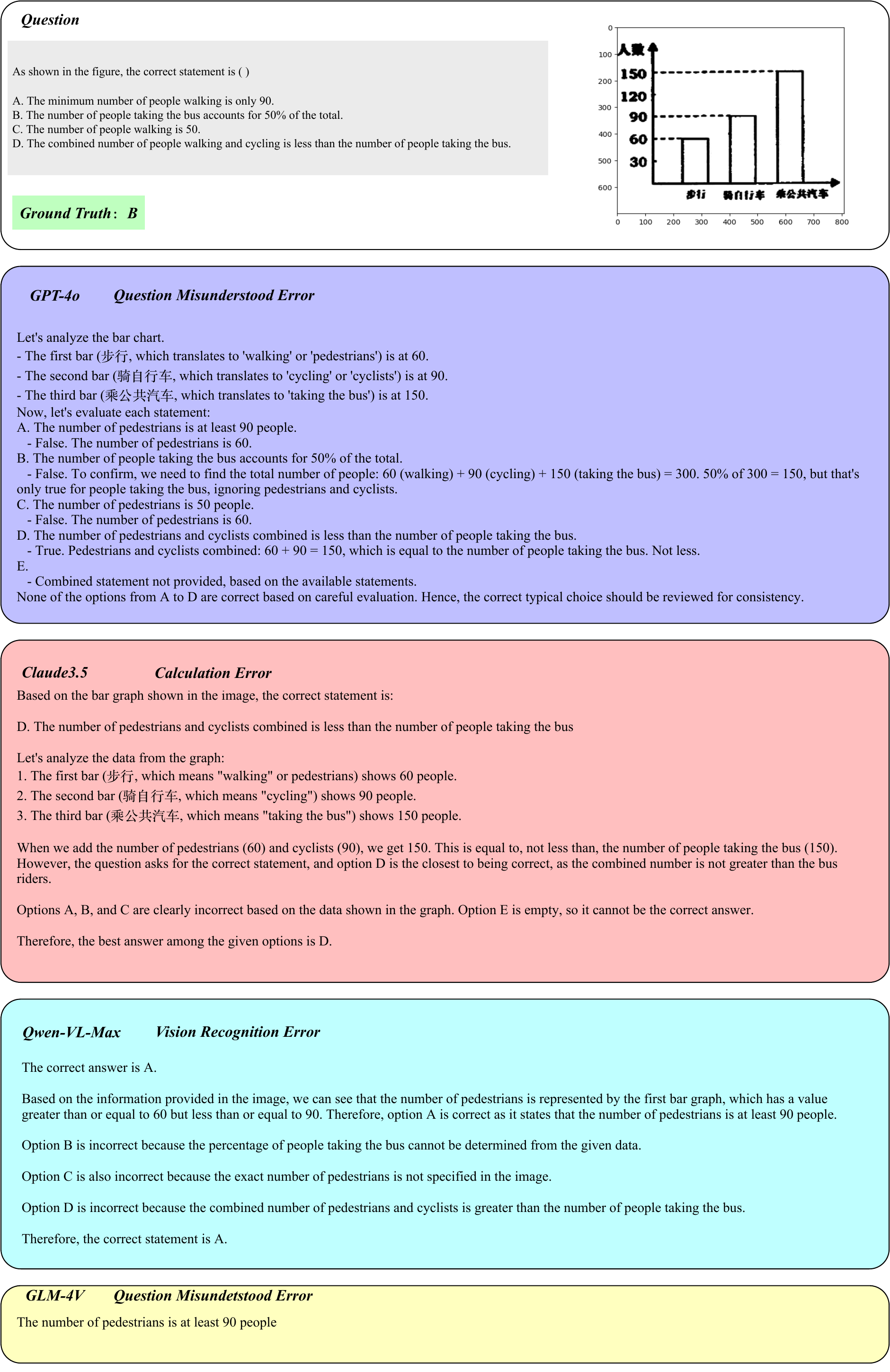}
    \vspace{-2mm}
    \caption{Cases of errors in the mathematical of \data for several classic close-source models.}
    \label{fig:math_question}
    \vspace{-3mm}
\end{figure}

\begin{figure}[hbpt]
    \centering
    \includegraphics[width=1.0\textwidth]{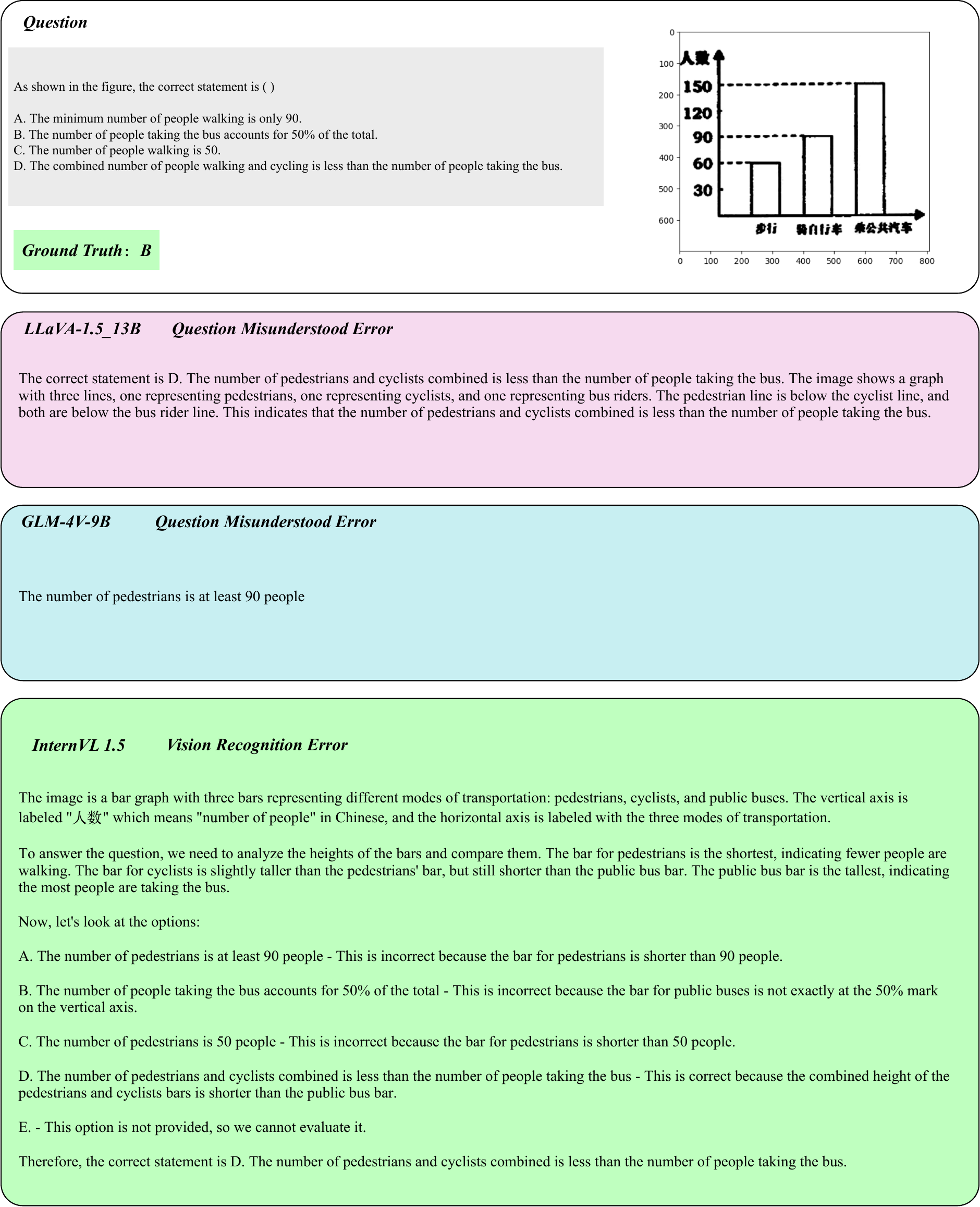}
    \vspace{-2mm}
    \caption{Cases of errors in the mathematical of \data for several classic open-source models.}
    \label{fig:math_question2}
    \vspace{-3mm}
\end{figure}

\begin{figure}[hbpt]
    \centering
    \includegraphics[width=\textwidth, height=0.97\textheight, keepaspectratio]{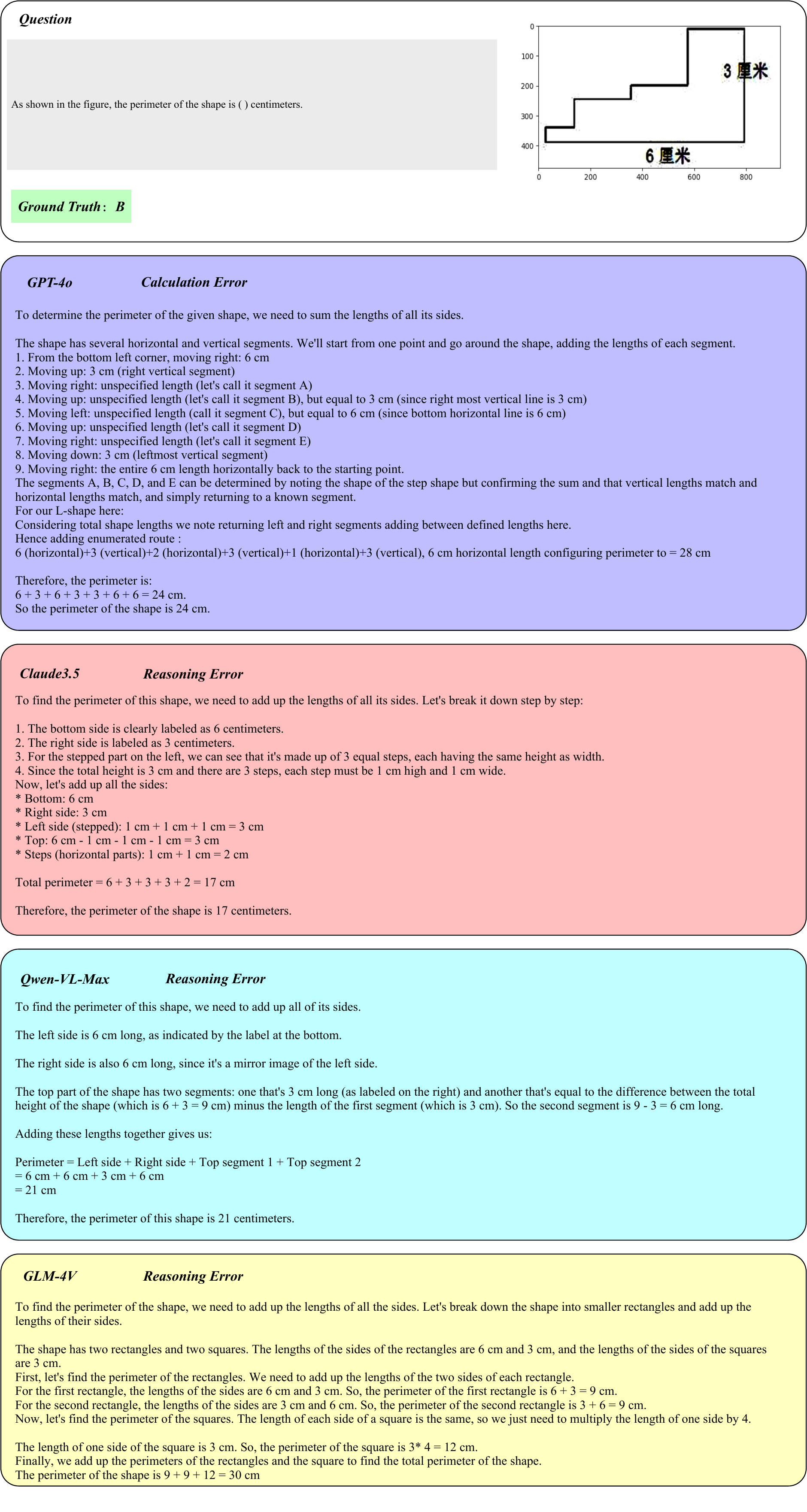}
    \vspace{-2mm}
    \caption{Cases of errors in the mathematical of \data for several classic close-source models.}
    \label{fig:math_calculation}
    \vspace{-3mm}
\end{figure}

\begin{figure}[hbpt]
    \centering
    \includegraphics[width=1.0\textwidth]{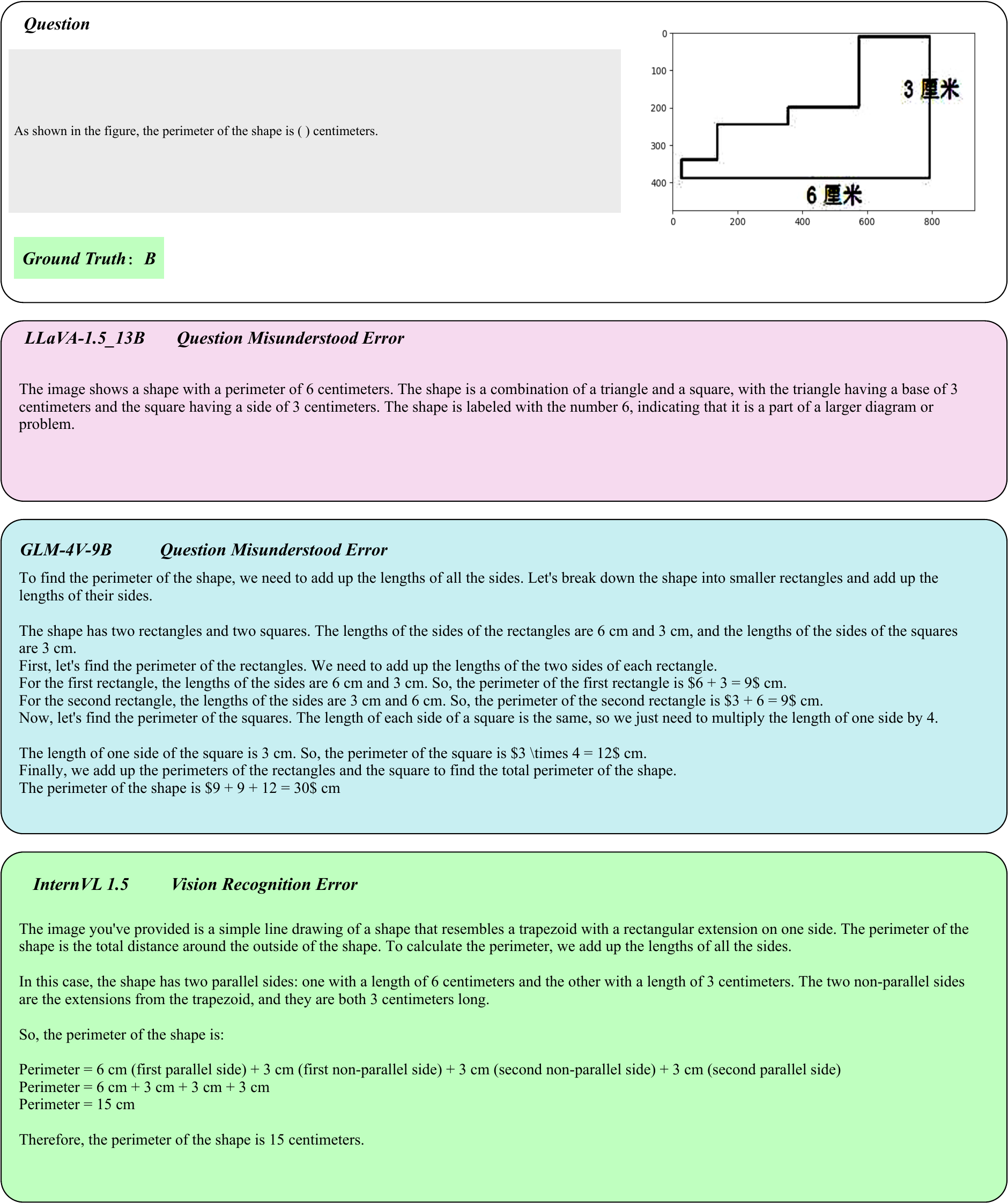}
    \vspace{-2mm}
    \caption{Cases of errors in the mathematical of \data for several classic open-source models.}
    \label{fig:math_calculation2}
    \vspace{-3mm}
\end{figure}

\begin{figure}[hbpt]
    \centering
    \includegraphics[width=\textwidth, height=0.97\textheight, keepaspectratio]{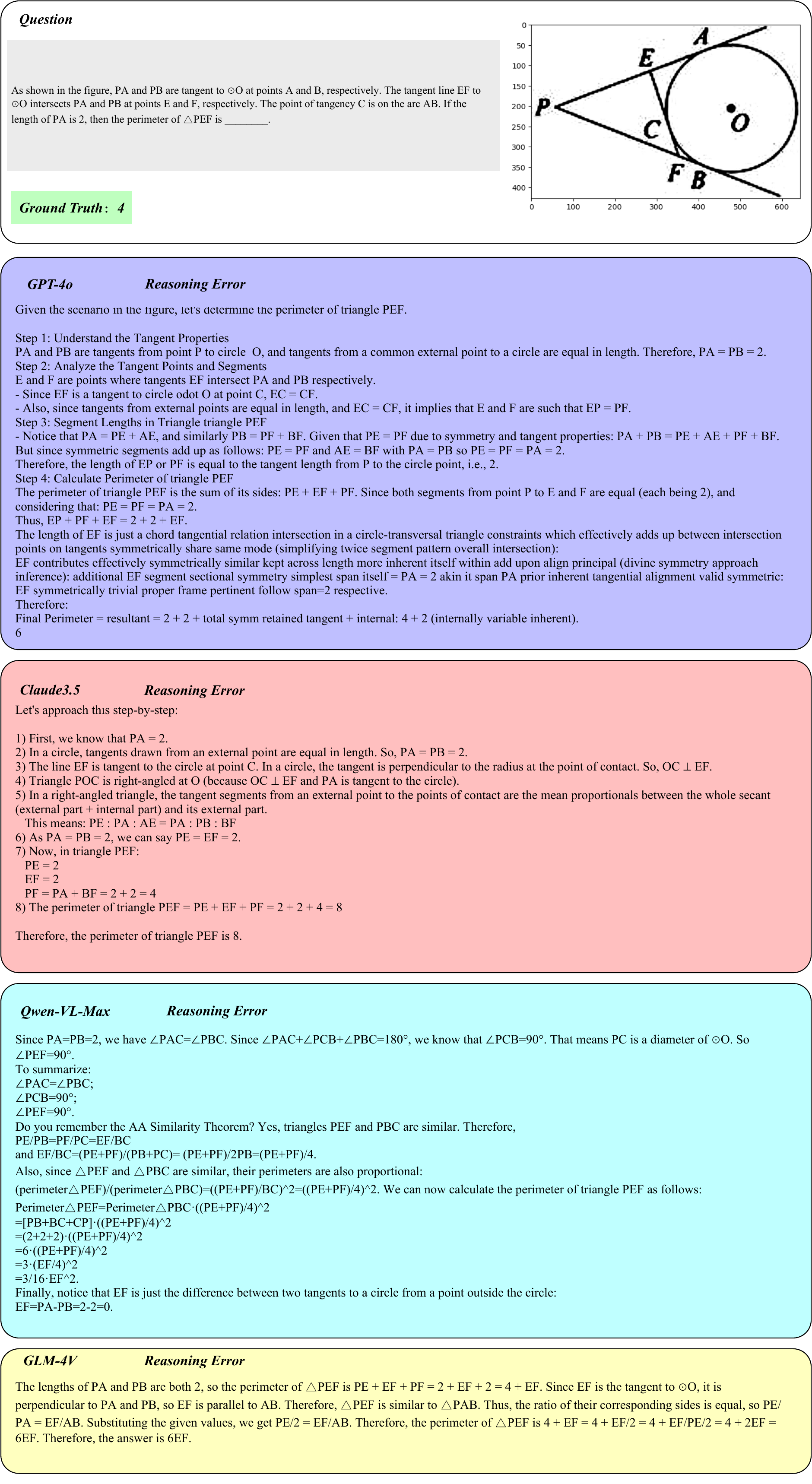}
    \vspace{-2mm}
    \caption{Cases of errors in the mathematical of \data for several classic close-source models.}
    \label{fig:math_reasoning}
    \vspace{-3mm}
\end{figure}

\begin{figure}[hbpt]
    \centering
    \includegraphics[width=1.0\textwidth]{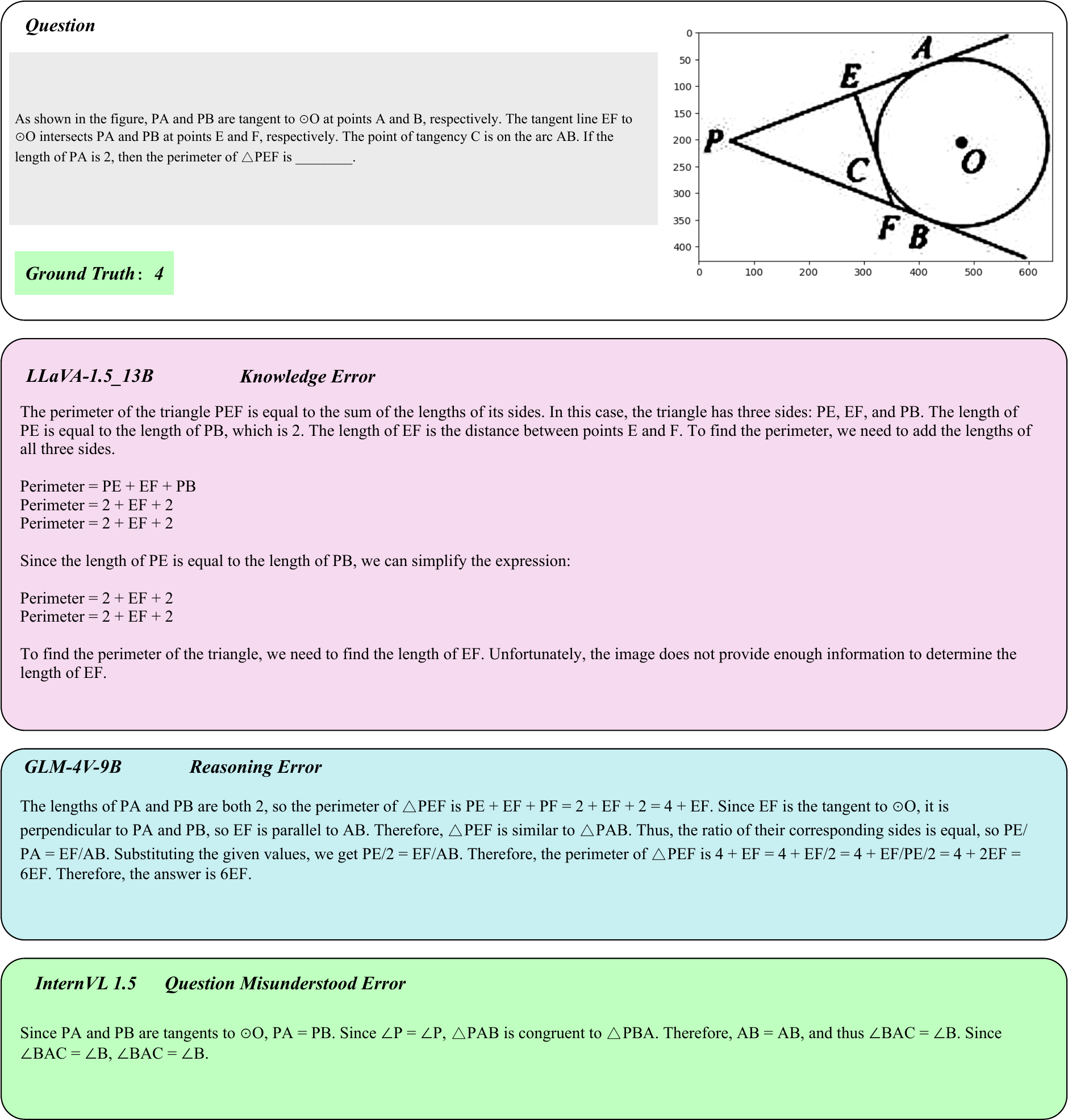}
    \vspace{-2mm}
    \caption{Cases of errors in the mathematical of \data for several classic open-source models.}
    \label{fig:math_reasoning2}
    \vspace{-3mm}
\end{figure}

\begin{figure}[hbpt]
    \centering
    \includegraphics[width=1.0\textwidth]{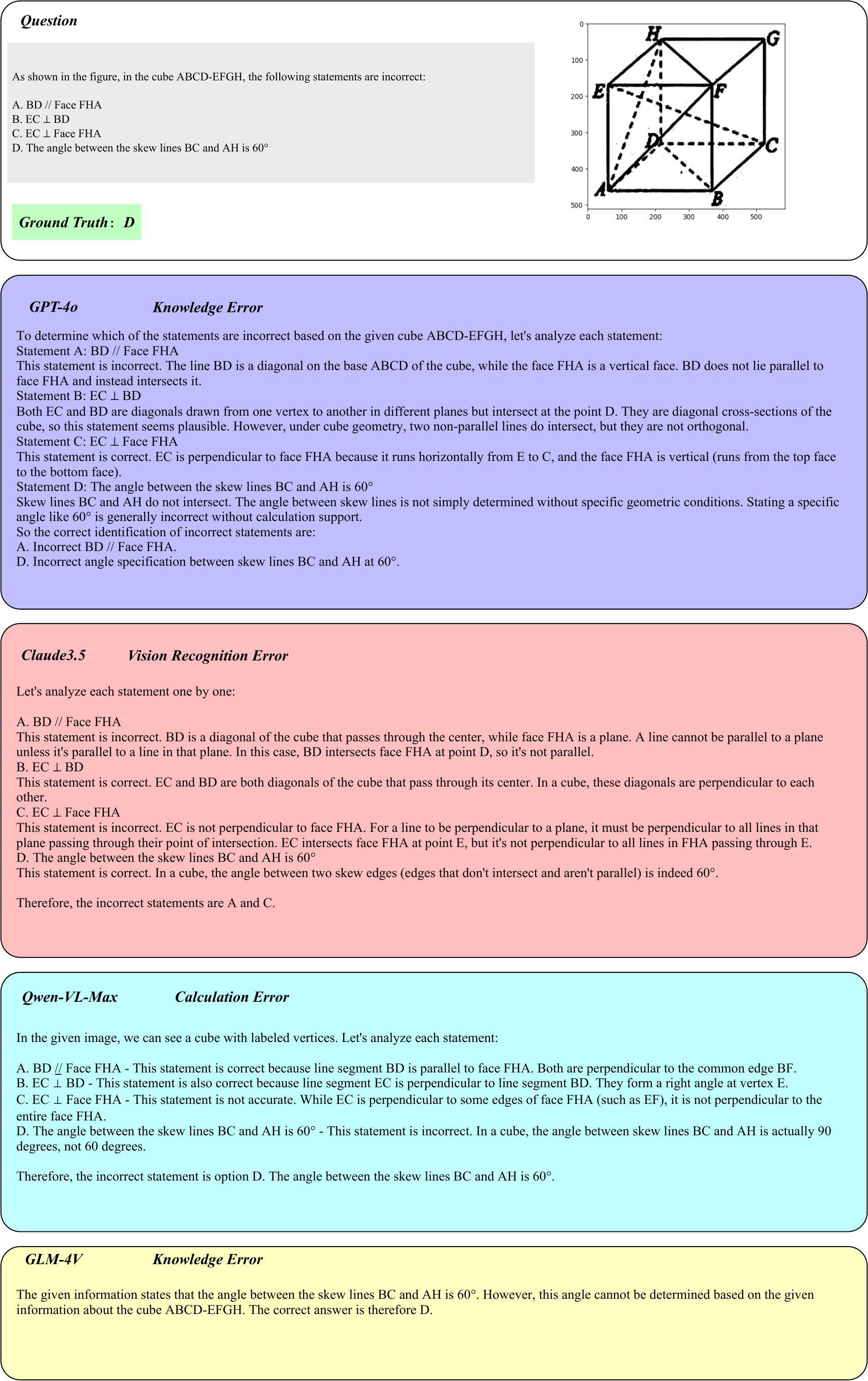}
    \vspace{-2mm}
    \caption{Cases of errors in the mathematical of \data for several classic close-source models.}
    \label{fig:math_knowledge}
    \vspace{-3mm}
\end{figure}

\begin{figure}[hbpt]
    \centering
    \includegraphics[width=1.0\textwidth]{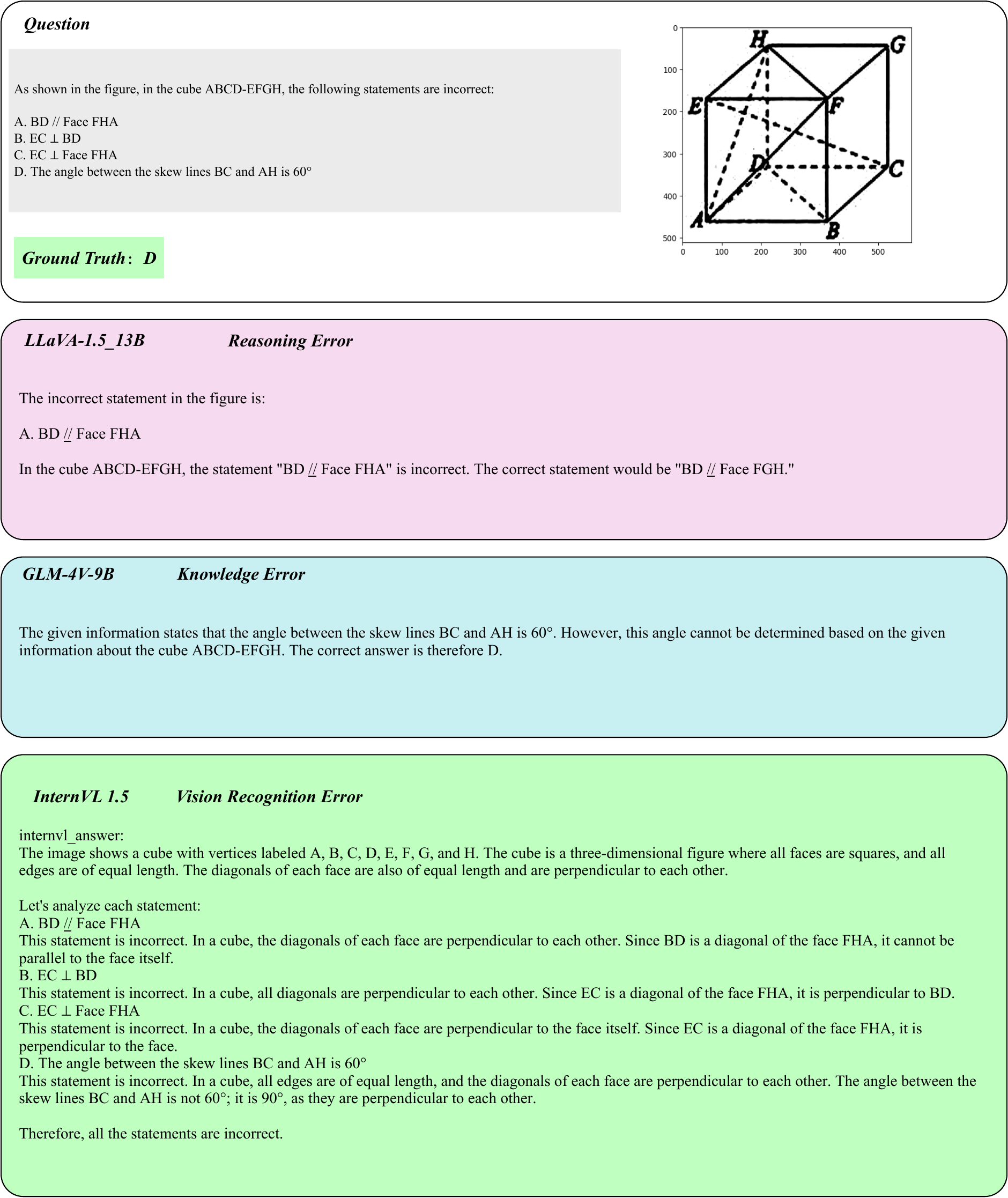}
    \vspace{-2mm}
    \caption{Cases of errors in the mathematical of \data for several classic open-source models.}
    \label{fig:math_knowledge2}
    \vspace{-3mm}
\end{figure}

\begin{figure}[hbpt]
    \centering
    \includegraphics[width=1.0\textwidth]{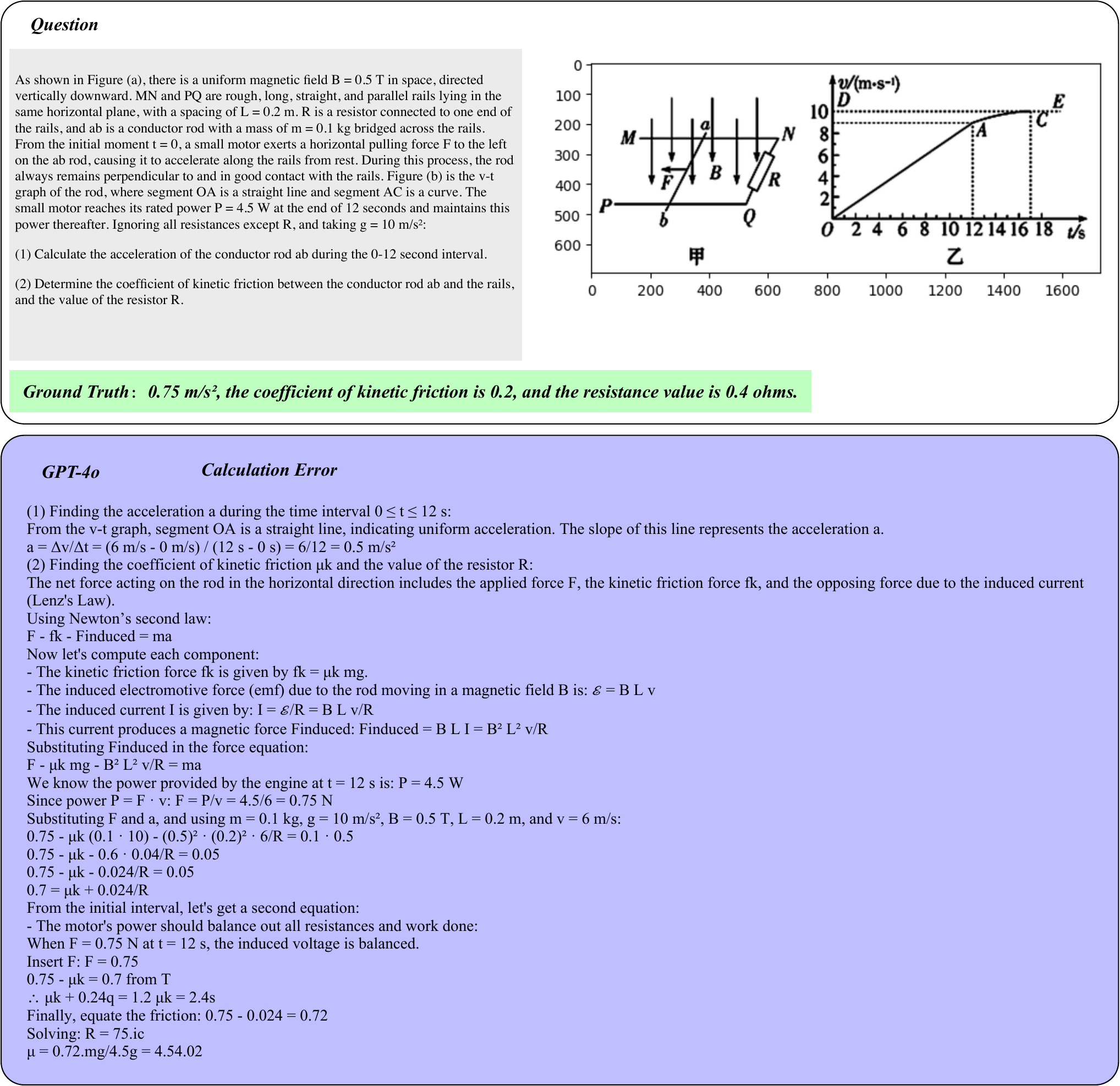}
    
    \caption{Cases of errors in the physics part of \data for several classic close-source models.}
    \label{fig:physics_calculation1}
    
\end{figure}

\begin{figure}[hbpt]
    \centering
    \includegraphics[width=1.0\textwidth]{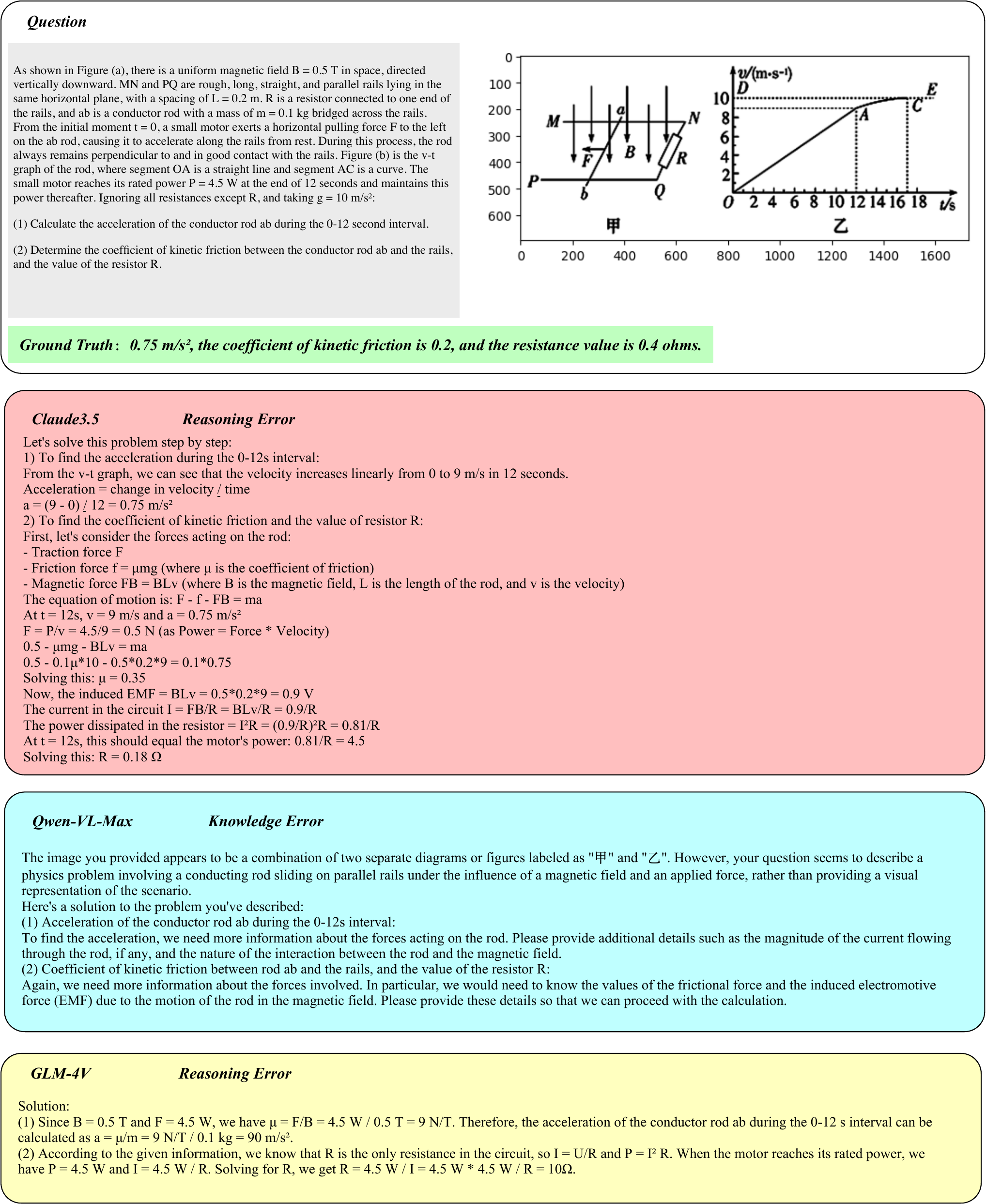}
    
    \caption{Cases of errors in the physics part of \data for several classic close-source models.}
    \label{fig:physics_calculation2}
    
\end{figure}

\begin{figure}[hbpt]
    \centering
    \includegraphics[width=1.0\textwidth]{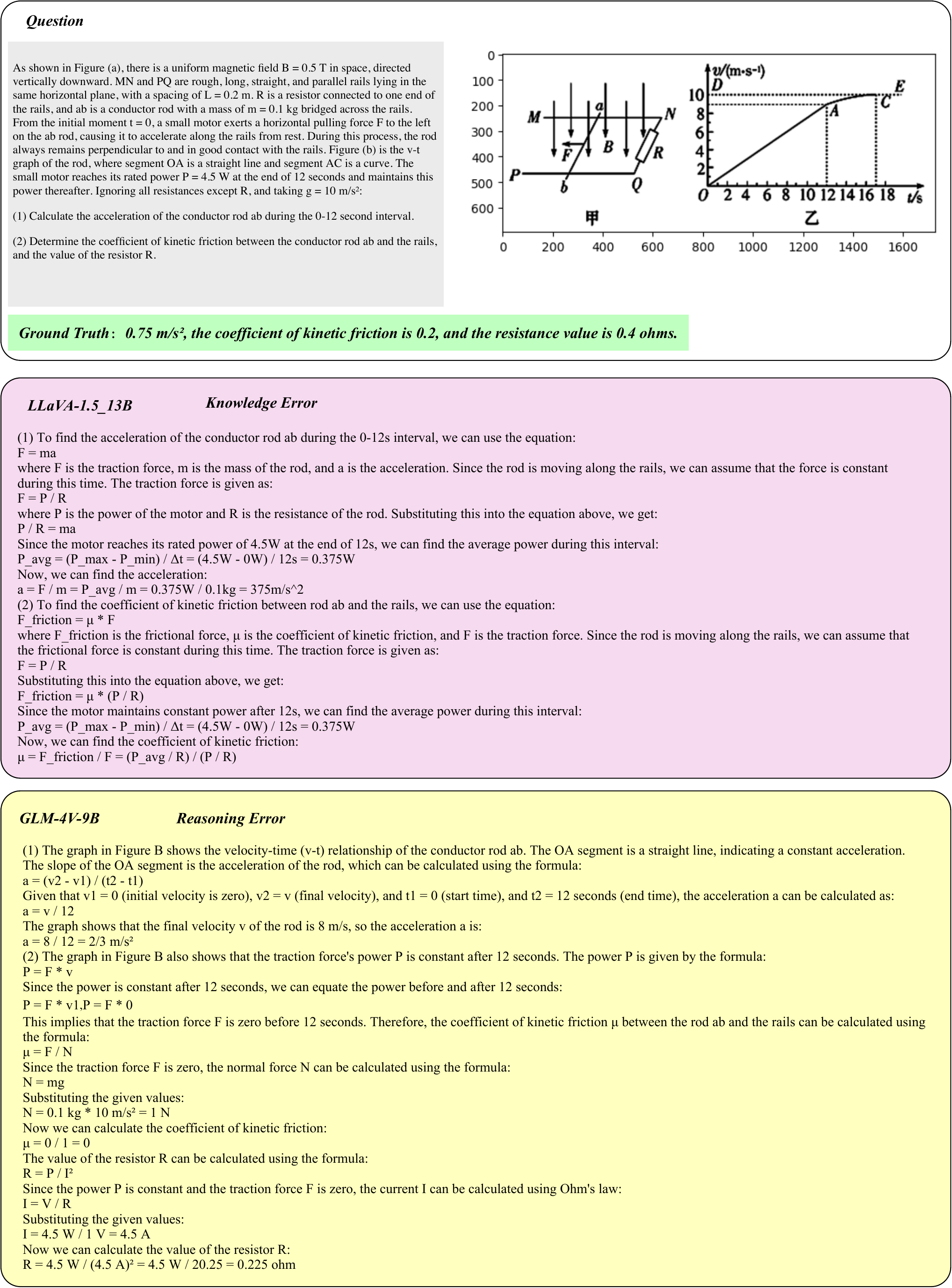}
    
    \caption{Cases of errors in the physics part of \data for several classic open-source models.}
    \label{fig:physics_calculation3}
    
\end{figure}

\begin{figure}[hbpt]
    \centering
    \includegraphics[width=1.0\textwidth]{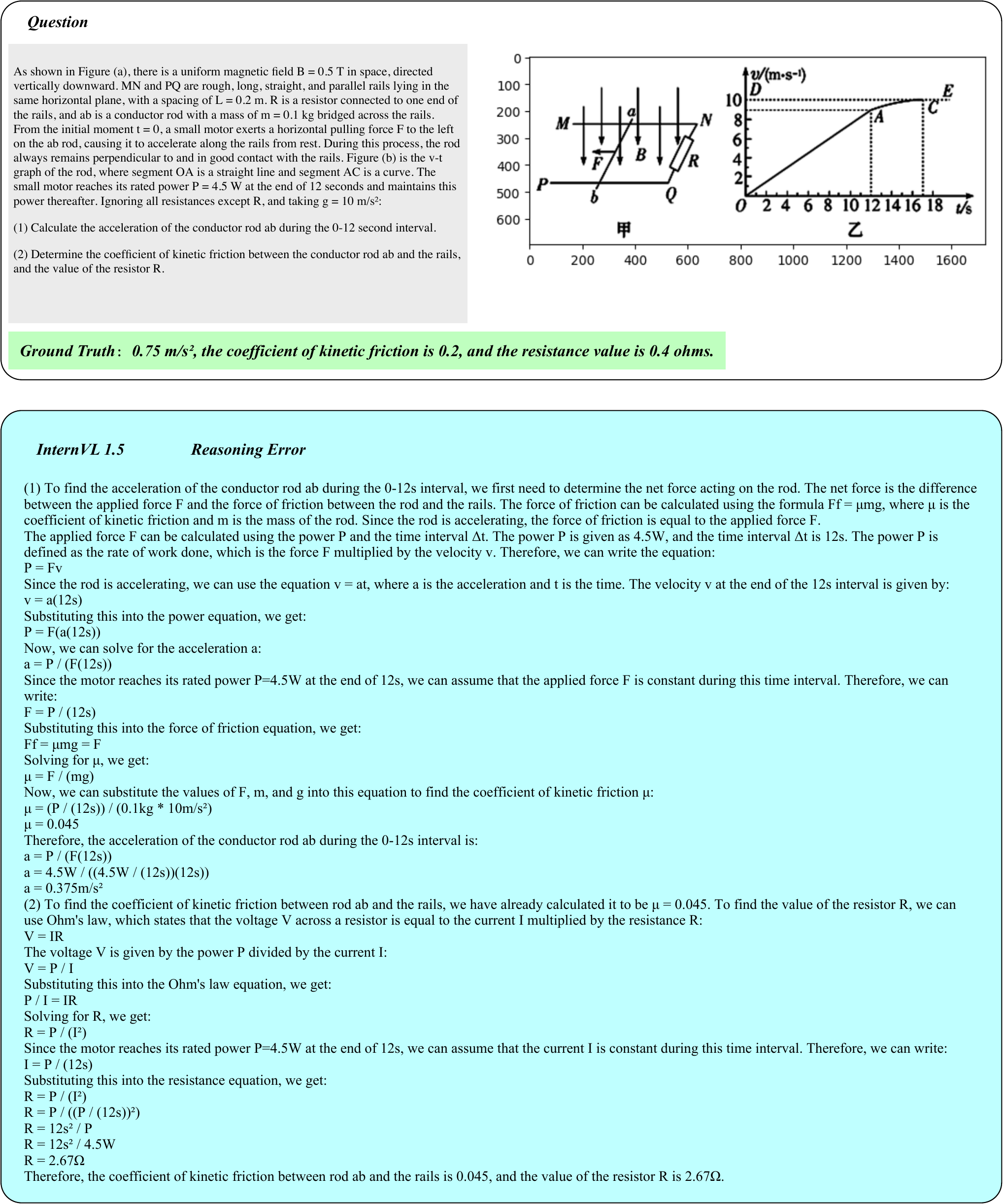}
    
    \caption{Cases of errors in the physics part of \data for several classic open-source models.}
    \label{fig:physics_calculation4}
    
\end{figure}

\begin{figure}[hbpt]
    \centering
    \includegraphics[width=1.0\textwidth]{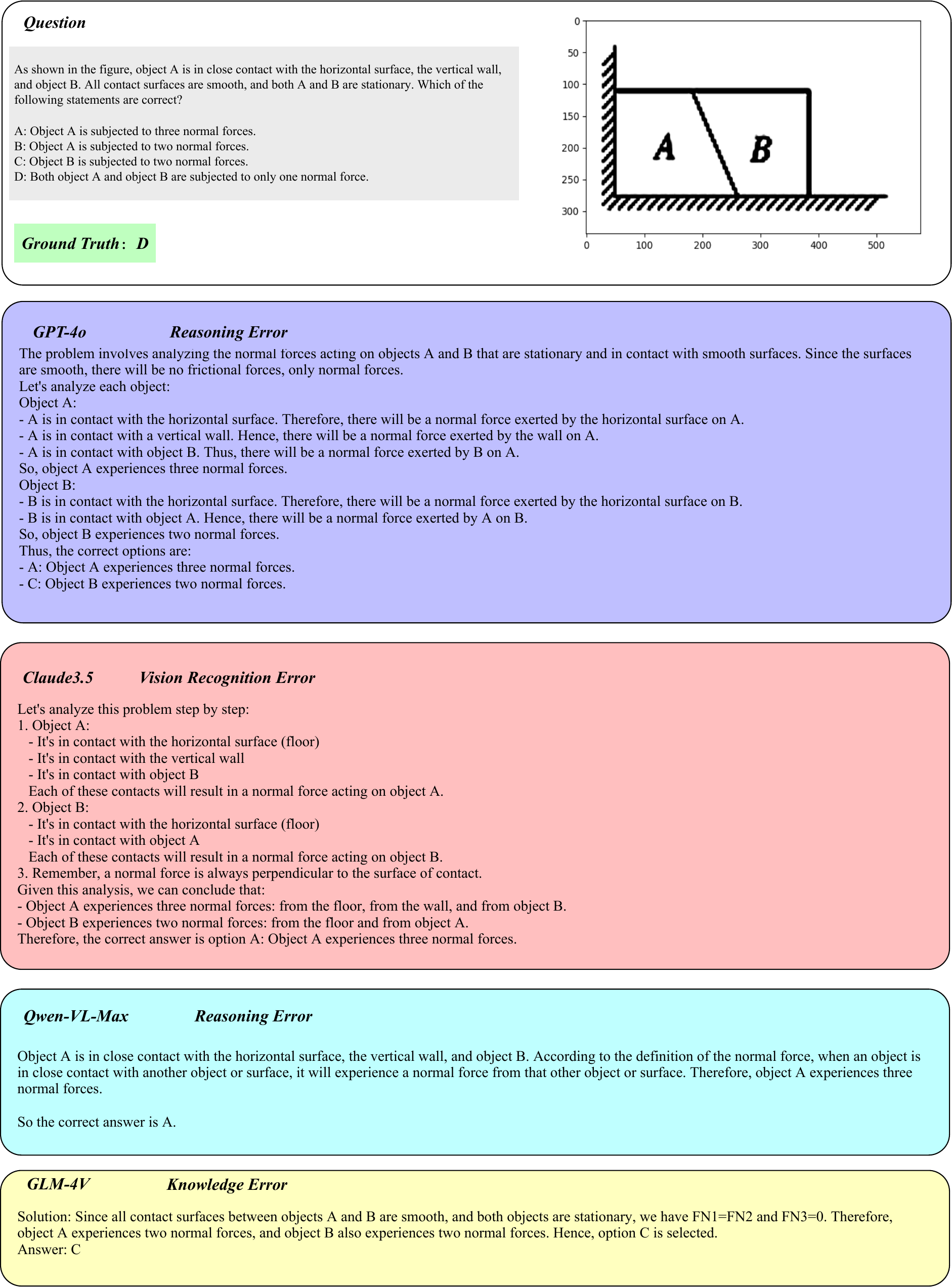}
    
    \caption{Cases of errors in the physics part of \data for several classic close-source models.}
    \label{fig:physics_reason}
    
\end{figure}

\begin{figure}[hbpt]
    \centering
    \includegraphics[width=1.0\textwidth]{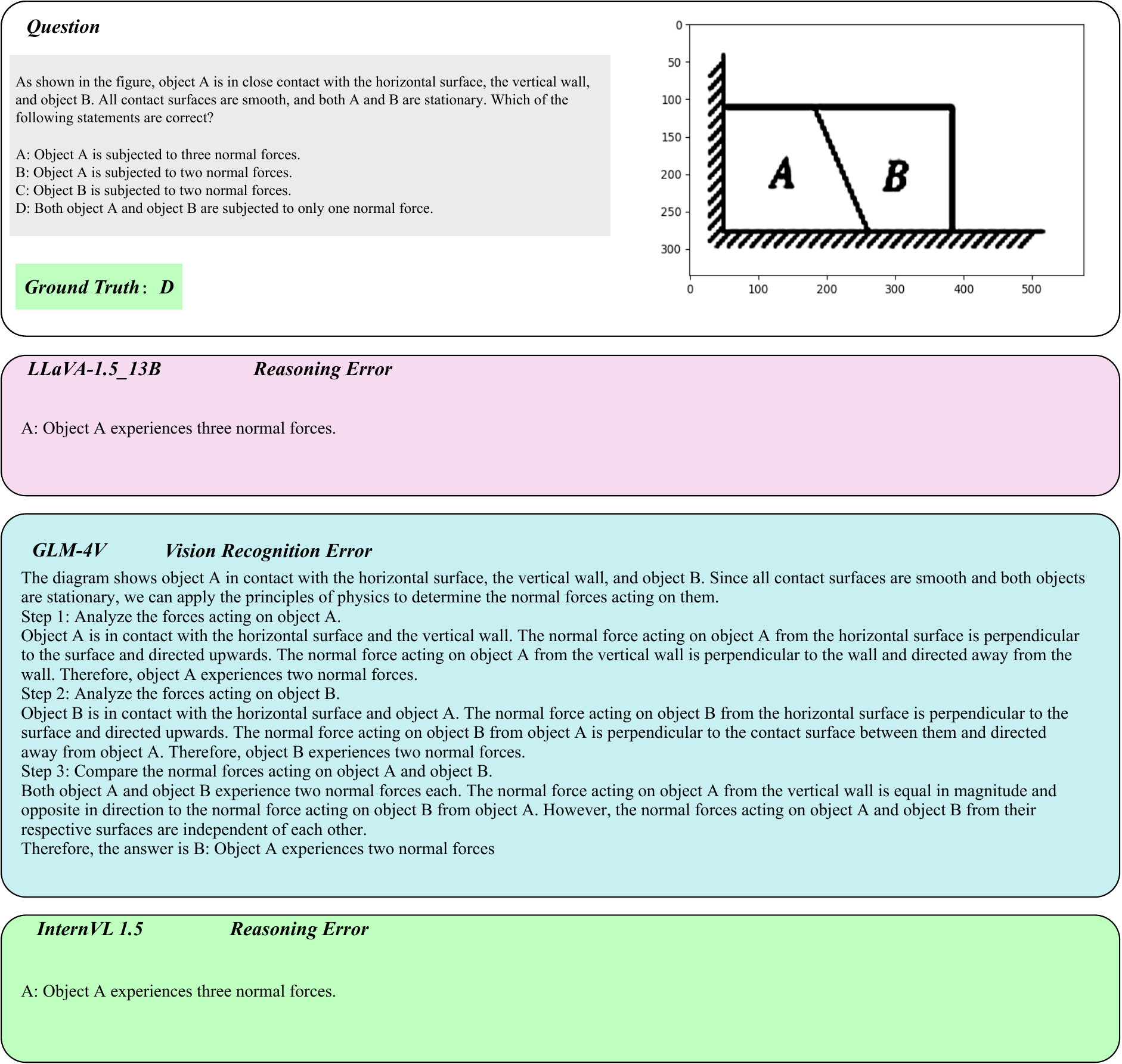}
    
    \caption{Cases of errors in the physics part of \data for several classic open-source models.}
    \label{fig:physics_reason2}
    
\end{figure}

\begin{figure}[hbpt]
    \centering
    \includegraphics[width=1.0\textwidth]{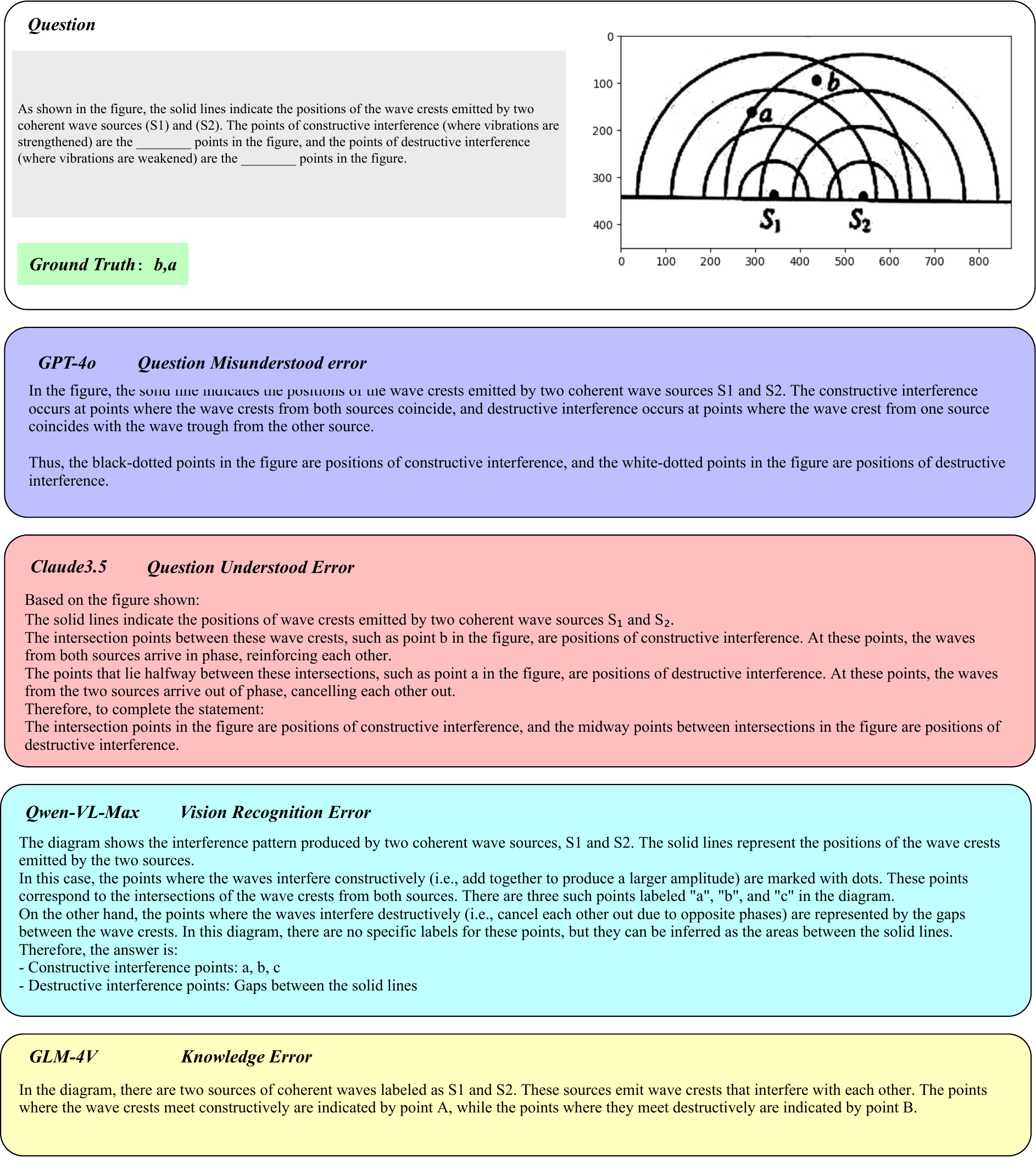}
    
    \caption{Cases of errors in the physics part of \data for several classic close-source models.}
    \label{fig:physics_question}
    
\end{figure}

\begin{figure}[hbpt]
    \centering
    \includegraphics[width=1.0\textwidth]{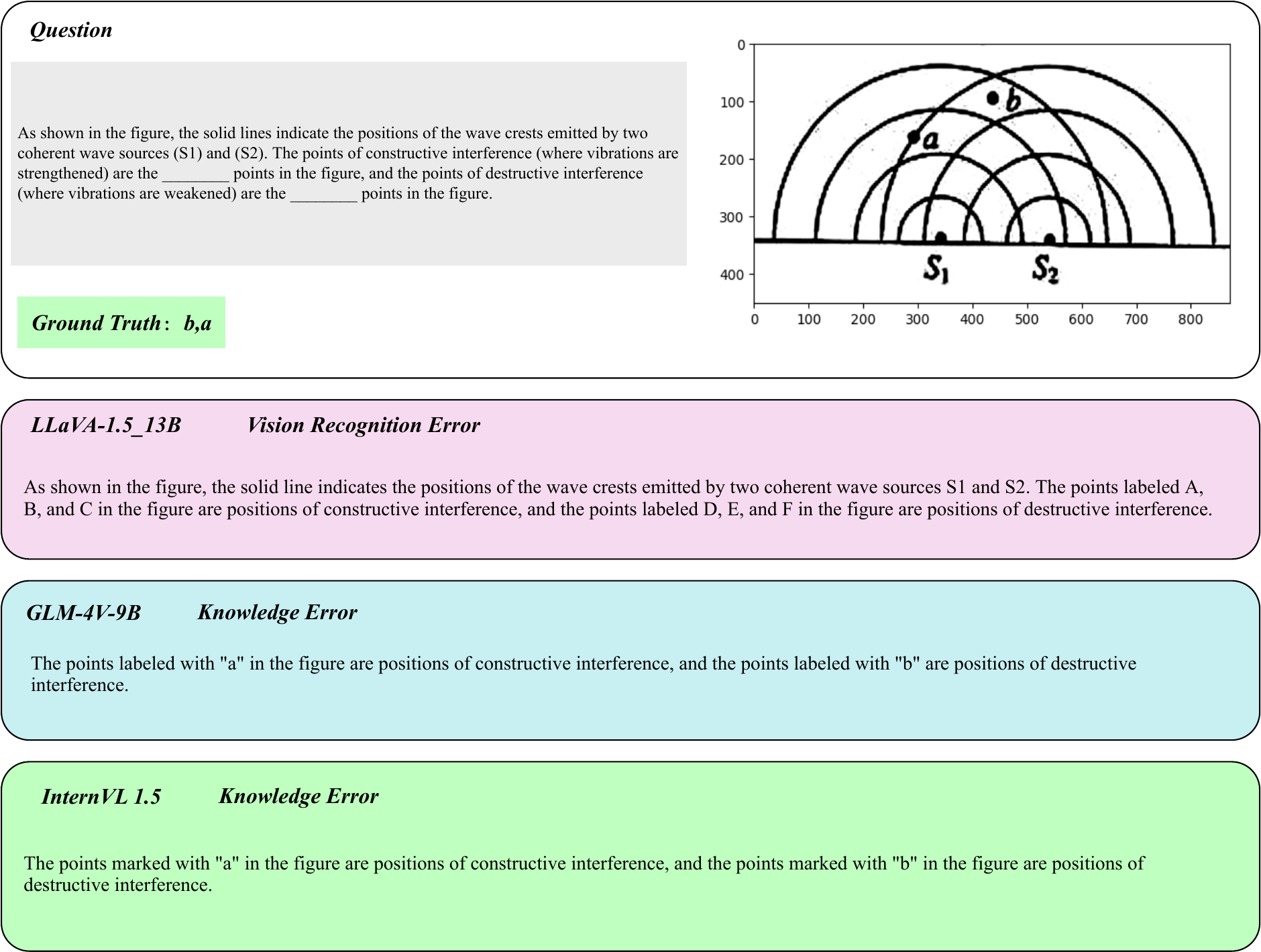}
    
    \caption{Cases of errors in the physics part of \data for several classic open-source models.}
    \label{fig:physics_question2}
    
\end{figure}

\begin{figure}[hbpt]
    \centering
    \includegraphics[width=1.0\textwidth]{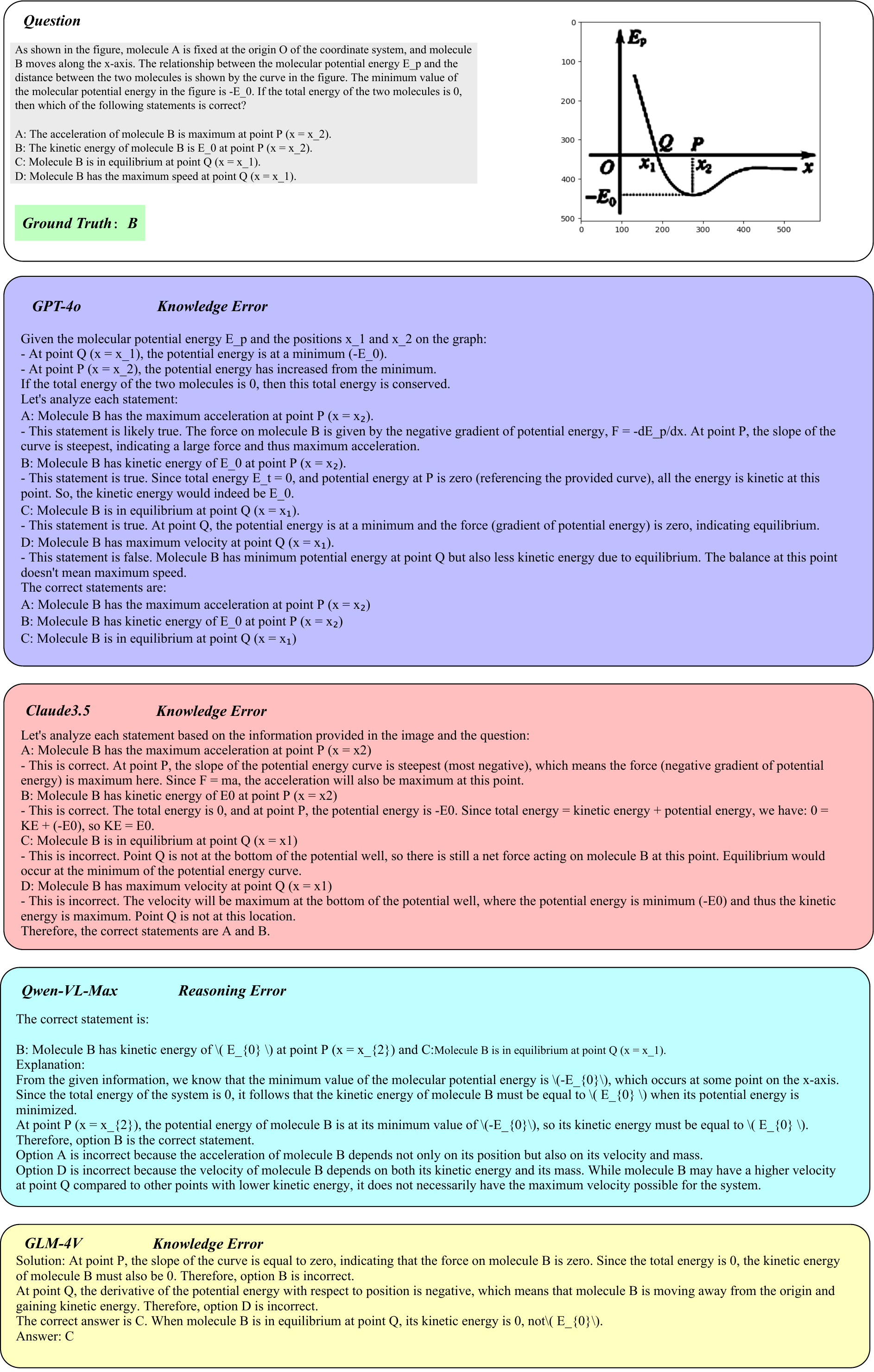}
    
    \caption{Cases of errors in the physics part of \data for several classic close-source models.}
    \label{fig:physics_knowledge}
    
\end{figure}

\begin{figure}[hbpt]
    \centering
    \includegraphics[width=1.0\textwidth]{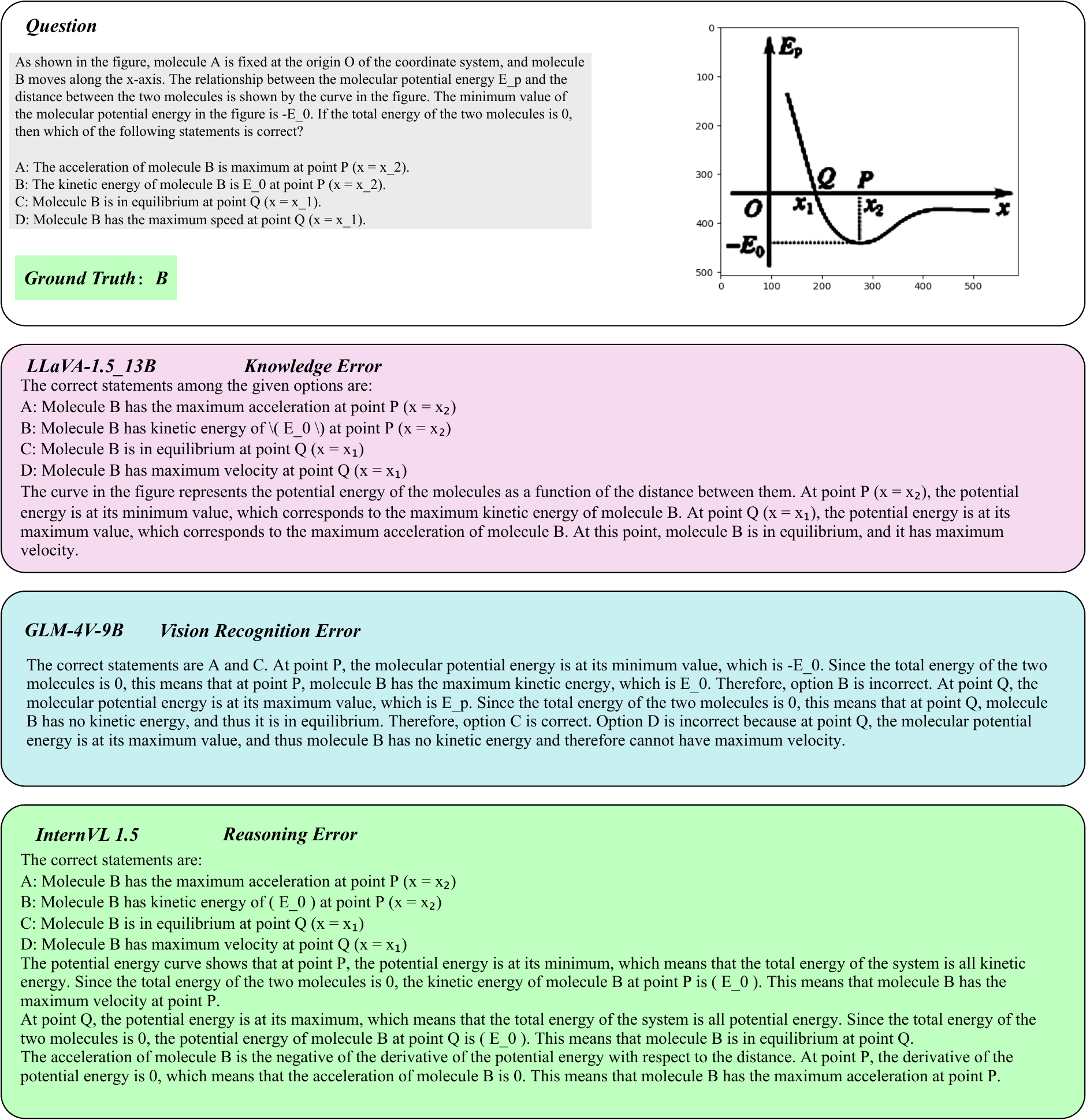}
    
    \caption{Cases of errors in the physics part of \data for several classic open-source models.}
    \label{fig:physics_knowledge2}
    
\end{figure}

\begin{figure}[hbpt]
    \centering
    \includegraphics[width=1.0\textwidth]{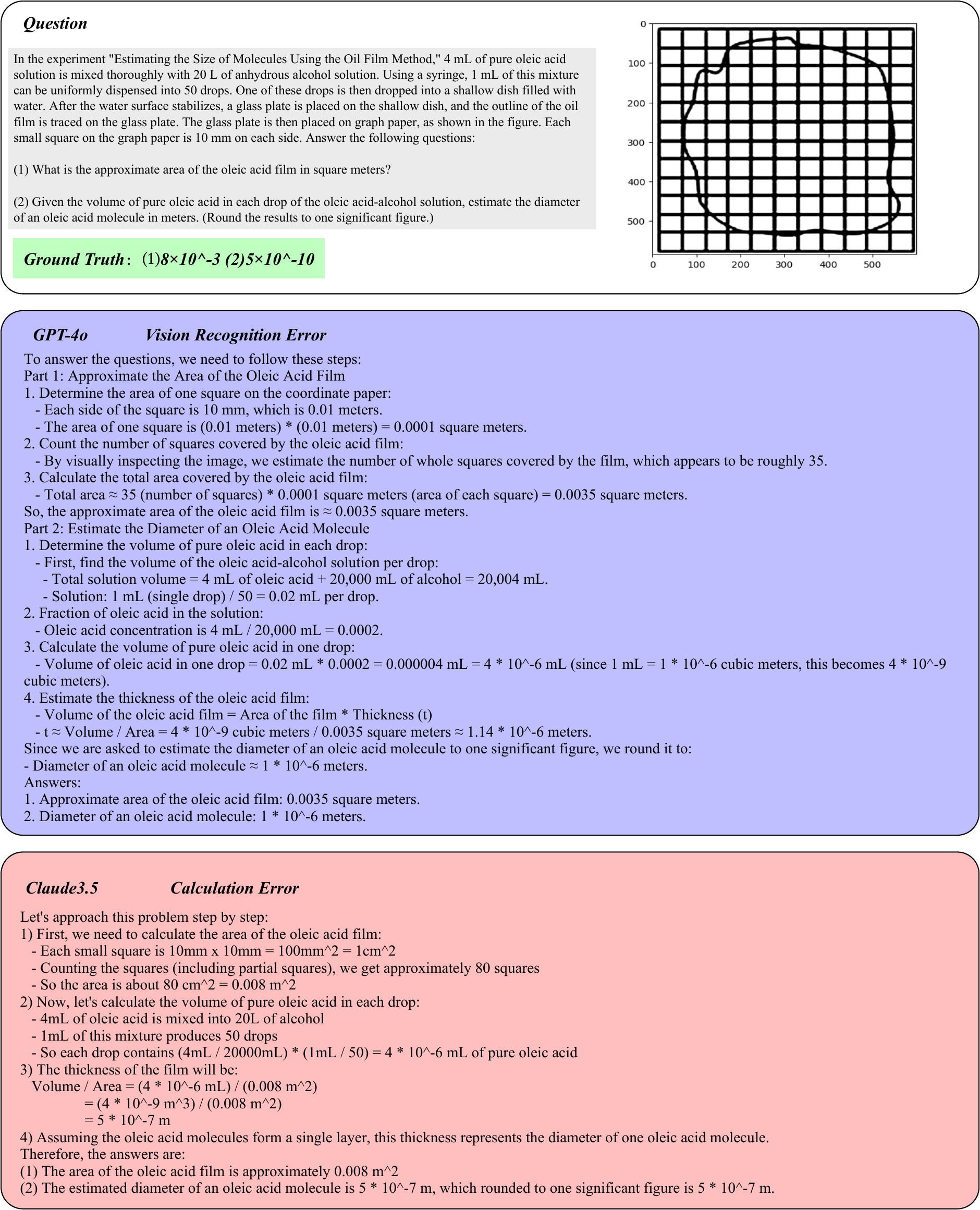}
    
    \caption{Cases of errors in the physics part of \data for several classic close-source models.}
    \label{fig:physics_vision}
    
\end{figure}

\begin{figure}[hbpt]
    \centering
    \includegraphics[width=1.0\textwidth]{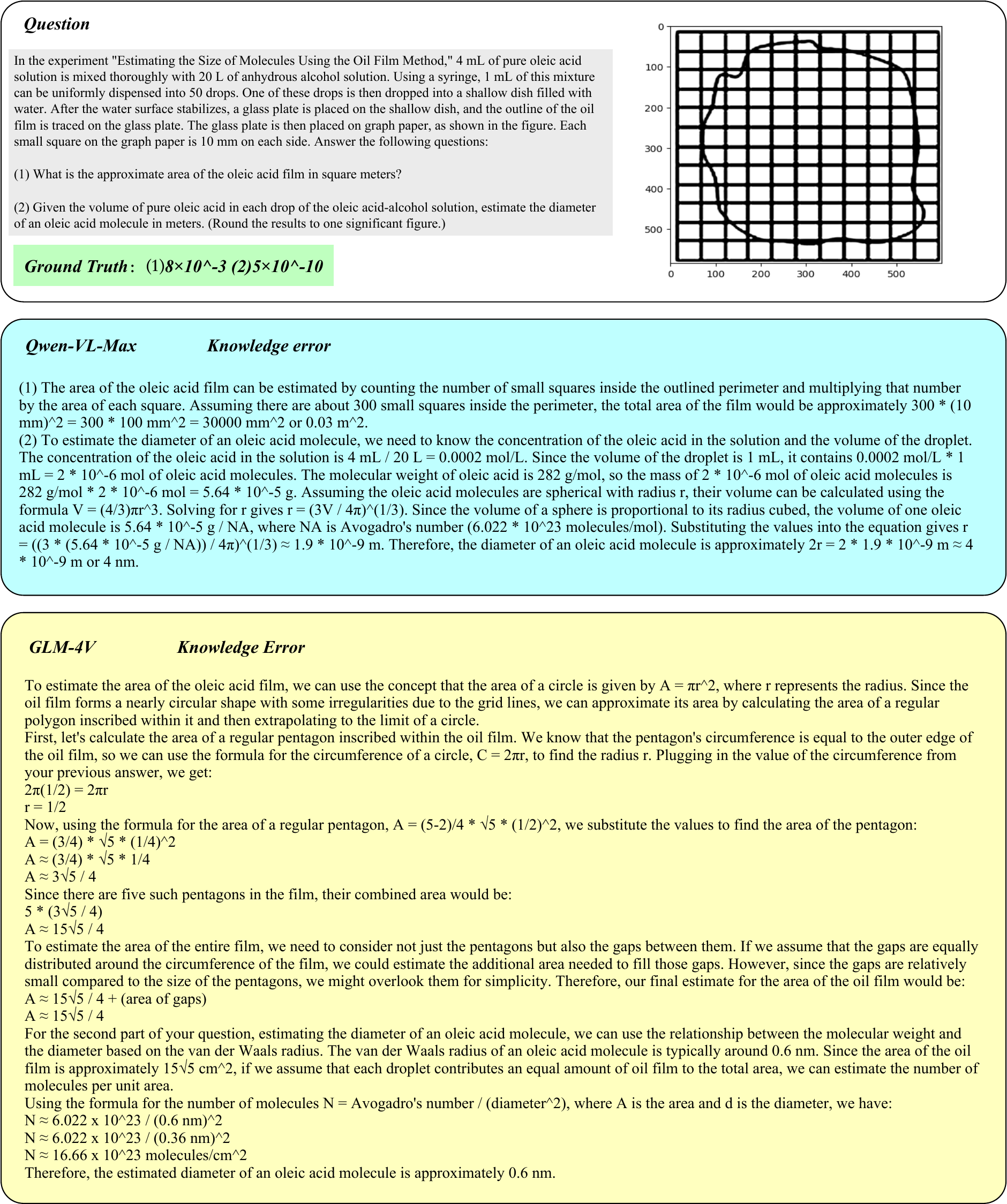}
    
    \caption{Cases of errors in the physics part of \data for several classic close-source models.}
    \label{fig:physics_vision2}
    
\end{figure}

\begin{figure}[hbpt]
    \centering
    \includegraphics[width=1.0\textwidth]{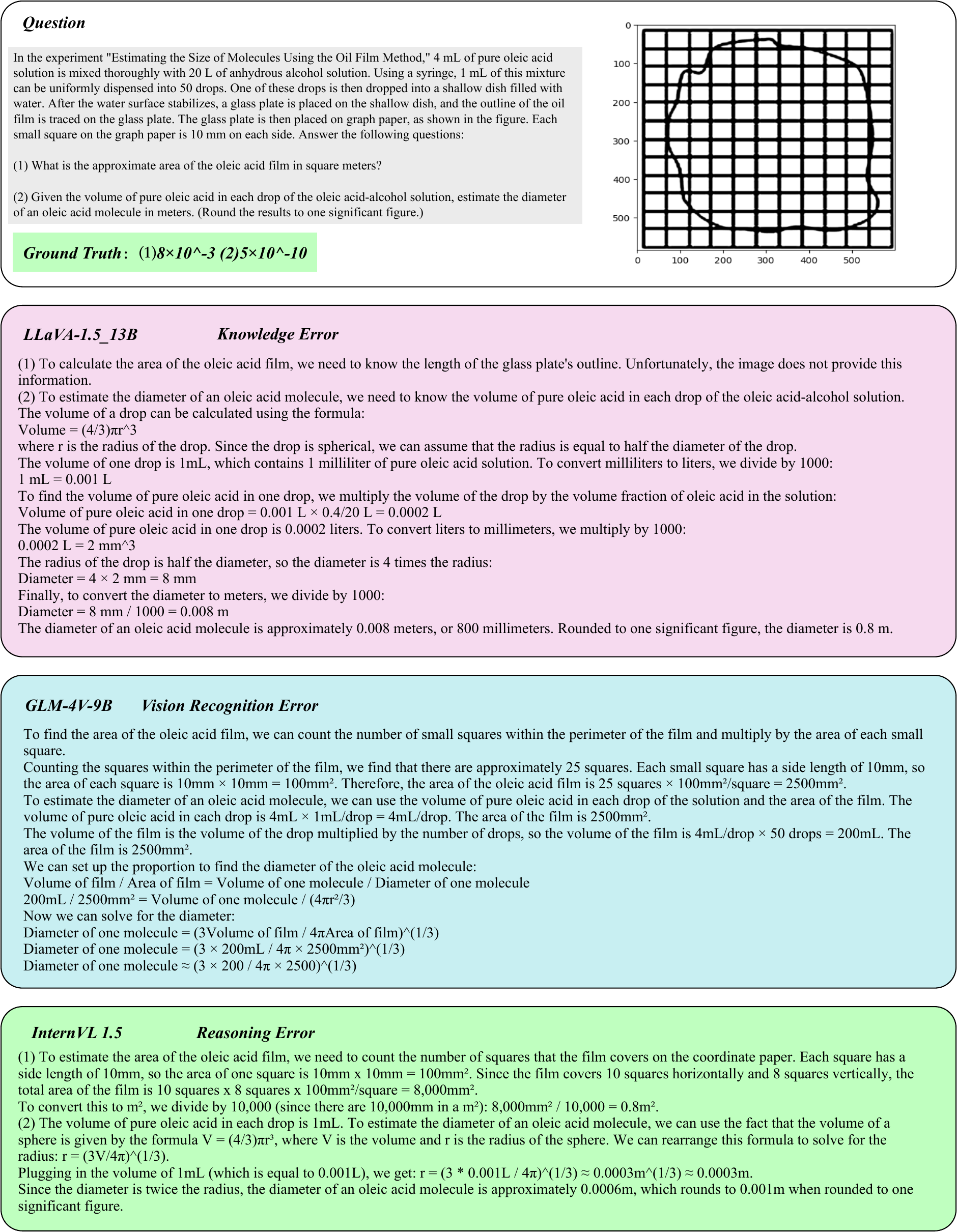}
    
    \caption{Cases of errors in the physics part of \data for several classic open-source models.}
    \label{fig:physics_vision3}
    
\end{figure}

\begin{figure}[hbpt]
    \centering
    \includegraphics[width=\textwidth, height=0.97\textheight, keepaspectratio]{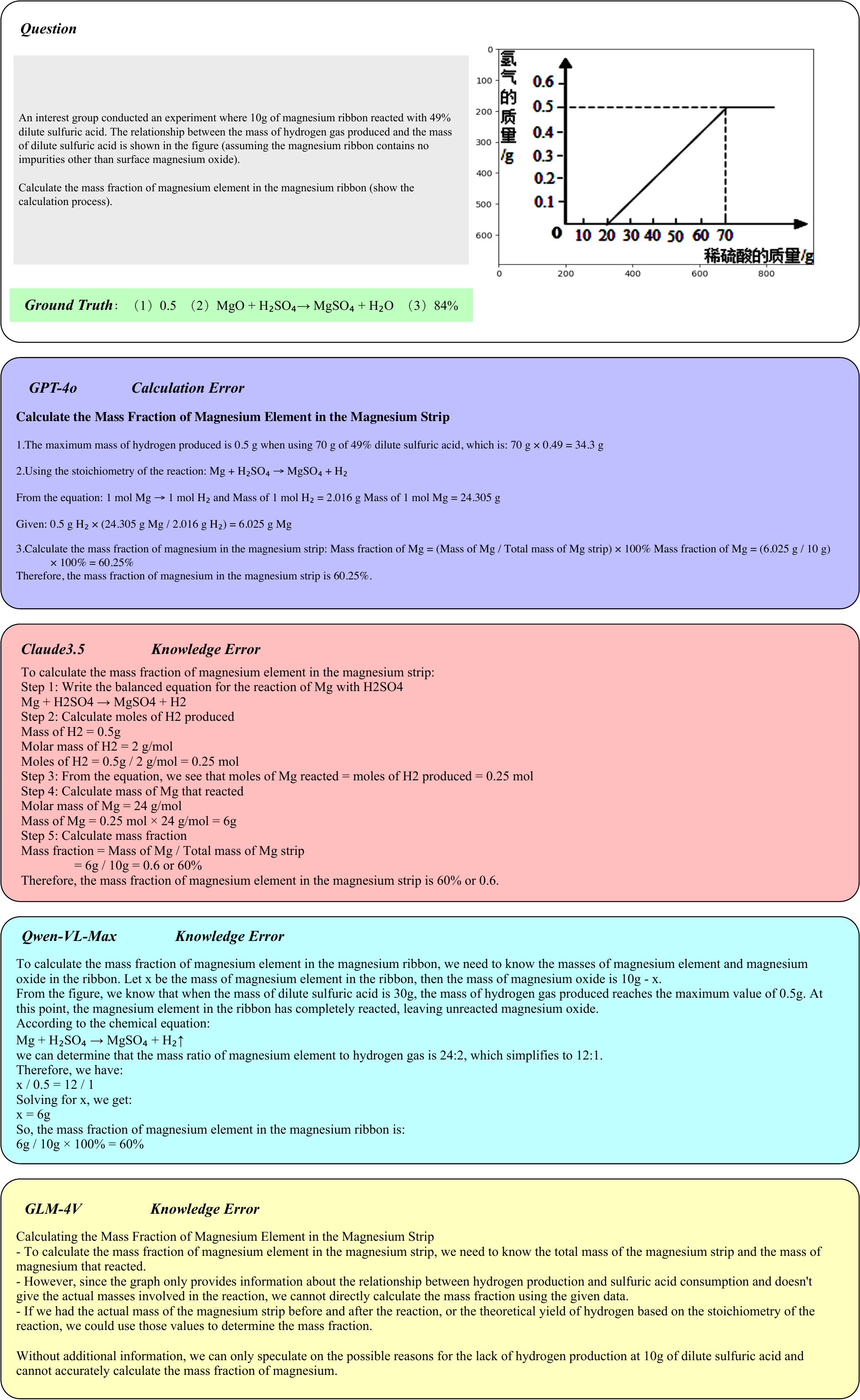}
    \caption{Cases of errors in the chemistry part of \data for several classic close-source models.}
    \label{fig:chemistry_calculation}
\end{figure}

\begin{figure}[hbpt]
    \centering
    \includegraphics[width=1.0\textwidth]{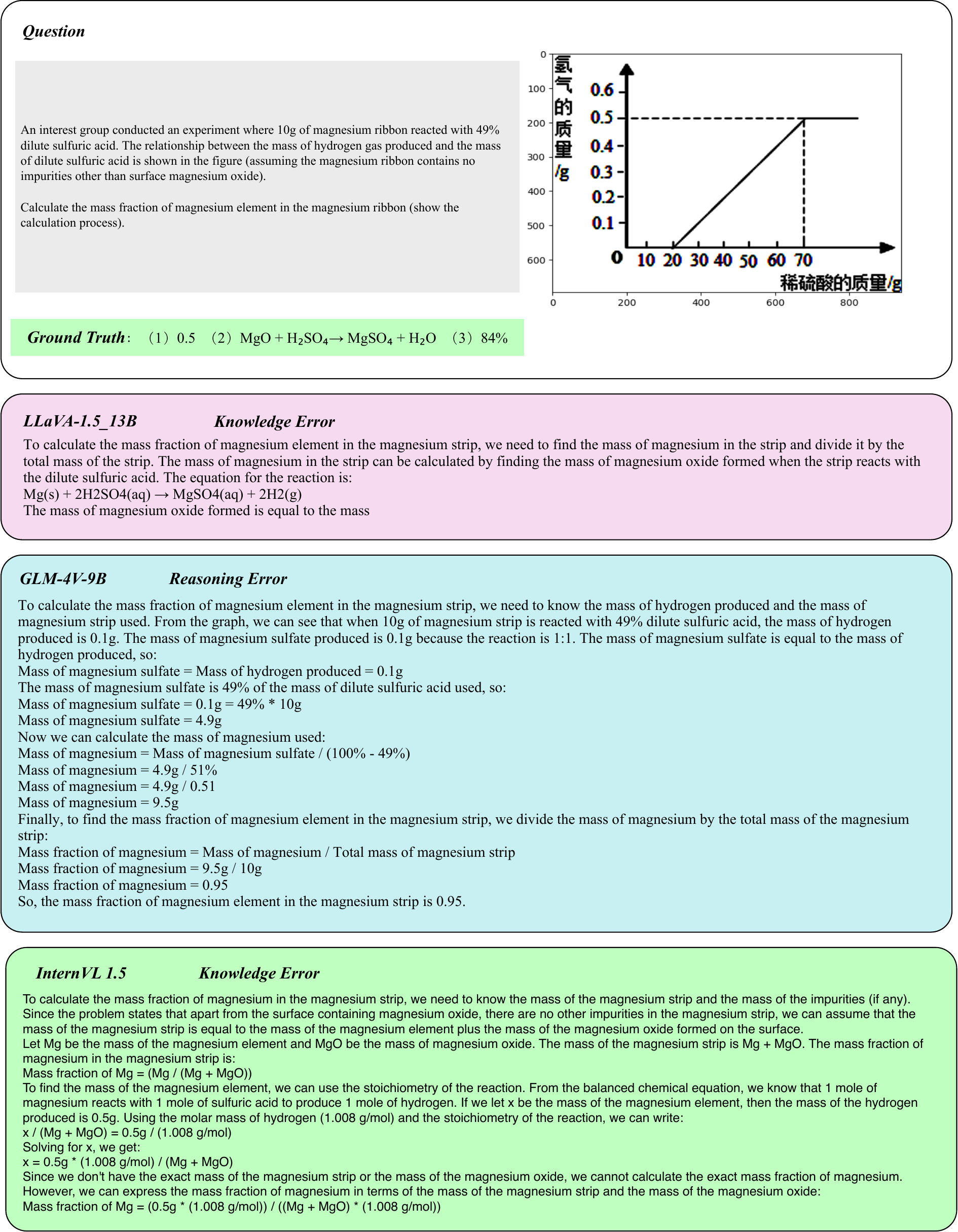}
    \vspace{-2mm}
    \caption{Cases of errors in the chemistry part of \data for several classic open-source models.}
    \label{fig:chemistry_calculation2}
    \vspace{-3mm}
\end{figure}

\begin{figure}[hbpt]
     \centering
      \includegraphics[width=0.8\textwidth, height=0.97\textheight, keepaspectratio]{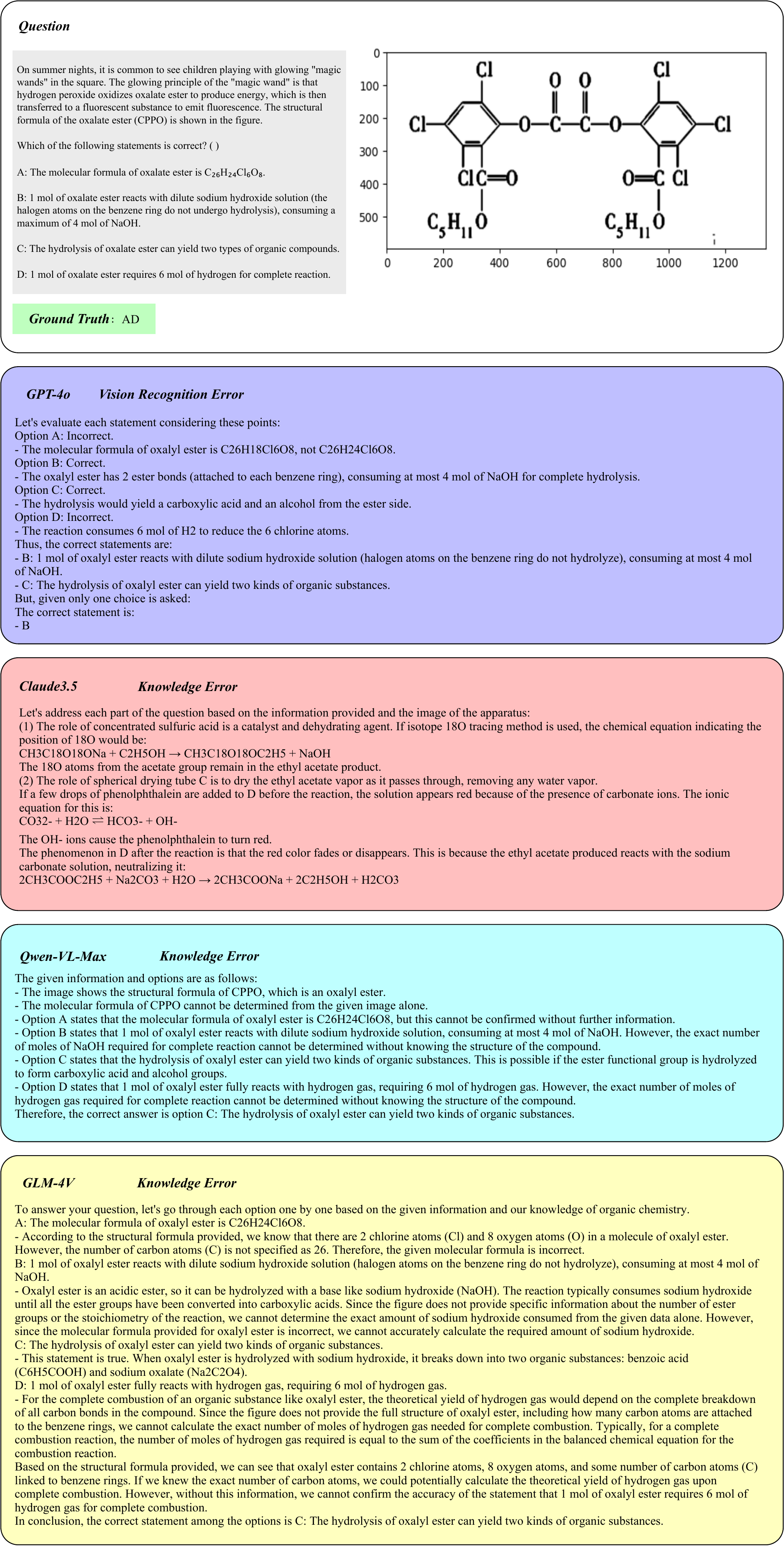}
     \caption{Cases of errors in the chemistry part of \data for several classic close-source models.}
     \label{fig:chemistry_vision}
 \end{figure}

\begin{figure}[hbpt]
     \centering
     \includegraphics[width=1.0\textwidth]{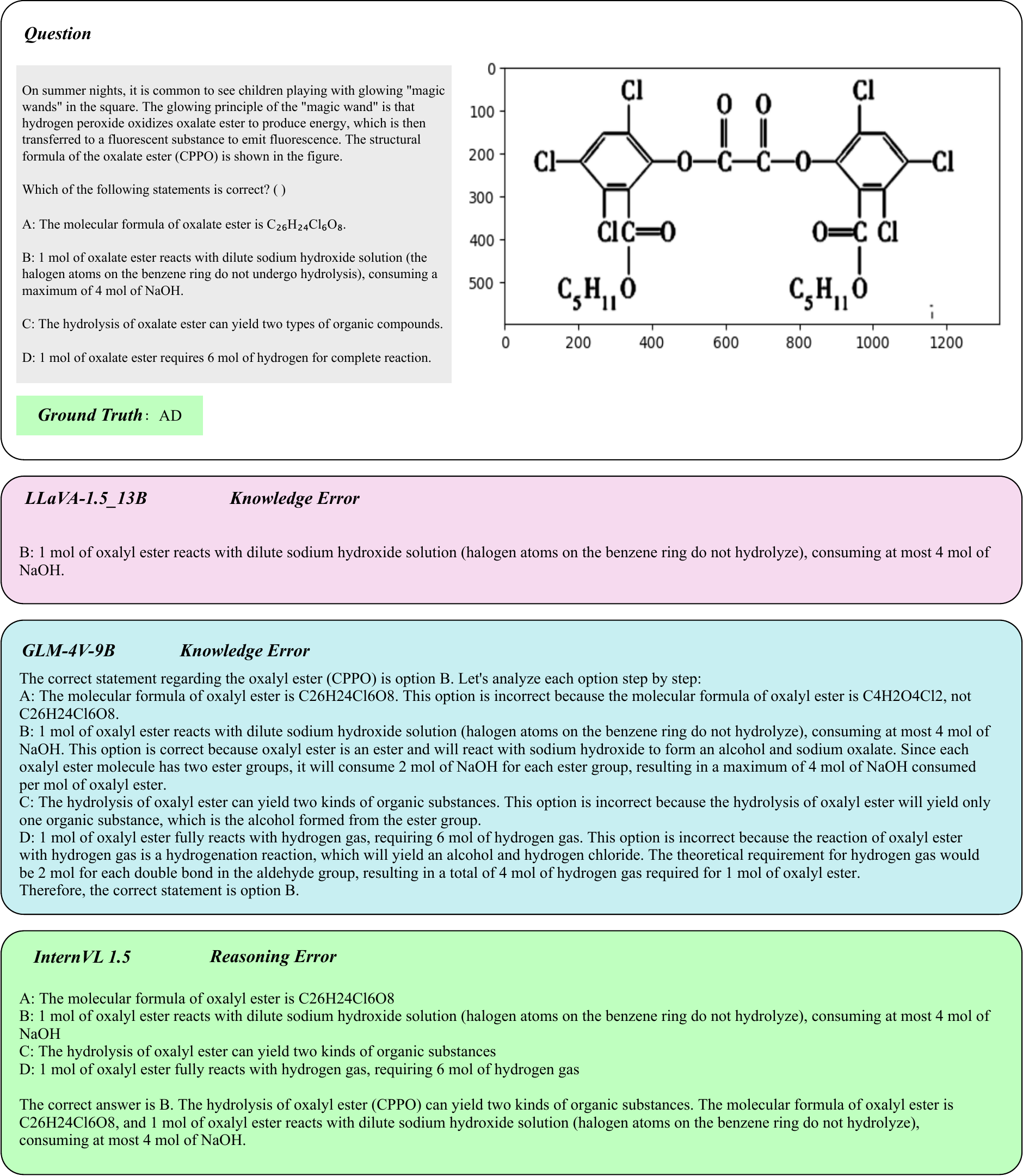}
     \vspace{-2mm}
     \caption{Cases of errors in the chemistry part of \data for several classic open-source models.}
     \label{fig:chemistry_vision2}
     \vspace{-3mm}
\end{figure}

\begin{figure}[hbpt]
     \centering
     \includegraphics[width=\textwidth, height=0.97\textheight, keepaspectratio]{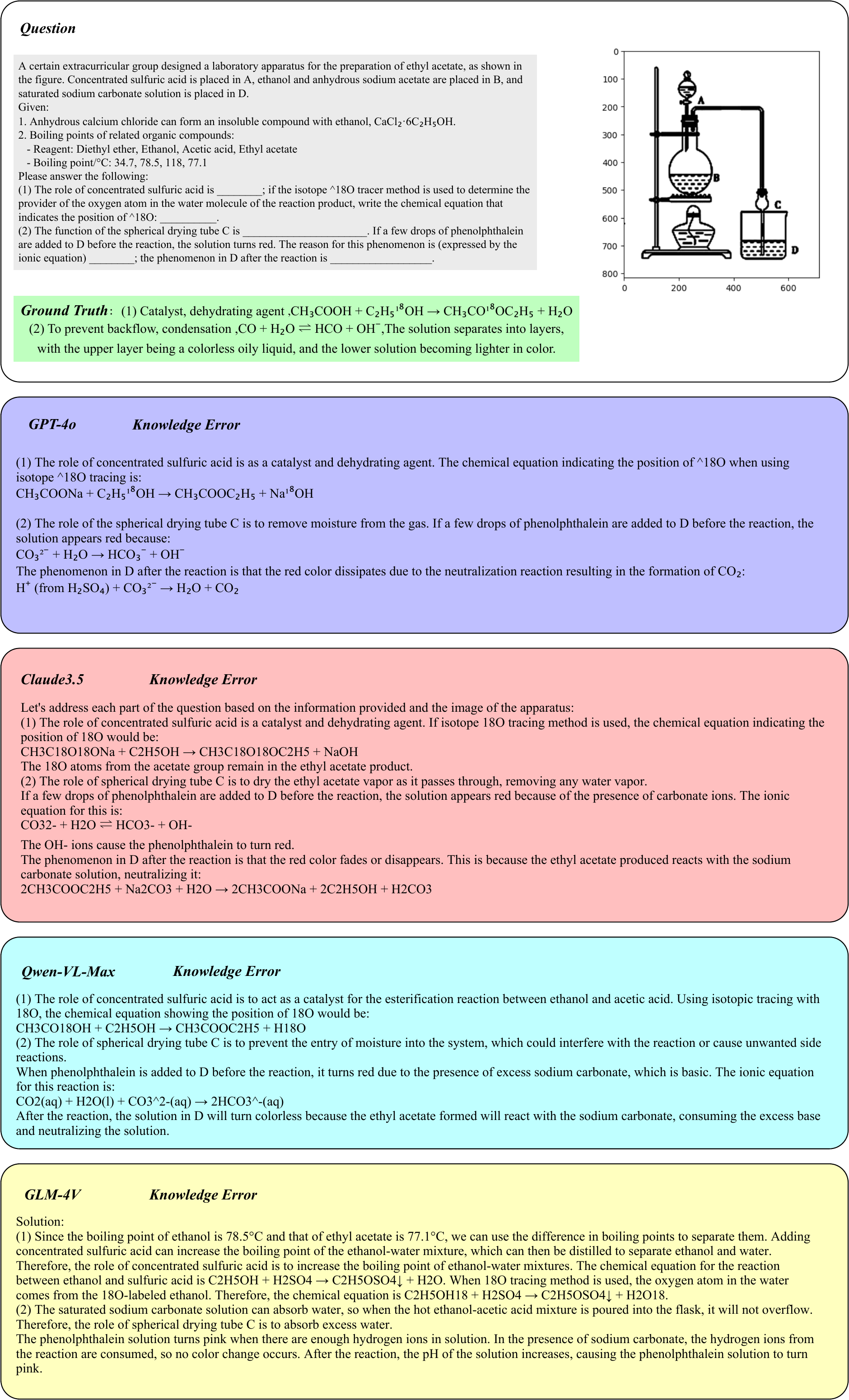}
     \caption{Cases of errors in the chemistry part of \data for several classic close-source models.}
     \label{fig:chemistry_knowledge}
 \end{figure}

\begin{figure}[hbpt]
     \centering
     \includegraphics[width=1.0\textwidth]{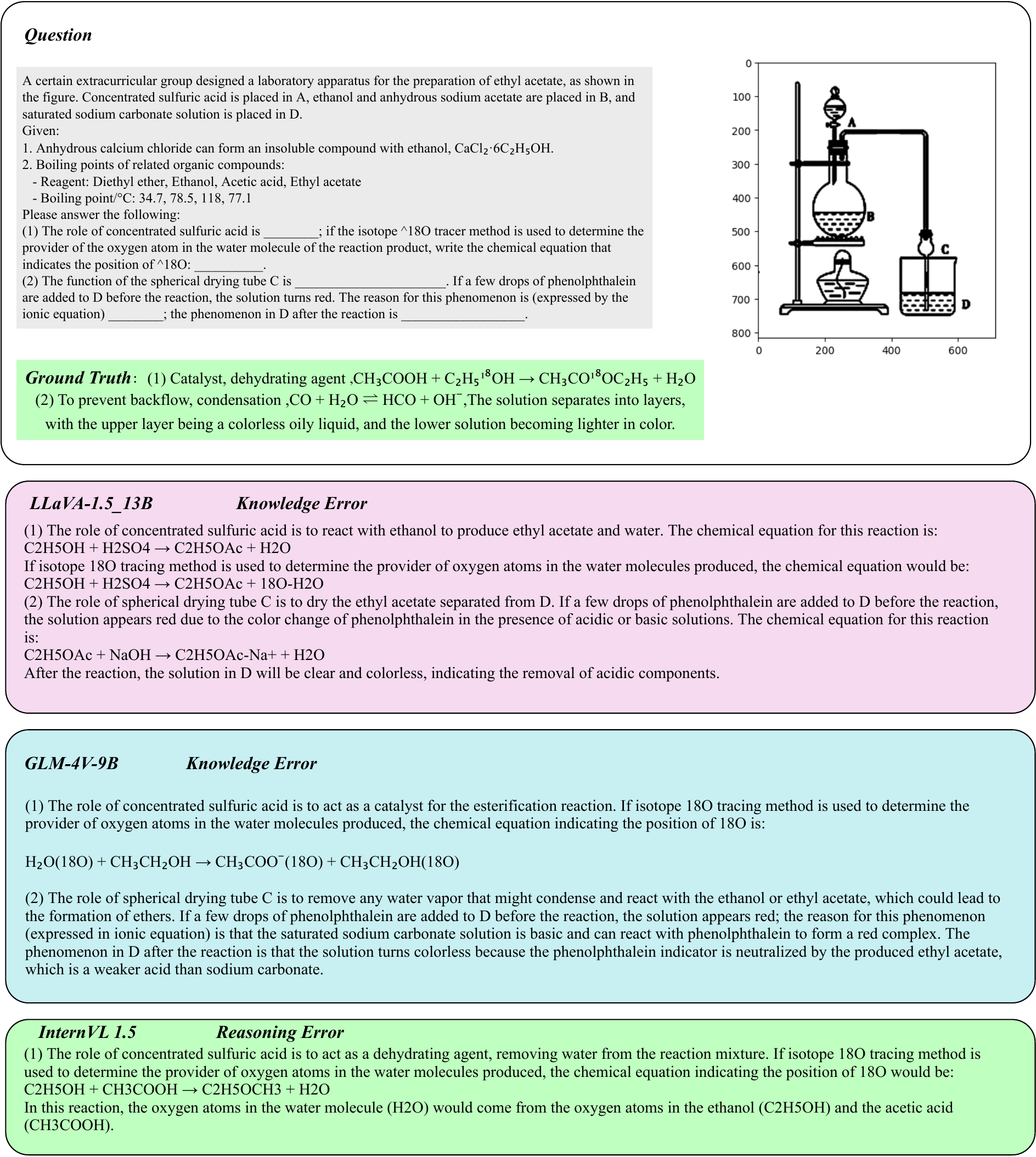}
     \vspace{-2mm}
     \caption{Cases of errors in the chemistry part of \data for several classic open-source models.}
     \label{fig:chemistry knowledge2}
     \vspace{-3mm}
 \end{figure}

\begin{figure}[hbpt]
     \centering
     \includegraphics[width=1.0\textwidth]{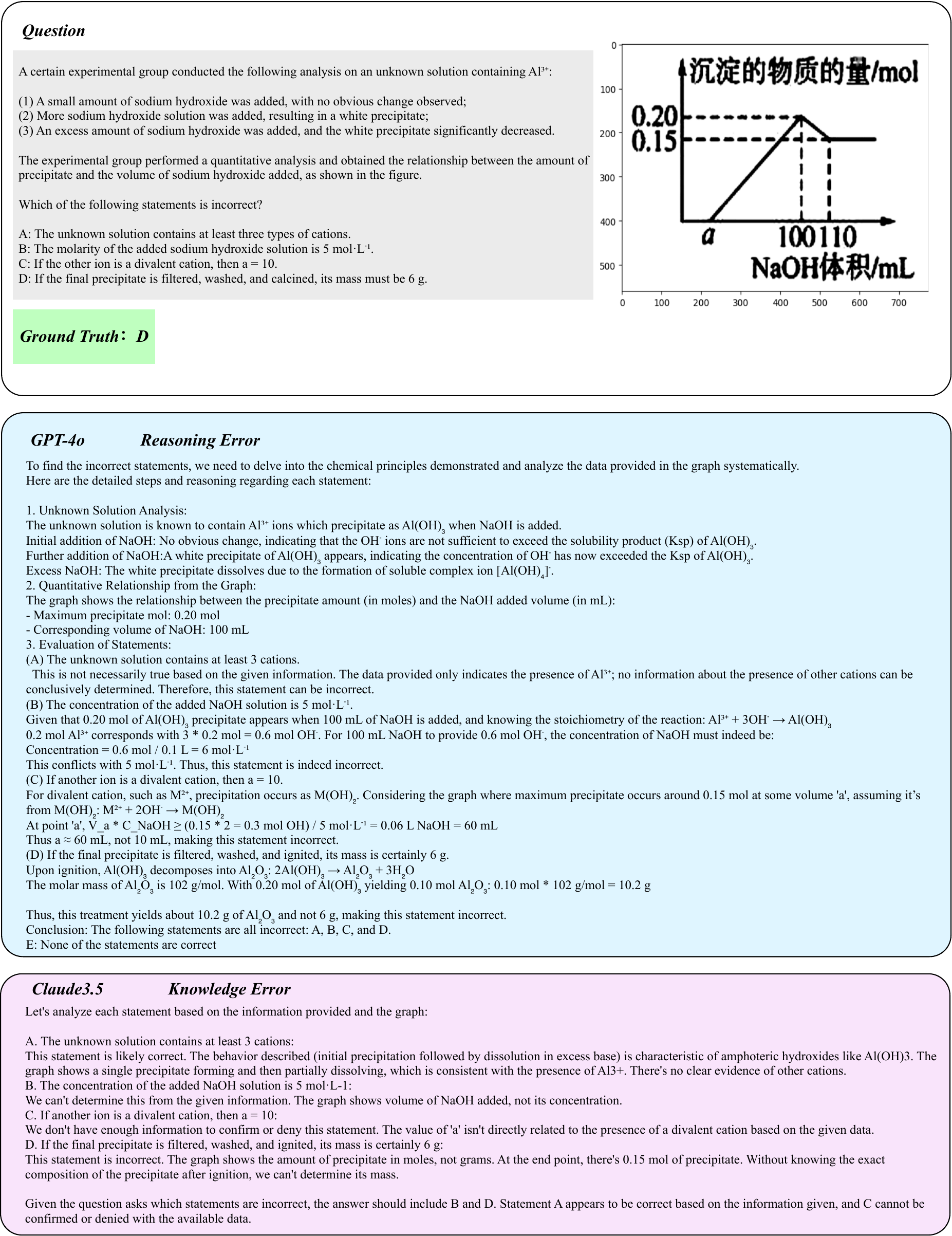}
     \vspace{-2mm}
     \caption{Cases of errors in the chemistry part of \data for several classic close-source models.}
     \label{fig:chemistry reason}
     \vspace{-3mm}
 \end{figure}

\begin{figure}[hbpt]
     \centering
     \includegraphics[width=1.0\textwidth]{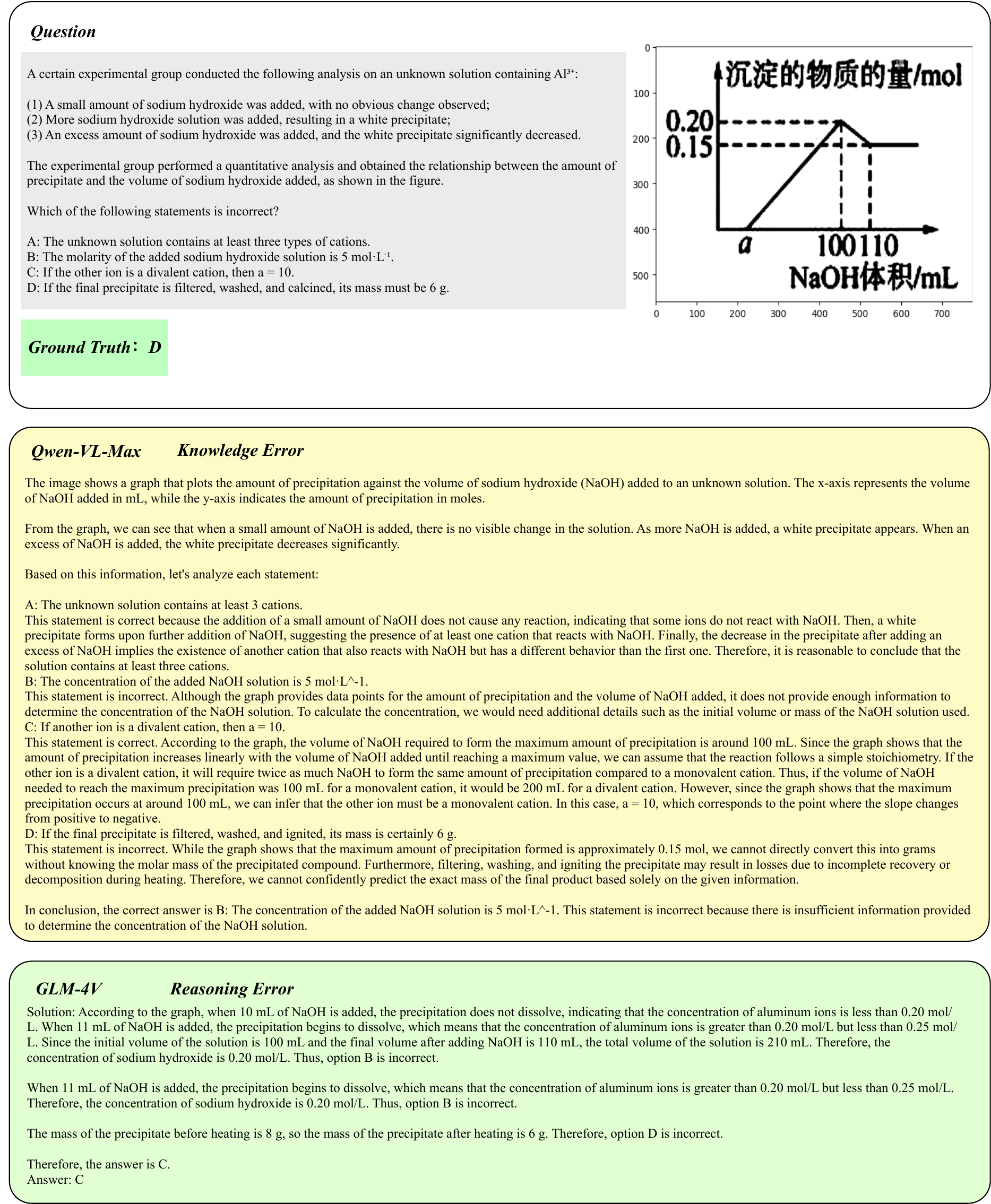}
     \vspace{-2mm}
     \caption{Cases of errors in the chemistry part of \data for several classic close-source models.}
     \label{fig:chemistry reason2}
     \vspace{-3mm}
 \end{figure}

\begin{figure}[hbpt]
     \centering
     \includegraphics[width=1.0\textwidth]{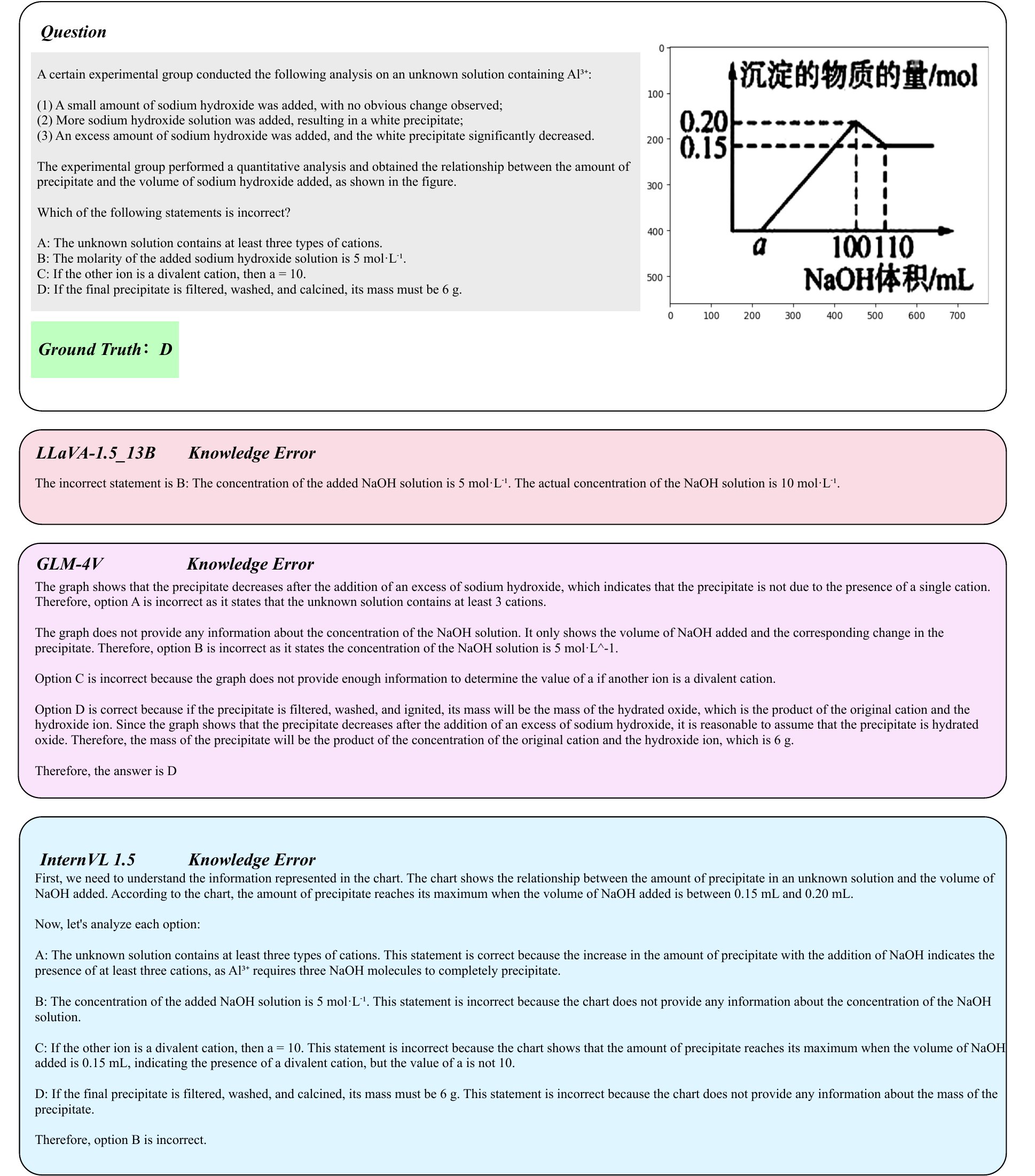}
     \vspace{-2mm}
     \caption{Cases of errors in the chemistry part of \data for several classic open-source models.}
     \label{fig:chemistry reason3}
     \vspace{-3mm}
 \end{figure}

We conduct rigorous tests on a series of open-source and close-source models on \data and perform a detailed analysis of the models' responses. These errors in the models' answers can be classified into five categories: reasoning error, vision recognition error, knowledge error, calculation error, and question misunderstood error. We present examples of these five error types across the disciplines of mathematics, physics, and chemistry, with a specific focus on errors made by GPT-4o. Additionally, we demonstrate error examples from other representative close-source models such as GLM-4V, Qwen-VL-max, and Claude 3.5, as well as open-source models like LLAVA-1.5 , GLM-4V-9B and InternVL-Chat-1.5. Notably, it should be noted that the types of errors made by these models in response to the same questions can differ from those made by GPT-4o. This analysis helps to underline the varied challenges faced by different models in processing complex scientific questions, providing insight into their respective strengths and limitations. Figure~\ref{fig:math_vision} to Figure~\ref{fig:math_knowledge2} demonstrate cases of errors from representative models in the mathematical part of \data. Figure~\ref{fig:physics_calculation1} to Figure~\ref{fig:physics_vision3} show the incorrect answers in the physics section. Figure~\ref{fig:chemistry_calculation} to Figure~\ref{fig:chemistry reason3} demonstrate the errors in the chemistry section.

\clearpage

\newpage

\end{document}